\documentclass[journal]{IEEEtran}

\makeatletter
\newif\if@restonecol
\makeatother

\usepackage[linesnumbered,ruled,vlined]{algorithm2e}
\SetKwBlock{Critic}{Update critic network:}{}
\SetKwBlock{Actor}{Update actor network:}{}
\usepackage{algpseudocode}
\usepackage{amsmath}
\usepackage{amssymb}
\usepackage{graphicx}
\usepackage{floatrow}
\usepackage{float}
\newfloatcommand{capbtabbox}{table}[][\FBwidth]
\usepackage{subfigure}
\usepackage{caption}
\usepackage[justification=centering]{caption}
\usepackage{booktabs}
\usepackage{threeparttable}
\usepackage{multirow}
\usepackage{array}
\usepackage{hyperref}
\hypersetup{hypertex=true,
            colorlinks=true,
            linkcolor=blue,
            anchorcolor=blue,
            citecolor=blue}

\ifCLASSINFOpdf
\else
\fi
\begin{document}
\title{Recursive Least Squares Advantage \\ Actor-Critic Algorithms}

\author{Yuan Wang, Chunyuan Zhang, Tianzong Yu, Meng Ma
\thanks{This work was supported by the National Natural Science Foundation of China under Grant 61762032 and Grant 11961018. (Corresponding author: Chunyuan Zhang).}
\thanks{Y. Wang, C. Zhang, T. Yu and M. Ma are with the School of Computer Science and Technology, Hainan University, Haikou 570228, China (email: 20085400210073@hainanu.edu.cn; 990721@hainanu.edu.cn; 20081200210007@hainanu.edu.cn; 20085400210053@hainanu.edu.cn).}
}

\markboth{IEEE Transactions on Systems, Man, and Cybernetics: Systems}%
{Shell \MakeLowercase{\textit{et al.}}: Bare Demo of IEEEtran.cls for IEEE Journals}

\maketitle

\begin{abstract}
As an important algorithm in deep reinforcement learning, advantage actor critic (A2C)
has been widely succeeded in both discrete and continuous control tasks with raw pixel inputs, but its sample
efficiency still needs to improve more. In traditional reinforcement learning, actor-critic algorithms generally use the recursive
least squares (RLS) technology to update the parameter of linear function approximators for accelerating
their convergence speed. However, A2C algorithms seldom use this technology to train deep neural
networks (DNNs) for improving their sample efficiency. In this paper, we propose two novel RLS-based A2C
algorithms and investigate their performance. Both proposed algorithms, called RLSSA2C and RLSNA2C,
use the RLS method to train the critic network and the hidden layers of the actor network.
The main difference between them is at the policy learning step. RLSSA2C uses an ordinary first-order gradient descent algorithm
and the standard policy gradient to learn the policy parameter.
RLSNA2C uses the Kronecker-factored approximation, the RLS method  and the natural policy gradient
to learn the compatible parameter and the policy parameter. In addition, we analyze the complexity
and convergence of both algorithms, and present three tricks for further improving their convergence speed.
Finally, we demonstrate the effectiveness of both algorithms on 40 games in the Atari 2600 environment and 11 tasks in the MuJoCo environment.
From the experimental results, it is shown that our both algorithms have better sample efficiency than the vanilla
A2C on most games or tasks, and have higher computational efficiency than other two
state-of-the-art algorithms.
\end{abstract}

%
\IEEEpeerreviewmaketitle

\section{Introduction}
Reinforcement learning (RL) is an important machine learning methodology for solving sequential decision-making problems. In RL, the agent
aims to learn an optimal policy for maximizing the cumulative return by interacting with the initially unknown environment \cite{1}.
For the past 40 years, RL has roughly experienced three historical periods, namely, tabular representation RL (TRRL), linear function approximation RL (LFARL) and deep RL (DRL).  TRRL can only solve a few simple problems with small-scale discrete state and action spaces. LFARL
can only solve some control tasks with low-dimensional continuous state and action spaces since its approximation capability
is still limited \cite{2}. In recent years, by combining with various deep neural networks (DNNs), DRL has shown a huge potential to solve real-world complex problems \cite{3}-\cite{6} and has received more and more research interest. Similar to LFARL classified in \cite{7},  DRL can also be classified into three categories: deep value function approximation (DVFA)  \cite{8,9}, deep policy search (DPS)  \cite{10} and deep actor-critic (DAC) methods \cite{11}-\cite{17}. DAC algorithms can be viewed as a hybrid of DVFA and DPS. They generally have a critic for policy evaluation and an actor for policy learning. Among three classes of DRL methods, DAC is more effective than DVFA and DPS for online learning real-world problems, and thus has received extensive attention.

In recent years, many novel DAC algorithms have been proposed. According to the policy type used in the actor, they can be roughly divided into two main subclasses. One subclass is the DAC algorithms with the deterministic policy. The actor-critic algorithms with the deterministic policy gradient (DPG) are first proposed in \cite{11}. Based on this work,  a few variants, such as deep DPG (DDPG) \cite{12}, recurrent DPG (RDPG) \cite{13} and twin delayed DDPG (TD3) \cite{14}, have been suggested recently, but they can  solve only continuous  control tasks. The other subclass is the DAC algorithms with the stochastic policy gradient. In this subclass, the most famous algorithm is perhaps the asynchronous advantage actor-critic (A3C) algorithm \cite{15}. A3C employs multiple workers to interacting with each own environment in parallel, uses the accumulated samples of each worker to update the shared model and uses the parameters of the shared model to update the model of each worker asynchronously. Compared with DPG-type DAC algorithms, A3C can solve both discrete and continuous control tasks without the experience replay. However, A3C doesn't work better than its synchronized version \cite{18}, namely the synchronous advantage actor-critic (A2C) algorithm \cite{15}, which uses all samples obtained by each worker to update the shared model and uses the shared model to select the action of each worker synchronously. A2C is easier to implement and has become a baseline algorithm in OpenAI. Therefore, we will focus on A2C in this paper.

How to improve the sample efficiency of the agent is an open issue in RL. Although A2C and A3C are very excellent, their sample efficiency still needs to improve more. In recent years, there have been some studies such as combining with the experience replay \cite{19,20}
and the entropy maximization \cite{21} to tackle this problem. Here we mainly focus on another way in which researchers try to use more advanced optimization methods for gradient updates. Schulman et al. propose the proximal policy optimization (PPO) \cite{16}, which uses a novel objective with clipped probability ratios and forms a pessimistic estimate of the performance of the policy. Compared with the trust region policy optimization (TRPO) \cite{17}, PPO is much simpler to implement. Byun et al. propose the proximal policy gradient (PPG) \cite{22}, which shows that the performance similar to PPO can be obtained by using the gradient formula from the original policy gradient theorem. Wu et al. propose the actor critic using Kronecker-factored trust region (ACKTR), which uses a Kronecker-factored approximation \cite{23} to the natural policy gradient (NPG) that allows the covariance matrix of the gradient to be inverted efficiently \cite{24}. These algorithms are more effective than the vanilla A2C or A3C with ordinary first-order optimization algorithms, but have higher computational complexities and run slowly. By using Kostrikov's source code \cite{25} to test on  Atari games, we find that PPO and ACKTR are only 1/10 and 7/12 as fast as A2C with RMSProp \cite{26}.

In LFARL, traditional actor-critic algorithms usually use the recursive least squares (RLS) method to improve their convergence performance.
In 1996, Bradtke and Barto first propose the least squares temporal difference (LSTD) algorithm and define a recursive version (namely RLSTD) \cite{27}.
In 1999, Konda and Tsitsiklis first propose a class of two-time-scale actor-critic algorithms with linear function approximations, and point out that it is possible to use LSTD for policy evaluation \cite{28}. After that, Xu et al. propose an actor-critic algorithm by using the  RLSTD($\lambda$) algorithm as the critic \cite{29}.  In 2003, Peter et al. first propose the natural actor-critic (NAC) algorithm, which uses the LSTD-Q($\lambda$) algorithm for policy evaluation and uses NPG for policy learning \cite{30}. On this basis, Park et al. propose the RLS-based NAC algorithm \cite{31}, and Bhatnagar et al. provide four actor-critic algorithms as well as their convergence proofs \cite{32}. There are some other RLS-based  actor-critic algorithms such as
KDHP \cite{7} and CIPG \cite{33}. RLS has been the baseline optimization method for traditional actor-critic algorithms.
However, to the best of our knowledge, there aren't any RLS-based DAC algorithms to be proposed, since the DNN approxiamtion is much more complicated than the linear function approximation. In our previous work \cite{34}, we propose a class of RLS optimization algorithms for DNNs and validate their effectiveness on some classification benchmark datasets. Whereas, there is a big difference between RL and supervised learning, and there are still some
obstacles we need to overcome.

In this paper, we try to introduce the RLS optimization into the A2C algorithm.
We propose two RLS-based A2C algorithms, called RLSSA2C and RLSNA2C,
respectively. Both of them use the same loss function as the vanilla A2C, and employ the
RLS method to optimize their critic networks and the hidden layers of their actor networks.
The main difference between them is the policy learning. RLSSA2C uses the standard policy gradient (SPG) and an ordinary first-order
gradient descent algorithm to update the policy parameters, and
RLSNA2C uses the NPG, the Kronecker-factored approximation and the RLS method
to learn the compatible parameter and the policy parameters.
We show that their computational complexities have the same order as the vanilla A2C with
the first-order optimization algorithm. In addition, we also provide some tricks for accelerating their convergence speed. Finally, we demonstrate their effectiveness on 40 games in the Atari 2600 environment and
11 tasks in the MuJoCo environment. Experimental results show that
both RLSSA2C and RLSNA2C have better sample efficiency  than the vanilla A2C  on most games.
Furthermore, they can achieve a faster running speed than PPO and ACKTR.

The rest of this paper is organized as follows.
In Section \uppercase\expandafter{\romannumeral2}, we introduce some background knowledge.
In Section \uppercase\expandafter{\romannumeral3}, we present the detail derivation of our proposed algorithms.
In Section \uppercase\expandafter{\romannumeral4}, we analyze the computational complexity and convergence of our proposed algorithms,
and provide three tricks for further improving their convergence speed and solution.
In Section \uppercase\expandafter{\romannumeral5}, we demonstrate the effectiveness of our proposed algorithms on Atari games and MuJoCo tasks.
Finally, we conclude our work in Section \uppercase\expandafter{\romannumeral6}.

\section{Background}
In this section, we briefly review Markov decision processes (MDPs), stochastic policy gradient, convolutional neural networks (CNNs) and the vanilla A2C algorithm. In addition, we also introduce some notations used in this paper.
\subsection{Markov Decision Process}
In RL, a sequential decision-making problem is generally formulated into a Markov decision process (MDP)
 defined as $\left \langle \mathcal{S},\mathcal{A},p,r,\gamma \right \rangle$,  where $\mathcal{S}$ is the state space, $\mathcal{A}$ is the action space, $p(s'_t|s_t,a_t)\in[0,1]$ and $r_t\in \mathcal{R}$ are the state-transition probability distribution and the immediate reward from the state $s_t$ to the next state $s'_t$ by taking the action $a_t$ at time step $t$, and $\gamma\in(0,1]$ is the discount factor.
The action $a_t$ is selected by a policy, which can be either stochastic or deterministic. In this paper,
we focus on the former and use a tensor or matrix $\Theta$ to parameterize it. A stochastic policy $\pi(a_t|s_t;\Theta)$  describes the probability distribution of taking $a_t$ in $s_t$.

 For an MDP, the goal of RL is to find an optimal policy $\pi^*$ (also an optimal policy parameter $\Theta^*$) to maximize the cumulative expected return $J(\Theta)$ from an initial state $s_0$, namely
\begin{equation}
\Theta^*=\mathop{\arg\!\max}\limits_{\Theta}J(\Theta)=\mathop{\arg\!\max}\limits_{\Theta}\mathbb{E}_{\pi}\bigg[\sum_{t=0}^{\infty}\gamma^tr_t \big|s_0 \bigg]
\end{equation}
Unfortunately, $J(\Theta)=\mathbb{E}_{\pi}[\sum_{t=0}^{\infty}\gamma^t r_t|s_0]$ is difficult to be calculated directly, since $p(s'|s_t,a_t)$ is unknown in RL, and $s'_t$ and $r_t$ can be obtained only by the agent's interaction with the environment.
DRL generally uses the DAC method to solve $\Theta^*$.
At time step $t$, the critic uses the DVFA method to approximate the state-value function $V^{\pi}(s_t)=\mathbb{E}_{\pi}[\sum_{i=0}^{\infty}\gamma^ir_i|s_0=s_t]$ and the action-value function $Q^{\pi}(s_t,a_t)=\mathbb{E}_{\pi}[\sum_{i=0}^{\infty}\gamma^ir_i|s_0=s_t,a_0=a_t]$ for evaluating the performance of the current policy $\pi$,
and then the actor uses the policy gradient to update $\Theta_t$.

\subsection{Stochastic Policy Gradient}
Currently, there are two main types of policy gradients: SPG and NPG.
From the policy gradient theorem \cite{35}, SPG can be calculated as
\begin{equation}
\nabla_{\Theta_t}J(\Theta_t) = \mathbb{E}_{\pi}\big[A^{\pi}(s_t,a_t)\nabla_{\Theta_t}\textrm{log}\pi(a_t|s_t;\Theta_t)\big]
\end{equation}
where $\nabla_{\Theta_t}J(\Theta_t)$ denotes $ \partial J(\Theta_t)/\partial \Theta_t$, and  $A^{\pi}(s_t,a_t)$ is the advantage function. $A^{\pi}(s_t,a_t)$ measures how much better than the average it is to take an action $a_t$ \cite{36}.
It is defined as
\begin{equation}
A^{\pi}(s_t,a_t)=Q^{\pi}(s_t,a_t)-V^{\pi}(s_t)
\end{equation}
Unlike SPG, NPG does not follow the steepest direction in the policy parameter space but the steepest direction with respect to the Fisher information metric \cite{37}. It is defined as
\begin{equation}
\tilde{\nabla}_{\Theta_t}J(\Theta_t) = (\textrm{F}(\Theta_t))^{-1} \nabla_{\Theta_t}J(\Theta_t)
\end{equation}
where $\textrm{F}(\Theta_t)$ is the  Fisher information matrix defined by
\begin{equation}
\hspace{-0.15cm}\textrm{F}(\Theta_t)\!=\!\mathbb{E}_{\pi}\big[v(\nabla_{\!\Theta_t}\!\textrm{log}\pi(a_t|s_t;\!\Theta_t\!))
v(\nabla_{\Theta_t}\!\textrm{log}\pi(a_t|s_t;\!\Theta_t\!))^{\!\textrm{T}}\big]
\end{equation}
where $v(\cdot)$ denotes reshaping the given tensor or matrix into
a column vector.
To avoid computing the inverse of $\textrm{F}(\Theta_t)$, (4) can also be redefined as
\begin{equation}
\tilde{\nabla}_{\Theta_t}J(\Theta_t) = m(w_t)
\end{equation}
where  $m(\cdot)$ denotes reshaping the given vector into
a matrix or tensor, and $w_t$ is the parameter of the linear compatible function approximator \cite{37} defined by
\begin{equation}
\tilde{A}(s_t,a_t;w_t) =w_t^\textrm{T} v\big(\nabla_{\Theta_t}\log\pi(a_t|s_t;\Theta_t)\big)
\end{equation}
$\tilde{A}(s_t,a_t;w_t)$ is the compatible approximation of $A^{\pi}(s_t,a_t)$.

\subsection{Convolutional Neural  Network}
In DAC, CNNs are widely used to approximate $V^{\pi}(s)$ and $\pi$ for solving control tasks with raw pixel inputs. A CNN generally consists of some convolutional (conv) layers, pooling layers and fully-connected (fc) layers. Since there are no learnable parameters in pooling layers, we only review the forward learning of  conv  layers and fc layers. Let $l$, $M$, $\textrm{X}^l_t$, $\textrm{Z}^l_t$, $\textrm{Y}^l_t$, $\Theta^l_t$ and  $f_l(\cdot)$ denote the current layer, the mini-batch size, the mini-batch input, the pre-activation input, the activation output, the parameter, the activation function in this layer at current time $t$, respectively. For brevity, we omit the bias term of each layer in this paper.

In a conv layer, $\textrm{X}^l_t \in \mathcal{R}^{M\times C_{l-1} \times H_{l-1} \times W_{l-1}  }$ is convolved with the kernel
 $\Theta^l_t \in \mathcal{R}^{C_{l-1}\times C_l \times H_l^k \times W_l^k} $ and  puts through  $f_l(\cdot)$  to form  $\textrm{Y}^l_t
 \in \mathcal{R}^{M\times C_{l} \times H_{l} \times W_{l}}$, where $C_l$, $H_l$, $W_l$, $H_l^k$ and $W_l^k$ denote
 the number of output channels, the output image height, the output image width, the kernel height and the kernel width, respectively.
Let $\textrm{\v{X}}^l_{t(:,:,:,:,h,w)} \in \mathcal{R}^{M \times C_{l-1}\times H_l^k \times W_l^k}$ denote the input selection
of the output pixel $\textrm{Y}^l_{t(:,:,h,w)}$. Reshape $\textrm{\v{X}}^l_t$ and  $\Theta^l_t$ as $\textrm{\^{X}}^l_t \in \mathcal{R}^{M\times C_{l-1} H_l^k W_l^k\times H_lW_l}$ and ${\hat{\Theta}}^l_t \in \mathcal{R}^{ C_{l-1} H_l^k W_l^k\times C_l}$, respectively.
Then, $\textrm{Y}^l_{t(:,:,j)}$ is defined as
\begin{equation}
\textrm{Y}^l_{t(:,:,j)} = f_l(\textrm{Z}^l_{t(:,:,j)}) = f_l(\textrm{\^X}^l_{t(:,:,j)}{\hat\Theta}^l_t)
\end{equation}

In an fc layer, $\textrm{X}^l_t\in \mathcal{R}^{M\times N_{l-1}}$ is weighed to connect all output neurons by $\Theta^l_t\in \mathcal{R}^{N_{l-1}\times N_{l}}$  and puts through  $f_l(\cdot)$  to form  $\textrm{Y}^l_t \in \mathcal{R}^{M\times N_{l}}$, where $N_l$ denotes the number of output neurons. Namely, $\textrm{Y}^l_t$ is defined as
\begin{equation}
\textrm{Y}^l_t = f_l(\textrm{Z}^l_t) = f_l(\textrm{X}^l_t\Theta^l_t)
\end{equation}

\subsection{Advantage Actor Critic}
A2C is an important baseline algorithm in OpenAI. In A2C, there are $N$ parallel workers, a shared
critic network and a shared actor network. The critic network and the actor network can be joint or disjoint.
If both networks are joint, they will share lower layers but have distinct output layers.

The algorithm flow of  A2C can be summarized as follows.
At current iteration step $t$, it lets each worker interact with each own environment for $T$ timesteps, and uses all
state-transition pairs to form the mini batch $\mathcal{M}_t=\{(s_{t,i}^{(k)}, a_{t,i}^{(k)},{s'}_{t,i}^{(k)},r_{t,i}^{(k)},d_{t,i}^{(k)})\}_{i=1,\cdots,N}^{k=1,\cdots,T}$, where
$d_{t,i}^{(k)} \in \{0,1\}$ denotes that the next state ${s'}_{t,i}^{(k)}$ of  the $i^{th}$ worker is the terminal state or not at the $k^{th}$ timestep, and
the mini-batch size $M$ is equal to $NT$.
Then, it calculates the loss function defined by
\begin{equation}
L(\Psi_t,\Theta_t)= L(\Psi_t)+L(\Theta_t)+\eta E(\Theta_t)
\end{equation}
where $\Psi_t$ and $L(\Psi_t)$ are the parameter and the loss function of the critic network,
$\Theta_t$ and $L(\Theta_t)$ are the parameter and the loss function of the actor network,
and $\eta E(\Theta_t)$ is the entropy regularization term with a small $\eta>0$.
$L(\Psi_t)$ is defined as
\begin{equation}
L(\Psi_t)=\frac{1}{2NT}\big\|A(\textrm{S}_t,\textrm{A}_t)\big\|^2_F
\end{equation}
where  $\textrm{S}_t=[s_{t,1}^{(1)}, \cdots, s_{t,N}^{(T)}]^\textrm{T}$, $\textrm{A}_t=[a_{t,1}^{(1)}, \cdots, a_{t,N}^{(T)}]^\textrm{T}$, and
$A(\textrm{S}_t,\textrm{A}_t) \in \mathcal{R}^{NT}$ denotes the estimate value of  $A^{\pi}(\textrm{S}_t,\textrm{A}_t)$,
which is calculated as
\begin{equation}
A(\textrm{S}_t,\textrm{A}_t) = Q(\textrm{S}_t,\textrm{A}_t) - V(\textrm{S}_t;\Psi_t)
\end{equation}
where $V(\textrm{S}_t;\Psi_t)$ and $Q(\textrm{S}_t,\textrm{A}_t)$ denote the estimate values of $V^{\pi}(\textrm{S}_t)$  and  $Q^{\pi}(\textrm{S}_t,\textrm{A}_t)$, respectively. The former is the actual output of the critic network, and the latter is the desired output of the critic network. Each element in the latter is calculated as
\begin{equation}
\hspace{-0.2cm}
Q(s_{t,i}^{(k)},a_{t,i}^{(k)}) \!=\!
\!\left\{\!
\begin{aligned}
r_{t,i}^{(T)} \!+\!\gamma(1\!-\!d_{t,i}^{(T)}) V({s'}_{t,i}^{(T)};\Psi_t)~~, &~k\!=\!T\\
r_{t,i}^{(k)} \!+\!\gamma(1\!-\!d_{t,i}^{(k)}) Q({s}_{t,i}^{k+1},a_{t,i}^{k+1}), &~k\!<\!T
\end{aligned}
\right.
\end{equation}
where $V({s'}_{t,i}^{(T)};\Psi_t)$ is also approximated by the critic network.
$L(\Theta_t)$ and $E(\Theta_t)$ are defined as follows
\begin{eqnarray}
L(\Theta_t)\!=\!-\frac{1}{NT}\!\sum_{i=1}^{N}\!\sum_{k=1}^{T}\! A(s_{t,i}^{(k)},a_{t,i}^{(k)})\!\log\!\pi(a_{t,i}^{(k)}|s_{t,i}^{(k)};\!\Theta_t)~\\
E(\Theta_t)\!=\!-\frac{1}{NT}\!\sum_{i=1}^{N}\!\sum_{k=1}^{T}\! \pi(a_{t,i}^{(k)}|s_{t,i}^{(k)};\!\Theta_t)\!\log\!\pi(a_{t,i}^{(k)}|s_{t,i}^{(k)};\!\Theta_t)
\end{eqnarray}
where $\pi(a_{t,i}^{(k)}|s_{t,i}^{(k)};\Theta_t)$  is the actual output of the actor network.
Finally, A2C uses (10) to update $\Psi_t$ and $\Theta_t$ with some first-order optimization algorithms such as RMSProp.

\section{RLS-based Advantage Actor Critic}
In this section, we try to integrate the RLS method into the vanilla A2C with CNNs.
Under the loss function defined by (10), the RLS update rules for four different types of layers used in
critic and actor networks are derived respectively.
On the basis, we propose RLSSA2C and RLSNA2C algorithms.

\subsection{Optimizing Critic Output Layer}
As introduced in Section I, the critic of traditional actor-critic algorithms widely uses LSTD for policy evaluation.
From (11), (12) and (13), $A(\textrm{S}_t,\textrm{A}_t) \in \mathcal{R}^{NT} $ is a temporal difference error vector.
It means we can also use LSTD for updating the critic parameter.
Let $\mathcal{D}_t=\{\mathcal{M}_1,\cdots,\mathcal{M}_t\}$ denote the state-transition mini-batch dataset from
the start to the current iteration step $t$.
On the basis, we define an auxiliary least squares loss function as
\begin{equation}
\tilde{L}(\Psi)=\frac{1}{2NT}\sum_{n=1}^{t}\lambda^{t-n}\big\|A(\textrm{S}_n,\textrm{A}_n)\big\|^2_F
\end{equation}
where $\lambda\in(0,1]$ is the forgetting factor.
Then, the learning problem of the current critic parameter  can be described as
\begin{equation}
\Psi_{t+1}=\mathop{\arg\!\min}\limits_{\Psi}\tilde{L}(\Psi)
\end{equation}

In the critic network of A2C, the output layer is generally a linear fc layer with one output neuron. In other words,
the activation function of this layer is the identity function.
Thus, from (9), $V(\textrm{S}_n;\Psi)$ is calculated as
\begin{equation}
V(\textrm{S}_n;\Psi)=\textrm{X}^l_n\Psi^l
\end{equation}
where $\Psi^l$ is the parameter of this layer. Then, from (12),
(16) can be rewritten as
\begin{equation}
\tilde{L}(\Psi)=\frac{1}{2NT}\sum_{n=1}^{t}\lambda^{t-n}\big\|Q(\textrm{S}_n,\textrm{A}_n) - \textrm{X}^l_n\Psi^l\big\|^2_F
\end{equation}
By the chain rule for $\Psi^l$, $\nabla_{\Psi^l}\tilde{L}(\Psi)$ can be derived as
\begin{equation}
\nabla_{\Psi^l}\tilde{L}(\Psi)=-\frac{1}{NT}\!\sum_{n=1}^{t}\lambda^{t-n}(\textrm{X}^l_n)^{\textrm{T}}\!\big(Q(\textrm{S}_n,\textrm{A}_n) - \textrm{X}^l_n\Psi^l\big) \end{equation}
Let $\nabla_{\Psi^l}\tilde{L}(\Psi)=\textrm{0}$. We can easily obtain the least squares solution of $\Psi^l_{t+1}$, namely
\begin{equation}
\Psi^l_{t+1}=(\textrm{H}^l_{t+1})^{-1}b^l_{t+1}
\end{equation}
where $\textrm{H}_{t+1}^l \in \mathcal{R}^{N_{l-1}\times N_{l-1}}$ and $b_{t+1}^l \in \mathcal{R}^{N_{l-1}}$ are defined as
\begin{eqnarray}
\textrm{H}_{t+1}^l=\frac{1}{NT}\sum_{n=1}^{t}\lambda^{t-n}(\textrm{X}^l_n)^{\textrm{T}}\textrm{X}^l_n ~~~~ \\
b_{t+1}^l=\frac{1}{NT}\sum_{n=1}^{t}\lambda^{t-n}(\textrm{X}^l_n)^{\textrm{T}}Q(\textrm{S}_n,\textrm{A}_n)
\end{eqnarray}

Next, to avoid computing the inverse of $\textrm{H}^l_{t+1}$ and realize online learning, we try to derive the RLS solution of  $\Psi^l_{t+1}$.
Rewrite (22) and (23) as the following incremental update equations
\begin{eqnarray}
\textrm{H}_{t+1}^l=\lambda \textrm{H}^l_{t}+\frac{1}{NT}(\textrm{X}^l_{t})^{\textrm{T}}\textrm{X}^l_{t} ~~~~~\\
b_{t+1}^l=\lambda b^l_{t}+\frac{1}{NT}(\textrm{X}^l_{t})^{\textrm{T}}Q(\textrm{S}_{t},\textrm{A}_{t})
\end{eqnarray}
However, by using the Sherman-Morrison matrix inversion
lemma \cite{38} for (24), the recursive update of $(\textrm{H}^{l}_{t+1})^{-1}$ still includes a new inverse matrix, since the rightmost term in (24) is a matrix product rather than a vector product. In our previous work \cite{34}, we propose an
average-approximation method to tackle this problem and reduce the computation burden. Using this method, we define
\begin{eqnarray}
\bar{x}^l_t=\frac{1}{NT}\sum_{i=1}^{NT}\textrm{X}^l_{t(i,:)}~~~~~~  \\
\bar{q}_t =\frac{1}{NT}\sum_{i=1}^{NT}Q(\textrm{S}_{t(i,:)},\textrm{A}_{t(i,:)})
\end{eqnarray}
where $\textrm{X}^l_{t(i,:)} \in \mathcal{R}^{N_{l-1}}$ and $Q(\textrm{S}_{t(i,:)},\textrm{A}_{t(i,:)}) \in \mathcal{R}$ are the column vectors
sliced from $\textrm{X}^l_t$ and $Q(\textrm{S}_t,\textrm{A}_t)$, respectively.
On the basis, we can rewrite (24) and (25) as follows
\begin{eqnarray}
\textrm{H}_{t+1}^l =\lambda \textrm{H}^l_t+k\bar{x}^l_t (\bar{x}^l_t)^\textrm{T} \\
b_{t+1}^l =\lambda b^l_t+k\bar{x}^l_t \bar{q}_t ~~~
\end{eqnarray}
where $k>0$ is the average scaling factor. Let $\textrm{P}_t = (\textrm{H}_t)^{-1}$. Then, using the Sherman-Morrison matrix inversion
lemma and (17), we finally obtain the following RLS update rules
\begin{eqnarray}
~~~~~~\textrm{P}^l_{t+1}\approx\frac{1}{\lambda}\bigg(\textrm{P}^l_{t}-\frac{k \textrm{P}^l_{t}\bar{x}^l_t(\bar{x}^l_t)^{\textrm{T}}\textrm{P}^l_{t}}{\lambda+k(\bar{x}^l_t)^{\textrm{T}}\textrm{P}^l_{t}\bar{x}^l_t}\bigg)\\
~~~~~~\Psi^l_{t+1}\approx\Psi^l_{t}+\frac{k \textrm{P}^l_{t}\bar{x}^l_t (\bar{q}_t-\bar{v}_t)}{\lambda+k(\bar{x}^l_t)^{\textrm{T}}\textrm{P}^l_{t}\bar{x}^l_t} ~~~
\end{eqnarray}
where $\bar{v}_t$ is defined as
\begin{equation}
\bar{v}_t=\frac{1}{NT}\sum_{i=1}^{NT}V(\textrm{S}_{t(i,:)};\Psi_t)
\end{equation}

In our previous work \cite{34}, we find that the RLS optimization can be converted into a gradient descent algorithm, which is
easier to implement by using PyTorch or TensorFlow. Based on (10)-(12), $\nabla_{\Psi^l_{t}} L(\Psi_t,\Theta_t)$ can be derived as
\begin{equation}
\nabla_{\Psi^l_{t}} L(\Psi_t,\Theta_t) = -\frac{1}{NT} (\textrm{X}_t^l)^{\textrm{T}} (Q(\textrm{S}_t,\textrm{A}_t) - \textrm{X}^l_t\Psi^l_t)
\end{equation}
In (28) and (29), we ever use $k\bar{x}^l_t (\bar{x}^l_t)^\textrm{T}$ and $k\bar{x}^l_t \bar{q}_t$ to
replace $\frac{1}{NT}(\textrm{X}^l_t)^{\textrm{T}}\textrm{X}^l_t$ and $\frac{1}{NT}(\textrm{X}_t^l)^{\textrm{T}} Q(\textrm{S}_t,\textrm{A}_t)$,
respectively. On the basis, we can easily get
\begin{equation}
k\bar{x}^l_t(\bar{q}_t-\bar{v}_t) \approx \frac{1}{NT} (\textrm{X}_t^l)^{\textrm{T}} (Q(\textrm{S}_t,\textrm{A}_t) - \textrm{X}^l_t\Psi^l_t)
\end{equation}
Thus, we can rewrite (31) as  the following gradient update form
\begin{equation}
\Psi^l_{t+1} \approx \Psi^l_{t}-\frac{\textrm{P}^l_{t} \nabla_{\Psi^l_{t}} L(\Psi_t,\Theta_t)}{\lambda+k(\bar{\textrm{x}}^l_t)^{\textrm{T}}\textrm{P}^l_{t}\bar{\textrm{x}}^l_t}
\end{equation}
where $\frac{\textrm{P}^l_{t}}{\lambda+k(\bar{\textrm{x}}^l_t)^{\textrm{T}}\textrm{P}^l_{t}\bar{\textrm{x}}^l_t}$ is the learning rate.
That means we needn't change the loss function defined by (10).

\subsection{Optimizing Actor Output Layer}
In the actor network of A2C, the output layer is also an fc layer, which is to output $\pi(\textrm{A}_t|\textrm{S}_t;\Theta_t)$.
For discrete control problems, A2C often uses the Softmax function to define $\pi$, called the Softmax policy.
For continuous  control problems, A2C usually uses the Gaussian function to define $\pi$, called the Gaussian policy.
Unlike (11),  (14) and (15) are difficult to be converted into least squares loss functions. In fact, for the same reason,
many traditional actor-critic algorithms only use the RLS method to update the critic parameter, but use the SPG
descent method to update the actor parameter. Our first algorithm RLSSA2C will follow this method. Its actor output layer is
optimized by an ordinary first-order optimization algorithm. For example, its update rules based on RMSProp are
defined as follows
\begin{eqnarray}
\textrm{C}^l_{t+1} = \rho\textrm{C}^l_{t} +(1-\rho) \nabla_{\Theta^l_{t}} L(\Psi_t,\Theta_t) \odot  \nabla_{\Theta^l_{t}} L(\Psi_t,\Theta_t)    \\
\Theta^l_{t+1} = \Theta^l_{t}-\frac{\epsilon}{\sqrt{\delta +\textrm{C}^l_{t+1}}} \odot  \nabla_{\Theta^l_{t}} L(\Psi_t,\Theta_t) ~~~~~
\end{eqnarray}
where $\textrm{C}^l_{t}$ is the accumulative squared gradient, $\rho$ is the decay rate, $\epsilon$ is the learning rate, $\delta>0$ is a small constant,
$\nabla_{\Theta^l_{t}} L(\Psi_t,\Theta_t)$ is SPG,  and
$\odot$ denotes the Hadamard product.

As introduced in Section II. \textit{B}, there is another type of policy gradients, namely NPG \cite{30,37}, which has yielded
a few novel NAC algorithms. To avoid computing the inverse of the Fisher information matrix, some traditional NAC algorithms,
often use the RLS method to approximate the compatible parameter $\textrm{W}_t$, and use $\textrm{W}_t$ as NPG for updating the actor parameter \cite{31,32}. Following this way, we define an auxiliary least squares loss function based on $\mathcal{D}_t$  as
\begin{equation}
\tilde{L}(w)=\frac{1}{2NT}\sum_{n=1}^{t}\lambda^{t-n}\big\|A(\textrm{S}_n,\textrm{A}_n) - \tilde{A}(\textrm{S}_n,\textrm{A}_n;w)\big\|^2_F
\end{equation}
where $A(\textrm{S}_n,\textrm{A}_n)$ is calculated by (12).
From (7), each element in $\tilde{A}(\textrm{S}_n,\textrm{A}_n;w)$ is defined as
\begin{equation}
\tilde{A}(\textrm{S}_{n(i,:)},\textrm{A}_{n(i,:)};w)
 = w^\textrm{T} \textrm{G}_{n(i,:)}^{\Theta^l}
\end{equation}
where $\textrm{G}_{n(i,:)}^{\Theta^l}$ denotes $v(\nabla_{\Theta_t^l}\log\pi(\textrm{A}_{n(i,:)}|\textrm{X}^l_{n(i,:)};\Theta_t^l))\in \mathcal{R}^{N_{l-1}N_l}$
for simplifying notations. Then, the learning problem of the current compatible parameter can be described as
\begin{equation}
w_{t+1}=\mathop{\arg\!\min}\limits_{w}\tilde{L}(w)
\end{equation}
Let $\nabla_{w}\tilde{L}(w)=0$. We can easily obtain the least squares solution of $w_{t+1}$, namely
\begin{equation}
w_{t+1}=(\textrm{H}^l_{t+1})^{-1}b^l_{t+1}
\end{equation}
where $\textrm{H}_{t+1}^l \in \mathcal{R}^{N_{l-1}N_l\times N_{l-1}N_l}$ and $b_{t+1} \in \mathcal{R}^{N_{l-1}N_l}$ are defined as follows
\begin{eqnarray}
\textrm{H}_{t+1}^l=\frac{1}{NT}\sum_{n=1}^{t}\lambda^{t-n}(\textrm{G}_{n}^{\Theta^l})^{\textrm{T}}\textrm{G}_{n}^{\Theta^l} ~~~~\\
b_{t+1}^l=\frac{1}{NT}\sum_{n=1}^{t}\lambda^{t-n} (\textrm{G}_{n}^{\Theta^l})^{\textrm{T}} A(\textrm{S}_{n},\textrm{A}_{n})
\end{eqnarray}
where $\textrm{G}_{n}^{\Theta^l}=[\textrm{G}_{n(1,:)}^{\Theta^l},\cdots,\textrm{G}_{n(NT,:)}^{\Theta^l}]^\textrm{T} \in \mathcal{R}^{NT\times N_{l-1}N_l}$.
By using the same derivation method in Section III. \textit{A}, we can easily obtain the RLS update rules for $\textrm{P}_{t+1}^l$ and $w_{t+1}$.
However, here $\textrm{P}_{t+1}^l=\big(\textrm{H}_{t+1}^l\big)^{-1}$ will be $N_l^2$ times as complex as in the critic output layer.

Since the forgetting factor $\lambda$ is 1 or close to 1 in general, $\textrm{H}_{t+1}^l$ can be approximately viewed as a Fisher information matrix.
In addition, by the chain rule, we easily  get
\begin{equation}
\textrm{G}_{n(i,:)}^{\Theta^l} = \textrm{X}_{n(i,:)}^l (\textrm{G}_{n(i,:)}^{\textrm{Z}_{n(i,:)}^l})^\textrm{T}
\end{equation}
where $\textrm{G}_{n(i,:)}^{\textrm{Z}_{n(i,:)}^l} \!=\! \nabla_{\textrm{Z}_{n(i,:)}^l}\!\log\pi(\textrm{A}_{n(i,:)}|\textrm{X}^l_{n(i,:)};\Theta^l)$ and
$\textrm{Z}_{n(i,:)}^l = (\Theta^l)^\textrm{T}\textrm{X}^l_{n(i,:)}$. Then, we have
\begin{equation}
(\textrm{G}_{n(i,:)}^{\Theta^l})^{\textrm{T}}\textrm{G}_{n(i,:)}^{\Theta^l}
= \textrm{X}_{n(i,:)}^l (\textrm{X}_{n(i,:)}^l)^\textrm{T} \otimes  \textrm{G}_{n(i,:)}^{\textrm{Z}_{n(i,:)}^l}(\textrm{G}_{n(i,:)}^{\textrm{Z}_{n(i,:)}^l})^\textrm{T}
\end{equation}
where $\otimes$ is the Kronecker product. To reduce the memory and computation burden,
we use the Kronecker-factored approximation \cite{23,24}
to rewrite (42) as
\begin{equation}
\textrm{H}^l_{t+1} \approx \textrm{H}^{(1)}_{t+1} \otimes \textrm{H}^{(2)}_{t+1}
\end{equation}
where $\textrm{H}^{(1)}_{t+1}$ and $\textrm{H}^{(2)}_{t+1}$  are defined as follows
\begin{equation}
\textrm{H}^{(1)}_{t+1} = \frac{k}{NT} \sum_{n=1}^{t}\lambda^{t-n}(\textrm{X}_{n}^l)^{\textrm{T}}\textrm{X}_{n}^l
\end{equation}
\begin{equation}
\textrm{H}^{(2)}_{t+1} = \frac{k}{NT}\sum_{n=1}^{t}\lambda^{t-n}(\textrm{G}_{n}^{\textrm{Z}_n^l})^{\textrm{T}}\textrm{G}_{n}^{\textrm{Z}_n^l}
\end{equation}
where $\textrm{G}_{n}^{\textrm{Z}_n^l}= [\textrm{G}_{n(1,:)}^{\textrm{Z}_{n(i,:)}^l},\cdots,\textrm{G}_{n(NT,:)}^{\textrm{Z}_{n(i,:)}^l}]^\textrm{T} \in \mathcal{R}^{NT\times N_l}$, and $k>0$ is another average scaling factor.
Let $\textrm{P}^{(1)}_{t}=\big(\textrm{H}^{(1)}_{t}\big)^{-1}$ and $\textrm{P}^{(2)}_{t}=\big(\textrm{H}^{2)}_{t}\big)^{-1}$.
Now, we rewrite (47) and (48) as the incremental forms,  use the Sherman-Morrison matrix inversion lemma for each sample
in the current mini batch, and average the recursive results. We can easily get
\begin{eqnarray}
\textrm{P}^{(1)}_{t+1} \approx \frac{1}{\lambda}\bigg(\textrm{P}^{(1)}_{t}-
\frac{k}{NT}\sum_{i=1}^{NT} \frac{ \textrm{P}^{(1)}_{t} \textrm{X}^l_{t(i,:)} (\textrm{X}^l_{t(i,:)})^{\textrm{T}}\textrm{P}^{(1)}_{t}}
{\lambda+k(\textrm{X}^l_{t(i,:)})^{\textrm{T}}\textrm{P}^{(1)}_{t}\textrm{X}^l_{t(i,:)} }\bigg)~
\\
\textrm{P}^{(2)}_{t+1} \approx \frac{1}{\lambda}\bigg(\textrm{P}^{(2)}_{t}-
\frac{k}{NT}\sum_{i=1}^{NT} \frac{ \textrm{P}^{(2)}_{t} \textrm{G}^{\textrm{Z}_{t(i,:)}^l}_{t(i,:)} (\textrm{G}^{\textrm{Z}_{t(i,:)}^l}_{t(i,:)})^{\textrm{T}}\textrm{P}^{(2)}_{t}}
{\lambda+k(\textrm{G}^{\textrm{Z}_{t(i,:)}^l}_{t(i,:)})^{\textrm{T}}\textrm{P}^{(2)}_{t}\textrm{G}^{\textrm{Z}_{t(i,:)}^l}_{t(i,:)} }\bigg)
\end{eqnarray}
Plugging (46) into (41) yields
\begin{equation}
w_{t+1}\approx \big(\textrm{P}^{(1)}_{t+1} \otimes \textrm{P}^{(2)}_{t+1}\big)b_{t+1}^l
=  \textrm{P}^{(1)}_{t+1} m(b_{t+1}^l) \textrm{P}^{(2)}_{t+1}
\end{equation}
where $m(b_{t+1}^l)$ denotes reshaping the vector $b_{t+1}^l$ into an $N_{l-1}\times N_l$ matrix.
Then, the rest deviation is similar to what we do in Section III. \textit{A}. Using (43), (49) and (50), we can obtain
\begin{equation}
w_{t+1} \approx w_{t}  - v\big(\textrm{P}^{(1)}_{t}m\big(\frac{1}{NT}\sum_{i=1}^{NT}
\textrm{G}_{t(i,:)}^w\big)\textrm{P}^{(2)}_{t}\big)
\end{equation}
where $\textrm{G}_{t(i,:)}^w$ is defined as
\begin{equation}
\textrm{G}_{t(i,:)}^w \!=\!\frac{\big(A(\textrm{S}_{t(i,:)},\textrm{A}_{t(i,:)})-w_t^\textrm{T} \textrm{G}_{t(i,:)}^{\Theta_t^l}\big)\textrm{G}_{t(i,:)}^{\Theta_t^l}}
{\!\big(\lambda\!+\!k\big(\textrm{X}^l_{t(i,:)}\big)^{\!\textrm{T}}\textrm{P}^{(1)}_{t}\!\textrm{X}^l_{t(i,:)}\!\big)
\!\big(\lambda\!+\!k\big(\textrm{G}^{\textrm{Z}_{t(i,:)}^l}_{t(i,:)}\big)^{\!\textrm{T}}\textrm{P}^{(2)}_{t}\!\textrm{G}^{\textrm{Z}_{t(i,:)}^l}_{t(i,:)} \!\big)} \nonumber
\end{equation}
Note that we don't use the average-approximation method used in Section III. \textit{A} to update $\textrm{P}^{l(1)}_{t}$, $\textrm{P}^{l(2)}_{t}$ and $w_t$, since $\textrm{G}^{\textrm{Z}_{t(i,:)}^l}_{t(i,:)}$ and $\textrm{G}_{t(i,:)}^{\Theta_t^l}$ of different samples
sometimes have large difference and this method will blur their difference.

Finally, the parameter of the actor output layer updated by using NPG can be defined as
\begin{equation}
\Theta^l_{t+1}=\Theta^l_{t} + \alpha m(w_{t+1})
\end{equation}
where $\alpha$ is the learning rate.

\subsection{Optimizing Fully-connected Hidden Layer}
In this subsection, we discuss the RLS optimization for fc hidden layers
in the critic and actor networks. Generally, there is a nonlinear activation function in each
hidden layer, which makes us difficult to derive the least squares solutions of $\Psi_{t+1}^l$ and $\Theta_{t+1}^l$  by using
the same method introduced in Section III. \textit{A}. In fact, this is the main reason why it is difficult to combine DAC and RLS.

Here we use the equivalent-gradient method, proposed in our previous work \cite{34},  to tackle this issue.
For the current layer $l$ of the critic network,  we define an auxiliary least squares loss function as
\begin{equation}
\tilde{L}(\Psi)=\frac{1}{2NT}\sum_{n=1}^{t}\lambda^{t-n}\big\|\textrm{Z}^{l*}_n - \textrm{Z}^{l}_n  \big\|^2_F
\end{equation}
where $\textrm{Z}^{l*}_n$ is the corresponding desired value of $\textrm{Z}^{l}_n=\textrm{X}^l_n\Psi^l$.
Then, by using the same derivation method in Section III. \textit{A}, $\Psi^l_{t+1}$ is defined as
\begin{equation}
\Psi^l_{t+1}\approx\Psi^l_{t}+ \frac{k \textrm{P}^l_{t} \bar{x}_t^l(\bar{z}^{l*}_t - \bar{z}^{l}_t)^{\textrm{T}} } {\lambda+k(\bar{x}^l_t)^{\textrm{T}}\textrm{P}^l_{t}\bar{x}^l_t}
\end{equation}
where $\bar{z}^{l*}_t$ and $\bar{z}^l_t$ are defined as follows
\begin{eqnarray}
\bar{z}^{l*}_t=\frac{1}{NT}\sum_{i=1}^{NT}\textrm{Z}^{l*}_{t(i,:)} \\
\bar{z}^l_t =\frac{1}{NT}\sum_{i=1}^{NT} \textrm{Z}^l_{t(i,:)} ~
\end{eqnarray}
and the update rule of  $\textrm{P}_t$ is the same as (30).
Furtherly, from our previous work $\cite{34}$,  $\nabla_{\textrm{Z}^l_{t}} L(\Psi_t,\Theta_t)$  can be equivalently defined as
\begin{equation}
\nabla_{\textrm{Z}^l_{t}} L(\Psi_t,\Theta_t) = - \frac{1}{\mu NT}(\textrm{Z}^{l*}_t - \textrm{Z}^{l}_t)
\end{equation}
where $\mu>0$ is the gradient scaling factor. Plugging (58) into (55), we finally get
\begin{equation}
\Psi^l_{t+1} \approx \Psi^l_{t}- \frac{\mu\textrm{P}^l_{t}}{\lambda+k(\bar{x}^l_t)^{\textrm{T}}\textrm{P}^l_{t}\bar{x}^l_t} \nabla_{\Psi^l_{t}} L(\Psi_t,\Theta_t)
\end{equation}

Note that the derivation of $\Theta^l_{t+1}$, which is the parameter of the current
fc hidden layer in the actor network, is the same as the above.
For brevity, we directly give the result, namely
\begin{equation}
\Theta^l_{t+1} \approx \Theta^l_{t}- \frac{\mu\textrm{P}^l_{t}}{\lambda+k(\bar{x}^l_t)^{\textrm{T}}\textrm{P}^l_{t}\bar{x}^l_t} \nabla_{\Theta^l_{t}} L(\Psi_t,\Theta_t)
\end{equation}
where the update rule of $\textrm{P}^l_t$ is also the same as (30).

\subsection{Optimizing Convolutional Hidden Layer}
Conv layers are at the front of a CNN. They are usually used to learn spatial features of original state inputs.
As defined by (8), a conv layer can be viewed as a special fc layer. That means
we can also use the same RLS update rules as (59) and (60) to optimize the conv layers of critic and actor networks.
However, there is a little different, since the current input $\hat{\textrm{X}}^l_t$ of the conv layer $l$ has three dimensions
rather than two dimensions. In our previous work \cite{34},
we define $\bar{x}^l_t=\frac{1}{NTH_lW_l}\sum_{i=1}^{NT}\sum_{j=1}^{H_l W_l}\hat{\textrm{X}}^l_{t(i,:,j)}$
to tackle this problem. But in practice,  we find that this definition will blur the difference among the input selections
of different output pixels, which will worsen the performance of our algorithms.

To avoid this situation, we define
\begin{equation}
\bar{\textrm{X}}^l_t=\frac{1}{NT}\sum_{i=1}^{NT}\hat{\textrm{X}}^l_{t(i,:,:)}
\end{equation}
where $\bar{\textrm{X}}^l_t \in \mathcal{R}^{C_{l-1} H_l^k W_l^k\times H_lW_l}$.
On the basis, similar to that
of $\textrm{P}^{l(1)}_{t}=\big(\textrm{H}^{l(1)}_{t}\big)^{-1}$ and $\textrm{P}^{l(2)}_{t}=\big(\textrm{H}^{l(2)}_{t}\big)^{-1}$ introduced
in Section  III. \textit{B},
the recursive derivation of $\textrm{P}^l_{t+1}=\big(\frac{1}{NT}\sum_{n=1}^{t}\lambda^{t-n}(\hat{\textrm{X}}^l_n)^{\textrm{T}}\hat{\textrm{X}}^l_n \big)^{-1} $ will yield
\begin{eqnarray}
\textrm{P}^{l}_{t+1}\!\approx\! \frac{1}{\lambda}\bigg(\textrm{P}^{l}_{t}\!-\!
\frac{k}{H_lW_l} \!\sum_{j=1}^{H_lW_l}\!\frac{\textrm{P}^{l}_{t} \bar{\textrm{X}}^l_{t(:,j)} (\bar{\textrm{X}}^l_{t(:,j)})^{\textrm{T}}\textrm{P}^{l}_{t}}
{\lambda+k(\bar{\textrm{X}}^l_{t(:,j)})^{\textrm{T}}\textrm{P}^{l}_{t}\bar{\textrm{X}}^l_{t(:,j)} }\bigg)
\end{eqnarray}
Finally, we can obtain the update rules for $\Psi_t^l$ and  $\Theta_t^l$ defined as follows
\begin{equation}
\Psi^l_{t+1} \approx \Psi^l_{t}- \tau\bigg(\frac{\mu H_lW_l\textrm{P}^l_{t}o\big(\nabla_{\Psi^l_{t}} L(\Psi_t,\Theta_t)\big)}{\!\sum_{j=1}^{H_lW_l}\!\big(\lambda+k(\bar{\textrm{X}}^l_{t(:,j)})^{\textrm{T}}\textrm{P}^{l}_{t}\bar{\textrm{X}}^l_{t(:,j)}\big)}\bigg)
\end{equation}
\begin{equation}
\Theta^l_{t+1} \approx \Theta^l_{t}- \tau\bigg(\frac{\mu H_lW_l\textrm{P}^l_{t}o\big(\nabla_{\Theta^l_{t}} L(\Psi_t,\Theta_t)\big)}{\!\sum_{j=1}^{H_lW_l}\!\big(\lambda+k(\bar{\textrm{X}}^l_{t(:,j)})^{\textrm{T}}\textrm{P}^{l}_{t}\bar{\textrm{X}}^l_{t(:,j)}\big)}\bigg)
\end{equation}
where  $o(\cdot)$ denotes reshaping a $C_{l-1}\times C_l \times H_l^k \times W_l^k$ tensor into a  $C_{l-1} H_l^k W_l^k\times C_l$ matrix, and
$\tau(\cdot)$ does the reverse.

\subsection{RLSSA2C and RLSNA2C Algorithms}
\begin{algorithm}[htb]
\caption{RLS-Based Advantage Actor Critic}
\textbf{Input:} critic parameters $\{\Psi_0^l, \textrm{P}_0^l\}_{l=1}^{L_{\textrm{C}}}$; actor parameters
$\{\Theta_0^l\}_{l=1}^{L_{\textrm{A}}}$, $\{\textrm{P}_0^l\}_{l=1}^{L_{\textrm{A}}\!-\!1}$, hyperparameters of the ordinary first-order algorithm or
$\{w_0,\textrm{P}_0^{(1)},\textrm{P}_0^{(2)},\alpha\}$;
initial states $\!\{s_{0,i}^{(1)} \}_{i=1}^N$ of $N$ workers, discount factor $\gamma$, scaling factors $k$ and $\mu$,
forgetting factor $\lambda$, regularization factor $\eta$.

\For{$t=0,1,2, ... $}
{
    \textbf{Excute}:  let each worker run $T$ timesteps and generate
    $\mathcal{M}_t=\{(s_{t,i}^{(k)},a_{t,i}^{(k)},{s'}_{t,i}^{(k)},r_{t,i}^{(k)},d_{t,i}^{(k)})\}_{i=1,\cdots,N}^{k=1,\cdots,T}$,
    where $a_{t,i}^{(k)} \sim \pi(a_{t,i}^{(k)}|s_{t,i}^{(k)},\Theta_t)$  decided by the actor network

    \textbf{Measure}: calculate the loss function by (10)

    \Critic{
         update $\Psi_t^{l_C}, \textrm{P}_t^{l_C}$ in fc output layer by (35), (30) \\
         update $\Psi_t^l, \textrm{P}_t^l$ in each fc hidden layer by \!(59), \!(30)\\
         update $\Psi_t^l, \textrm{P}_t^l$ in each conv layer by (63), (62)\\
    }
    \Actor{
         update  $\Theta_t^{L_{\!A}}$ by an ordinary first-order algorithm or
         $w_t,\textrm{P}_t^{l(1)},\textrm{P}_t^{l(2)},\Theta_t^{L_{\!A}}$\!  by (52), (49), (50), (53)\\
         update $\Theta_t^l, \textrm{P}_t^l$ in each fc hidden layer by \!(60), \!(30)\\
         update $\Theta_t^l, \textrm{P}_t^l$ in each conv layer by (64), (62)\\
    }
   \textbf{Set} $\{s_{t+1,i}^{(1)}\}_{i=1}^N = \{{s'}_{t,i}^{(T)}\}_{i=1}^N$  and  \textbf{discard} $\mathcal{M}_t$
}
\end{algorithm}

Based on the above derivation and the vanilla A2C, both RLSSA2C and RLSNA2C can be summarized in Algorithm 1, where $L_{\textrm{C}}$
and $L_{\textrm{A}}$ denote the total layer numbers in critic and actor networks. Note that the autocorrelation matrix  $\textrm{P}_t^{l}$ in the critic network is different from that in the actor network and we use the same notation only for brevity. For RLSSA2C,  $\Theta_t^{L_{\textrm{A}}}$ is updated by using SPG and an ordinary first-order optimization algorithm.
For RLSNA2C, $\Theta_t^{L_{\textrm{A}}}$ is updated by using the compatible parameter $w_{t+1}$, which is updated
by $\textrm{P}_t^{(1)}$ and $\textrm{P}_t^{(2)}$.  Except for those, the rest of RLSSA2C and RLSNA2C is the same.
In practice, to avoid instability in training, the critic network and the actor network sometimes are joint \cite{15,19}. In this case,
the shared layers are updated by the RLS optimization only once at each iteration.

\section{Analysis and Improvement}
In this section, we analyze the computational complexity and the convergence of RLSSA2C and RLSNA2C in brief. In addition, we also present three practical tricks for further improving their convergence speed.

\subsection{Theoretical Analysis}
First, we analyze the computational complexity of our proposed algorithms.  Although their derivation seems complex and tedious,
but in fact they are very simple and easy to implement. From (35), (52), (59), (60), (63) and (64), the RLS optimization for
actor and critic networks can be viewed as a special SGD optimization. For an fc hidden layer,  a critic fc output layer, an
actor fc output layer in RLSNA2C and a conv layer, the computational complexities of the RLS optimization are $1+\frac{N_{l-1}}{NT}$,  $1+\frac{N_{L-1}}{NT}$,  $2+\frac{N_{L-1}N_{L}}{NT}$ and $1+\frac{C_{l-1}H_{l}^kW_{l}^k}{NTH_lW_l}$
times as those of SGD, respectively. In practice, A2C generally uses 16 or 32 workers to interact with their environments for 5$\sim$20
timesteps at each iteration, that is, $NT$ is 80$\sim$640. Thus, our RLS optimization is only several times as complex as SGD.
In Section \uppercase\expandafter{\romannumeral5}, experimental results show that the running speeds of RLSSA2C and RLSNA2C
are only 10\%$\sim$30\% slower than that of the vanilla A2C with RMSProp, but they are significantly faster than those of PPO and ACKTR.

Next, we analyze the convergence of our proposed algorithms. As shown in Algorithm 1, RLSSA2C and RLSNA2C are two-time-scale actor-critic algorithms.
In LFARL, the convergence of this type of algorithm has been established \cite{7,32}. Namely, if the actor learning rate $\alpha_t^{\textrm{A}}$ and
the critic learning rate $\alpha_t^{\textrm{C}}$ satisfy the standard Robbins-Monro condition \cite{39} and  a time-scale seperation condition
$\textrm{lim}_{t\rightarrow\infty} \alpha_t^{\textrm{A}}/\alpha_t^{\textrm{C}}=0$,  actor and critic parameters will converge to
asymptotic stable equilibriums. However, unlike traditional actor-critic algorithms, DAC algorithms use non-linear function approximations.
Thus, their convergence proofs are difficult. In recent years, there have been a few studies on this issue.
Yang et al. establish the convergence of batch actor-critic with nonlinear function approximations and finite samples \cite{40}.
Liu et al. establish nonasymptotic upper bounds of the numbers of TD and SGD iterations, and prove that a variant
of PPO and TRPO with overparametrized neural networks converges to the globally optimal policy
at a sublinear rate \cite{41}. Assuming independent sampling, wang et al. prove that neural NPG converges to a globally optimal
policy at a sublinear rate \cite{42}. Compared with the proof methods for traditional actor-critic algorithms, these methods
have more assumptions and are more complex. Our algorithms are similar to the batch actor-critic algorithm studied in \cite{40},
and the RLS optimization can be viewed as a special SGD optimization. Therefore, we should establish the convergence of
RLSSA2C and RLSNA2C by using the methods in \cite{40}, \cite{41} and \cite{40}, \cite{42}, respectively.
Intuitively, if actor and critic networks can be mapped into two linear function approximator with some error, the
convergence proofs of RLSSA2C and RLSNA2C will be converted into the proofs of traditional actor-critic algorithms.

\subsection{Performance Improvement}
In Section III, to simplify the calculation, we import two scaling factors $k$ and $\mu$. From (28), (29), (47), (48) and (58), they
should be time-variant, so we present two new definitions for them.
From (30), we can get
\begin{eqnarray}
\textrm{P}^l_{t+1}\approx\frac{1}{\lambda}\bigg(\textrm{P}^l_{t}-\frac{
 \textrm{P}^l_{t}\bar{x}^l_t(\bar{x}^l_t)^{\textrm{T}}\textrm{P}^l_{t}}{\frac{\lambda}{k}+(\bar{x}^l_t)^{\textrm{T}}\textrm{P}^l_{t}\bar{x}^l_t}\bigg)
\end{eqnarray}
It is clear that the average scaling factor $k$ also plays a role similar to the forgetting factor $\lambda$. That is, a big $k$ will increase the forgetting rate,
and vice versa. At the beginning of learning, the policy $\pi$ is infantile and unstable, so we should select a big $k$ to forget the
historical information and emphasize the new samples for accelerating convergence. At the end of learning, the policy $\pi$ is close to the optimal,
so we should select a small $k$ to stabilize it. In (49), (50) and (62), $k$ also play the same role.
Thus, we redefine $k$ as
\begin{equation}
k_t= \max\big[k_0-\lfloor\frac{t}{t_\Delta}\rfloor \delta_k, k_{min}\big]
\end{equation}
where $k_0$, $t_\Delta$, $\delta_k$ and $k_{min}$ denote the initial value, the update interval, the decay size and the lower bound of $k$, respectively.
From (59), (60), (63) and (64), the gradient scaling factor $\mu$ is a part of the learning rate. A big $\mu$ can
accelerate convergence, but it will also cause $\pi$ to fall into the local optimum. Thus, $\mu$ should gradually decay to a steady value. Here, we redefine it as
\begin{equation}
\mu_t= \max\big[\mu_0-\lfloor\frac{t}{t_\Delta}\rfloor \delta_\mu, \mu_{min}\big]
\end{equation}
where $\mu_0$, $\delta_\mu$ and $\mu_{min}$ denote the initial value, the decay size and the lower bound of $\mu$, respectively.
In fact, $\mu_t$ is different for each layer. In a DNN,  deeper layers are more likely to suffer from the gradient vanishing,
so we suggest that users choose a big $\mu_0$ for these layers.

In addition, it is well known that the vanilla SGD can be accelerated by the momentum method \cite{43}.
Our RLS optimization is a special SGD, so we can also use this method to accelerate it.
Similar to in our previous work \cite{34}, (35) can be redefined as
\begin{eqnarray}
\Phi^l_{t+1} \approx \beta \Phi^l_{t}-\frac{\textrm{P}^l_{t} \nabla_{\Psi^l_{t}} L(\Psi_t,\Theta_t)}{\lambda+k(\bar{\textrm{x}}^l_t)^{\textrm{T}}\textrm{P}^l_{t}\bar{\textrm{x}}^l_t}\\
\Psi^l_{t+1} \approx \Psi^l_{t}+ \Phi^l_{t+1}  ~~~~~~~~
\end{eqnarray}
where $\Phi^l_{t}$ denote the velocity matrix in the $l^{th}$ layer, and $\beta$ is the momentum factor.
(59), (60), (63) and (64) can also be redefined as the similar forms. Note that we only
suggest RLSSA2C to use this method, since we find that it will worsen the RLSNA2C's stabilities empirically .

\section{Experimental Results}
In this section, we will compare our algorithms against the vanilla A2C with RMSProp (called RMSA2C) for evaluating the sample efficiency,
and  compare them against RMSA2C, PPO and ACKTR for evaluating the computational efficiency.
We first test these algorithms on 40 discrete control games in the Atari 2600 environment, and then test them on 11 continuous
control tasks in the MuJoCo environment.

\subsection{Discrete Control Evaluation}
Atari 2600 is the most famous discrete control benchmark platform for evaluating DRL.
It is comprised of a lots of highly diverse games, which have high-dimensional observations with raw pixels.
In this set of experiments, we select 40 games from the Atari environment for performance evaluation.
For each game, the state is a $4\!\times84\!\times\!84$ normalized image.

All tested algorithms have the same model architecture, which is defined in \cite{15}.
It has 3 shared conv  layers, 1 shared fc hidden layer, 1 separate fc critic output layer and 1
separate fc actor output layer.
The first conv layer has 32 $8\!\times\!8$ kernels with stride $4$,
the second conv layer has 64 $4\!\times\!4$ kernels with stride $2$,
the third conv layer has 32 $3\!\times\!3$ kernels with stride $1$,
and the fc hidden layer has $512$ neurons.
All hidden layers use ReLU activation functions,
the fc critic output layer uses the Identity activation function to predict the state-value function,
and the fc actor output layer uses the Softmax activation function to represent the policy.
The parameter of each layer is initialized  with the default settings of PyTorch.
All algorithms use the same loss function defined by (10), where the discount factor $\gamma=0.99$ and the entropy
regularization factor $\eta=0.01$.
At each iteration step, each algorithm lets 32 parallel workers run 5 timesteps, and uses the generated minibatch
including 160 samples for training. All workers  use an Intel Core i7-9700K CPU for trajectory sampling, and use a Nvidia RTX 2060 GPU
for accelerating the optimization computation.
To avoid gradient exploding, all parameter gradients are clipped by the $L_2$ norm with 0.5.

Besides the above common settings, individual settings of each algorithm are summarized as follows:
1) For RMSA2C, the learning rate $\epsilon$, decay factor $\rho$ and  small constant $\delta$ in RMSProp are set to 0.00025, 0.99 and 0.00005,  respectively.
2) For RLSSA2C,  all initial autocorrelation matrices are set to Identity matrices,
the forgetting factor $\lambda$ and  momentum factor $\beta$ are set to $1$ and $0.5$,
the hyperparameters $k_0$, $\delta_k$, $k_{min}$, $\mu_0$, $\delta_{\mu}$, $\mu_{min}$, $t_\triangle$ of scaling factors $k_t$ and $\mu_t$ are set to
0.1, 0.02 ,0.01, 5, 0.1, 1 and 5000,
and the actor output layer uses the same RMSProp as that in RMSA2C.
Note that $\mu_t$ is only used for conv layers but is fixed to 1 for all fc layers.
3) For RLSNA2C, the momentum factor $\beta$ is set to  $0.0$,
$\textrm{P}_0^{(1)}$ and $\textrm{P}_0^{(2)}$ are also set to Identity matrices,
and the learning rate $\alpha$ of the actor output layer is initialized to 0.01 and
decays by 0.002 per 5000 timesteps to 0.001. The other settings are the same as those  in RLSSA2C.
4) For PPO and ACKTR, we directly use Kostrikov's source code and settings  \cite{25}.
Note that the settings of RMSA2C are selected from \cite{15}, and the settings of RLSSA2C and RLSNA2C
are obtained by tuning Pong, Breakout and StarGunner games.

The convergence comparison of our algorithms against RMSA2C on 40 Atari games trained for 10 million timesteps is shown in Fig. 1.
It is clear that RLSSA2C and RLSNA2C outperform RMSA2C on most games.
Among these three algorithms, RLSSA2C has the best convergence performance (i.e., sample efficiency) and stability.
Compared with RMSA2C, RLSSA2C wins on 30 games. On Alien, Amidar, Assault, Asterix, BattleZone, Boxing, Breakout, Kangaroo, Krull,
KungFuMaster, MsPacman, Pitfall, Pong, Qbert, Seaquest and Tennis games, RLSSA2C is significantly superior to RMSA2C
in terms of convergence speed and convergence quality. Compared with RMSA2C, RLSNA2C also wins on 30 games.
On Asterix, Atlantis, DoubleDunk, FishingDerby, NameThisGame, Pong, Riverraid, Seaquest, UpNDown and Zaxxon games,
RLSNA2C performs very well. But compared with RLSSA2C, it is not very stable on some games.

In Table \uppercase\expandafter{\romannumeral1}, we present the last 100 average episode rewards of our algorithms and RMSA2C on 40 Atari games
trained for 10 million timesteps. From Table I, RLSSA2C and RLSNA2C obtain the highest rewards on 19 and 15 games respectively,
but RMSA2C only wins on 6 games. Notably, on Assault, BattleZone, Gopher and Q-bert games, RLSSA2C respectively acquires 2.6, 2.4, 3.6, and 3.1 times scores as much as RMSA2C. On Asterix, Atlantis, FishingDerby, UpNDown and Zaxxon games, RLSNA2C respectively acquired 2.8, 2.5, 2.7, 4.8, and 45 times scores as much as RMSA2C.

The running speed comparison of our algorithms against RMSA2C, PPO and ACKTR on six Atari games is shown in Table \uppercase\expandafter{\romannumeral2}.
Among these five algorithms, RMSA2C has the highest computational efficiency, RLSNA2C and RLSSA2C are listed 2nd and 3rd respectively, and PPO is listed last. In detail, RLSSA2C is only 28.1\% slower than RMSA2C, but is  592.7\% and 31.3\% faster than PPO and ACKTR.  RLSNA2C is only 27.7\% slower than RMSA2C, but is  596.3\% and 31.9\% faster than PPO and ACKTR. By using Kostrikov's source code to test PPO and ACKTR, we find that our algorithms have the
similar performance. Therefore, our algorithms achieve a better trade-off between high convergence performance and low computational cost.

\begin{figure*}
\centering
\subfigure{\includegraphics[width=4.2cm]{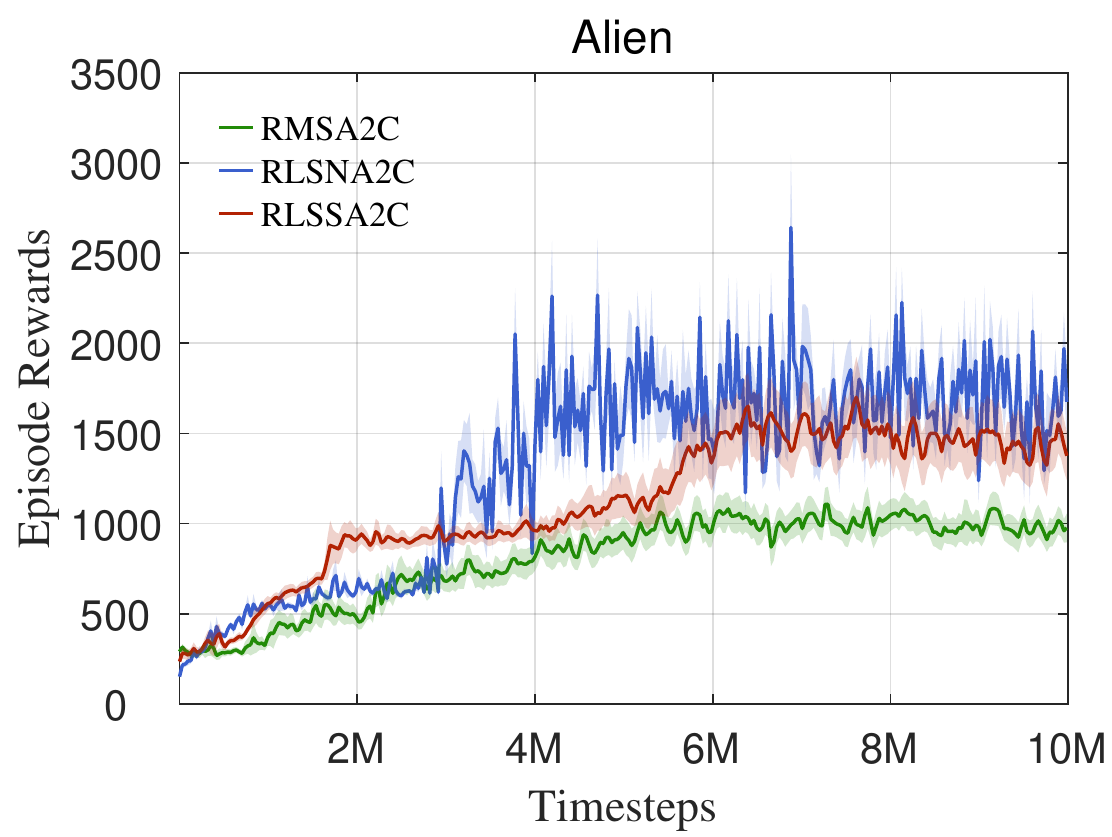}}
\subfigure{\includegraphics[width=4.2cm]{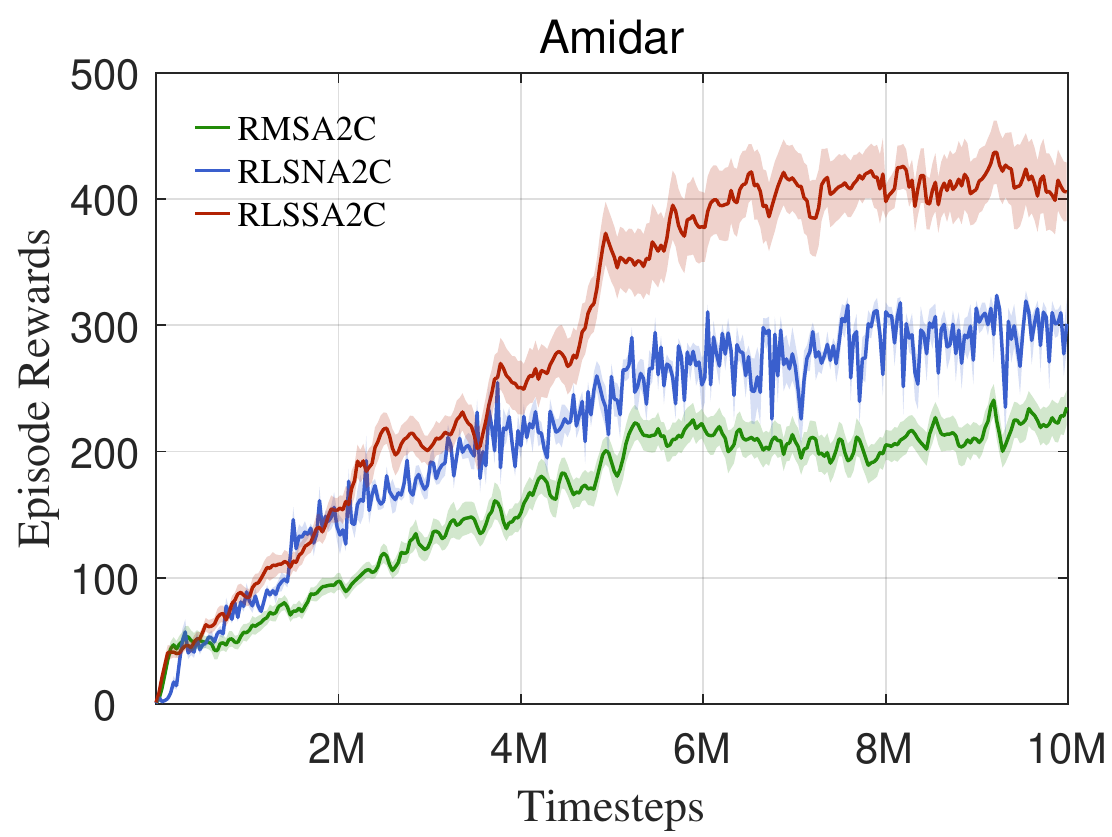}}
\subfigure{\includegraphics[width=4.2cm]{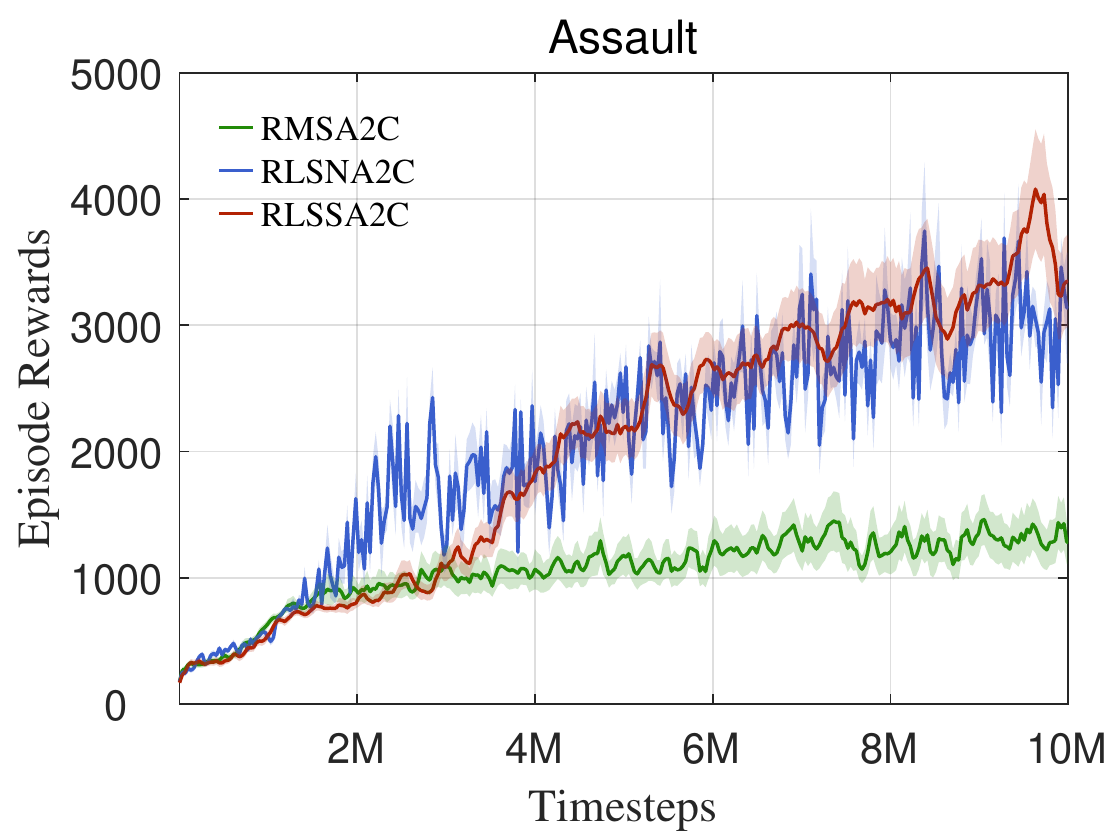}}
\subfigure{\includegraphics[width=4.2cm]{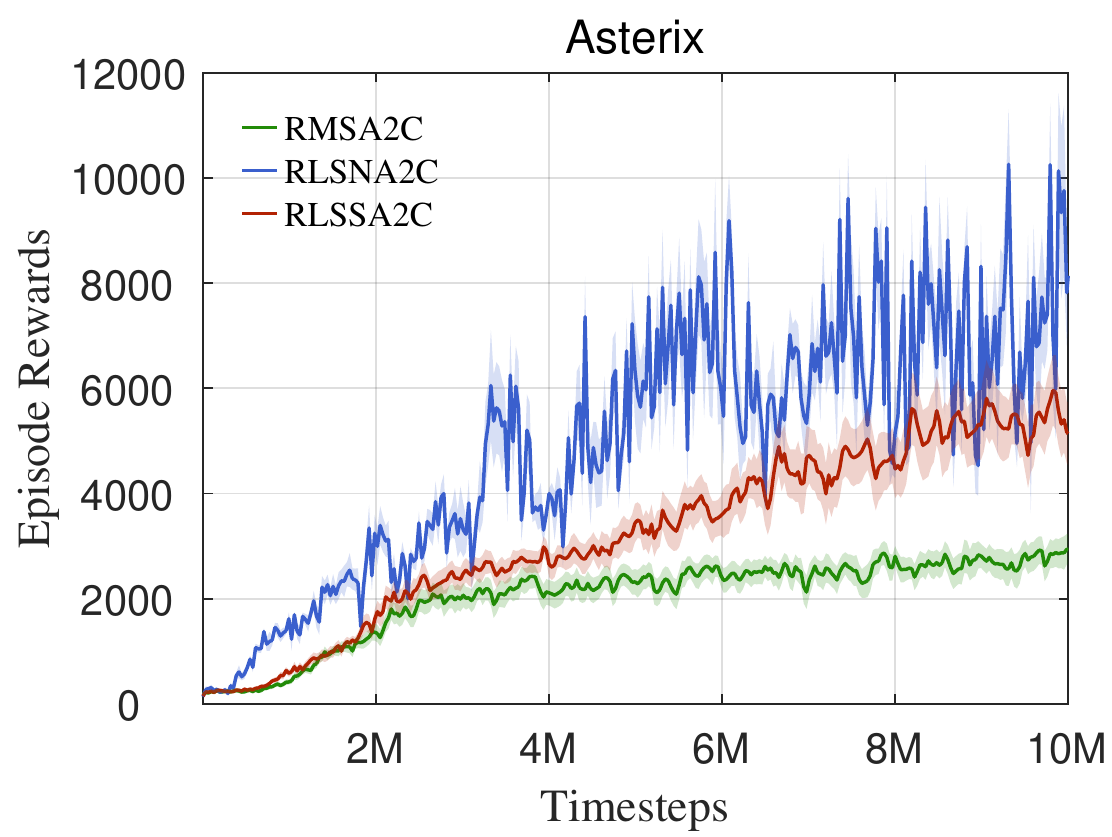}}
\centering
\subfigure{\includegraphics[width=4.2cm]{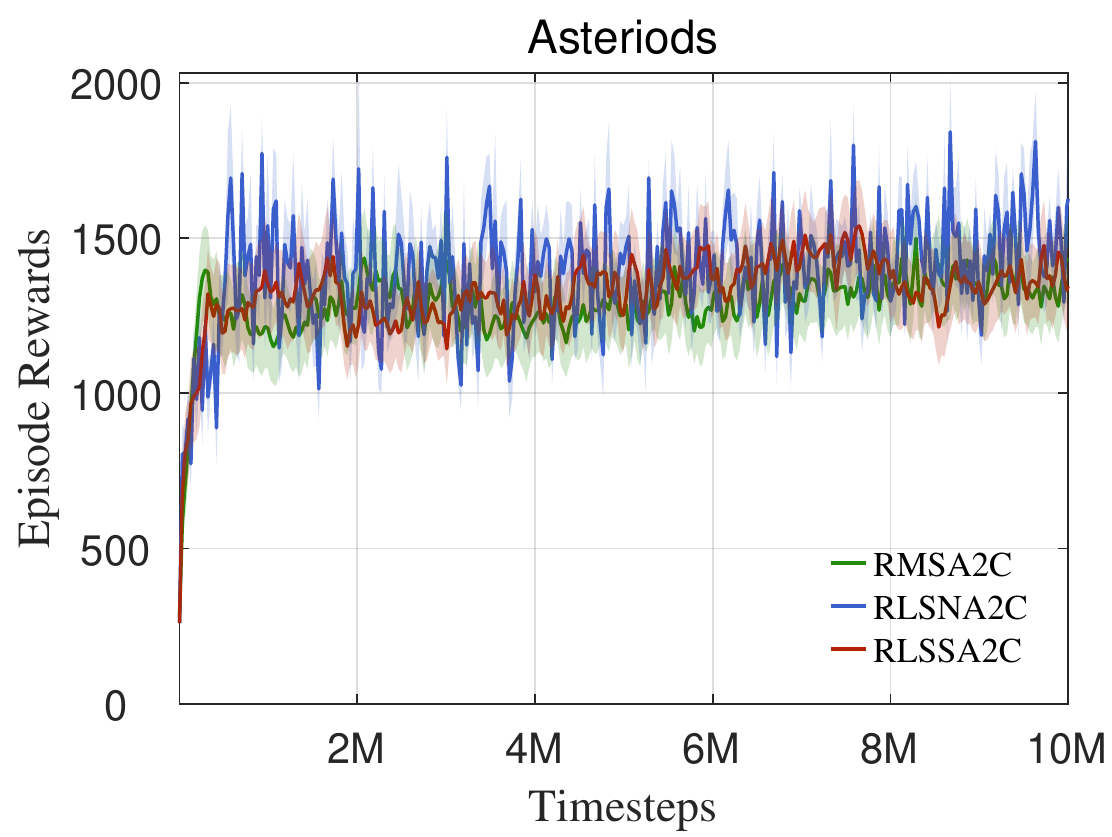}}
\subfigure{\includegraphics[width=4.2cm]{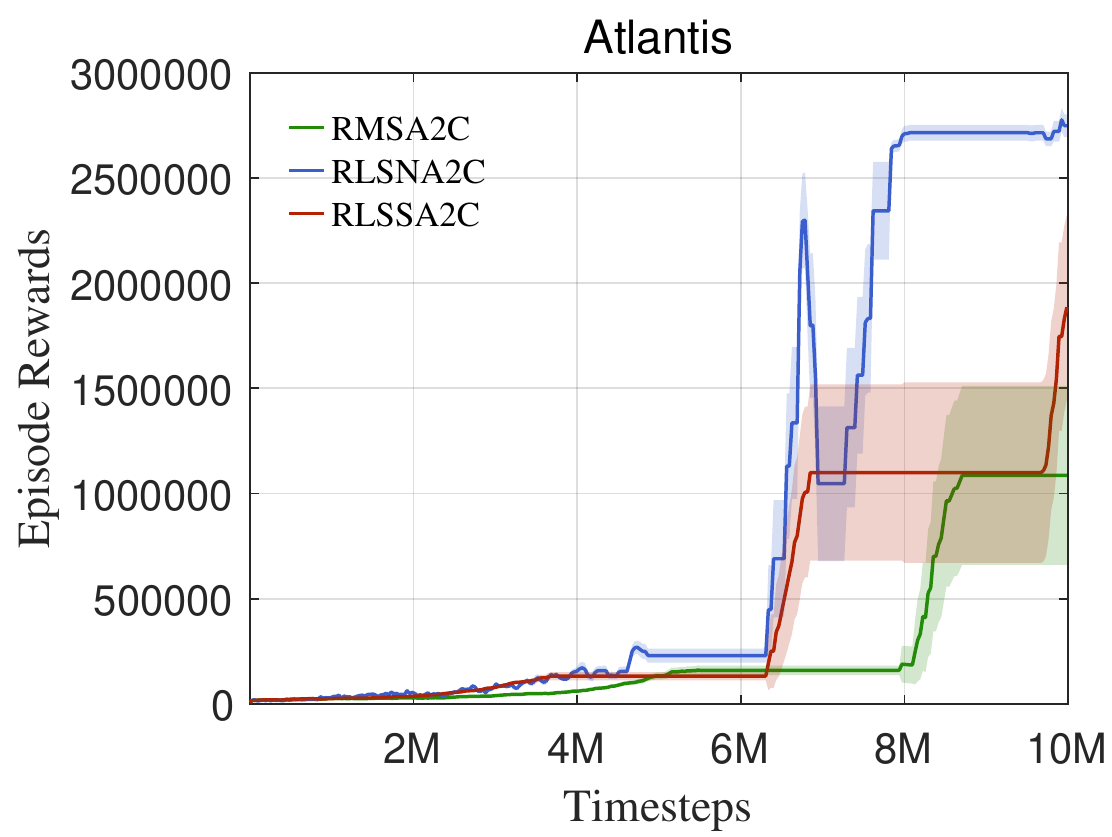}}
\subfigure{\includegraphics[width=4.2cm]{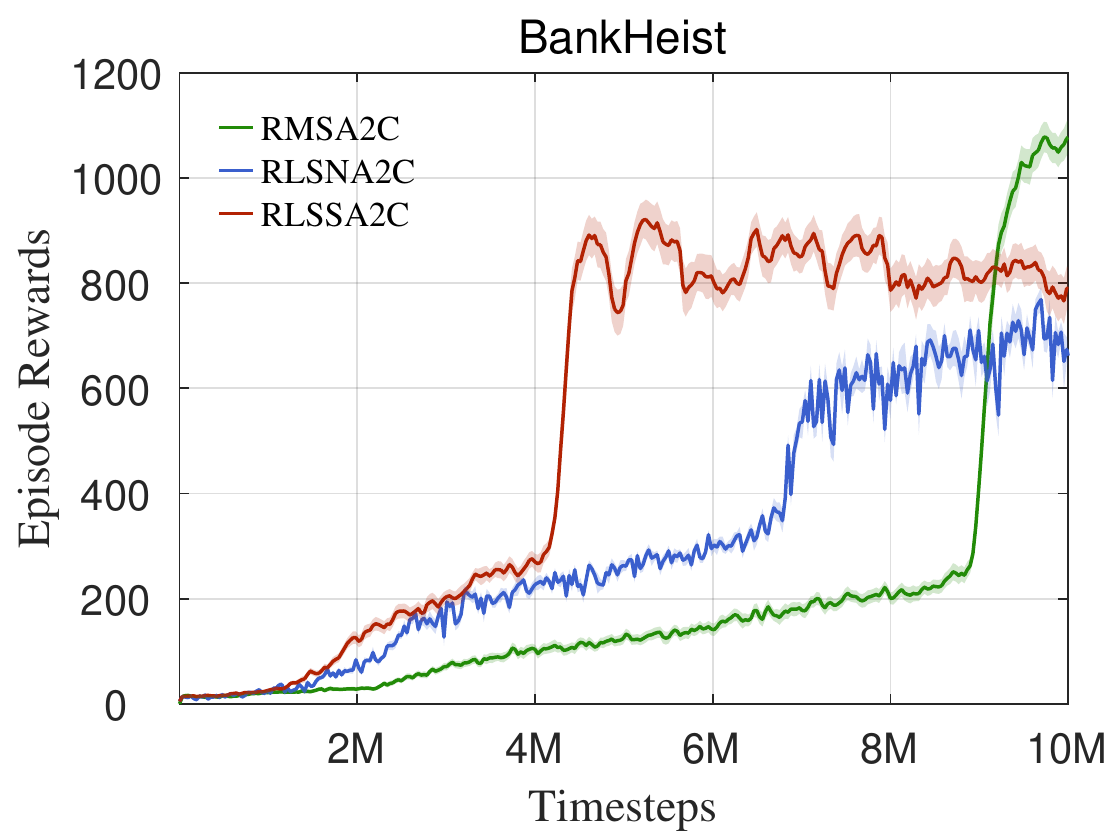}}
\subfigure{\includegraphics[width=4.2cm]{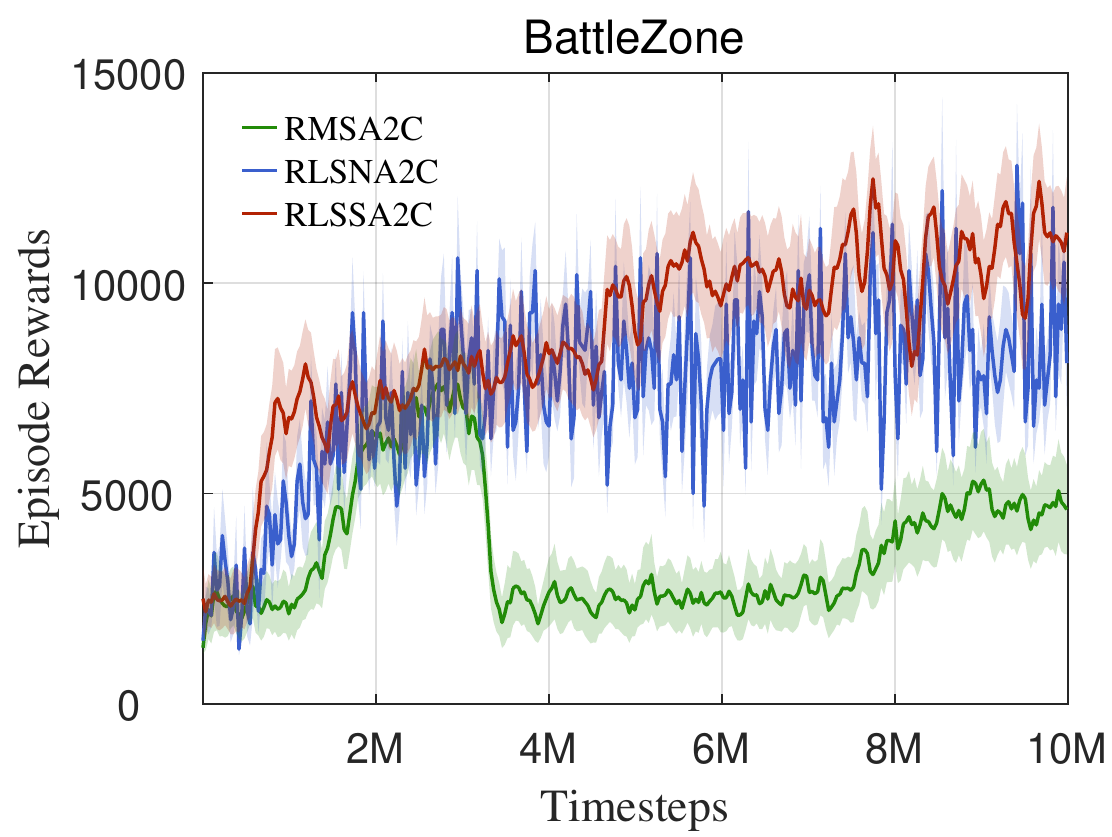}}
\centering
\subfigure{\includegraphics[width=4.2cm]{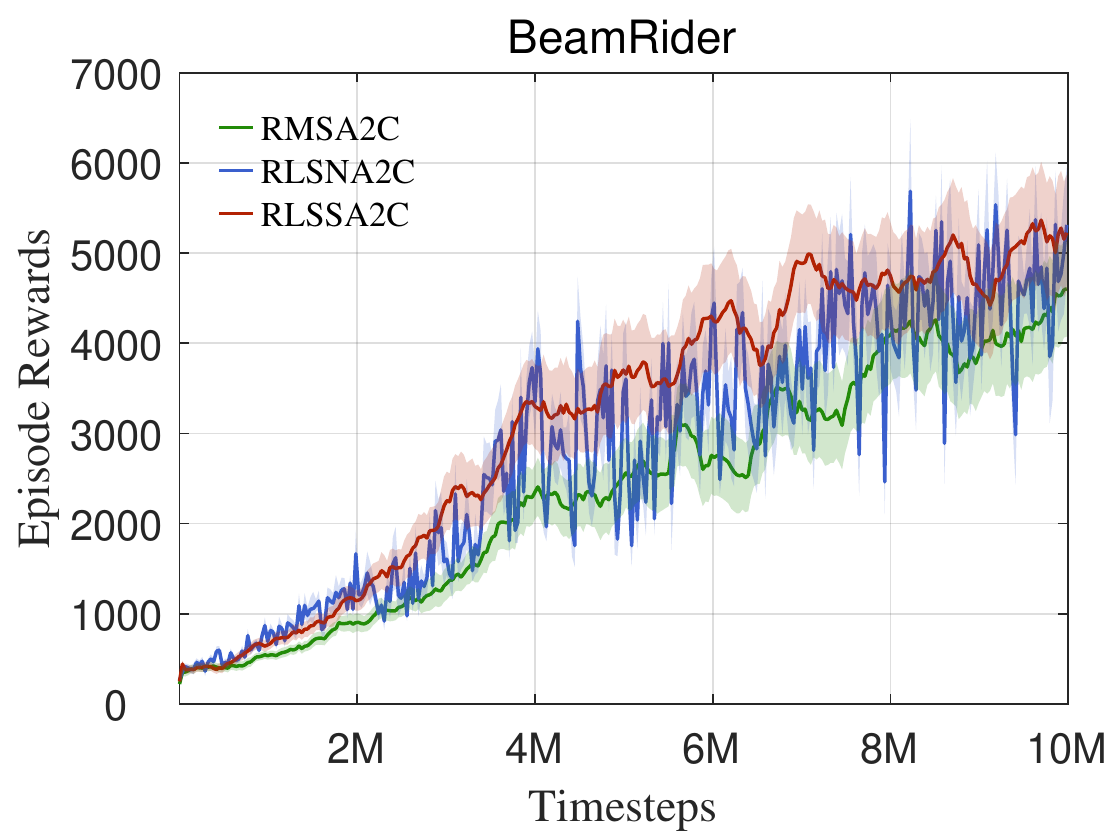}}
\subfigure{\includegraphics[width=4.2cm]{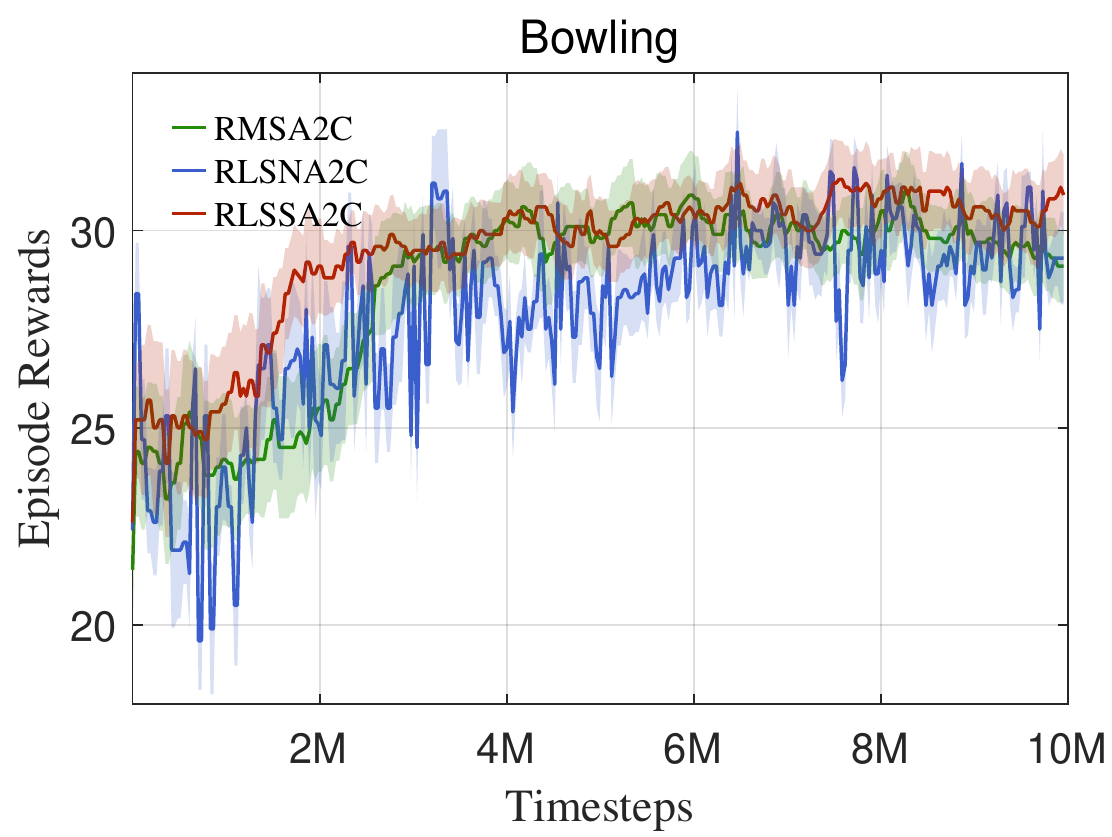}}
\subfigure{\includegraphics[width=4.2cm]{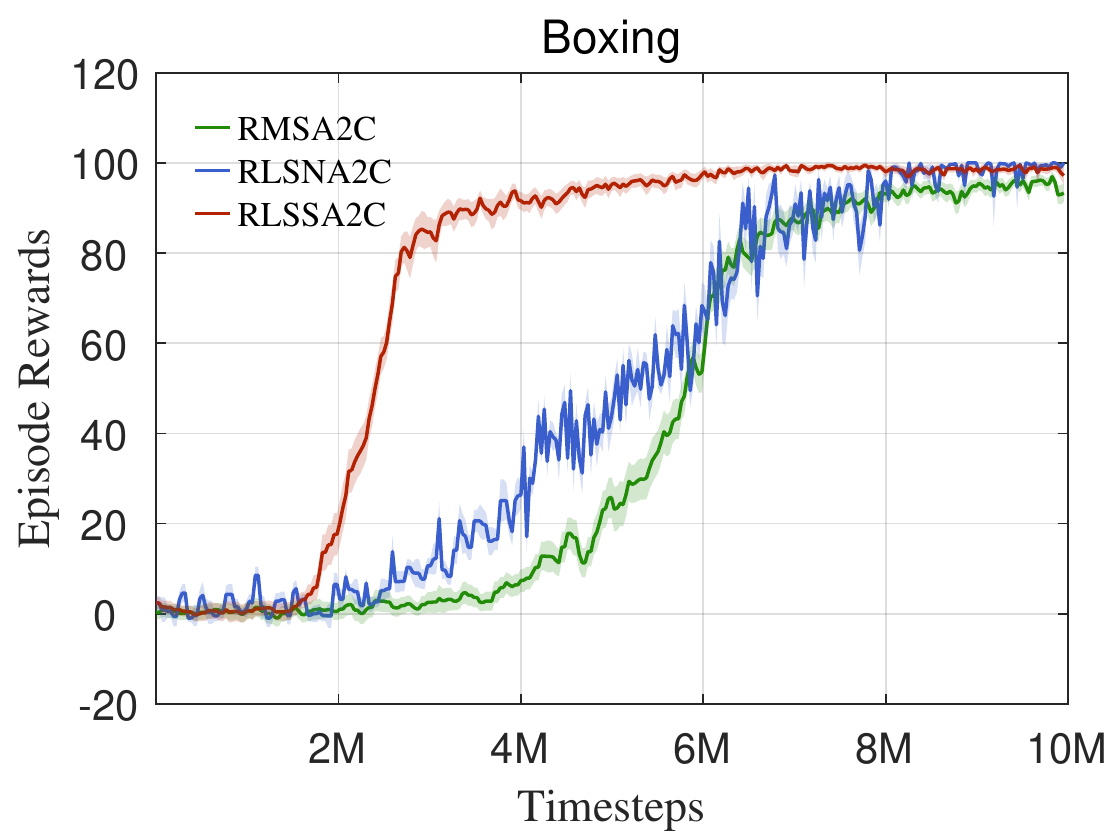}}
\subfigure{\includegraphics[width=4.2cm]{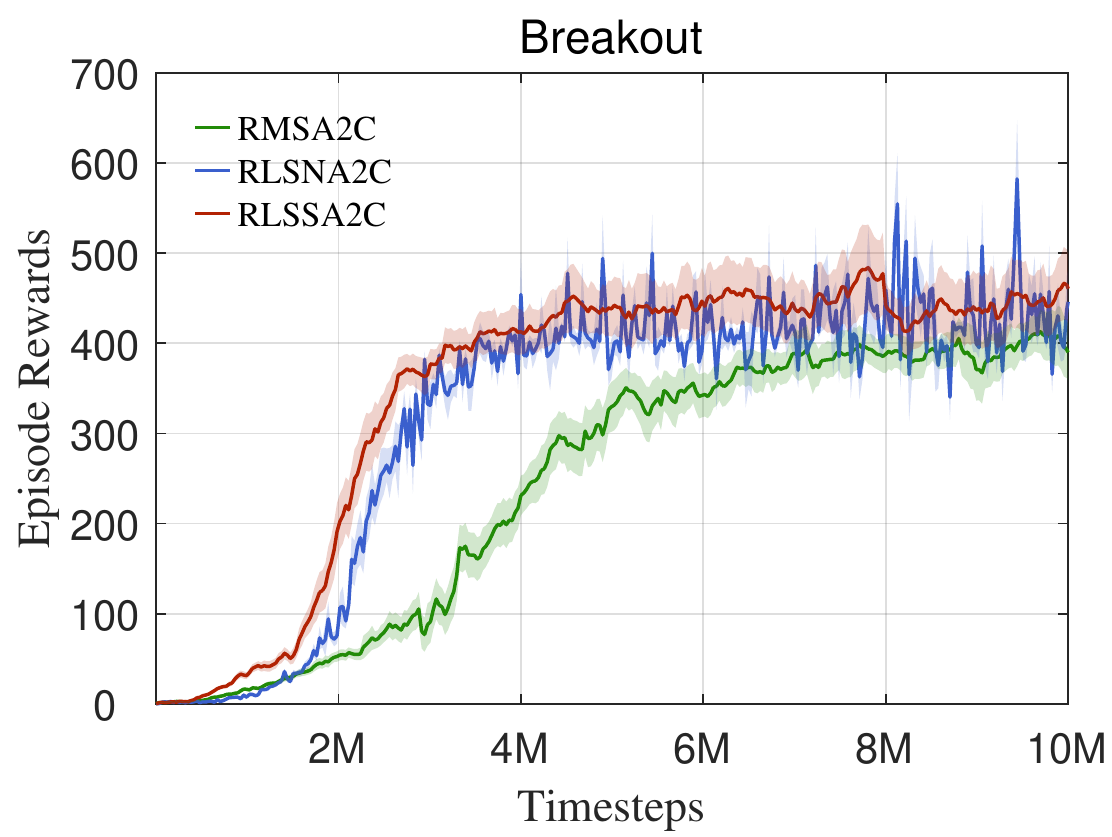}}
\centering
\subfigure{\includegraphics[width=4.2cm]{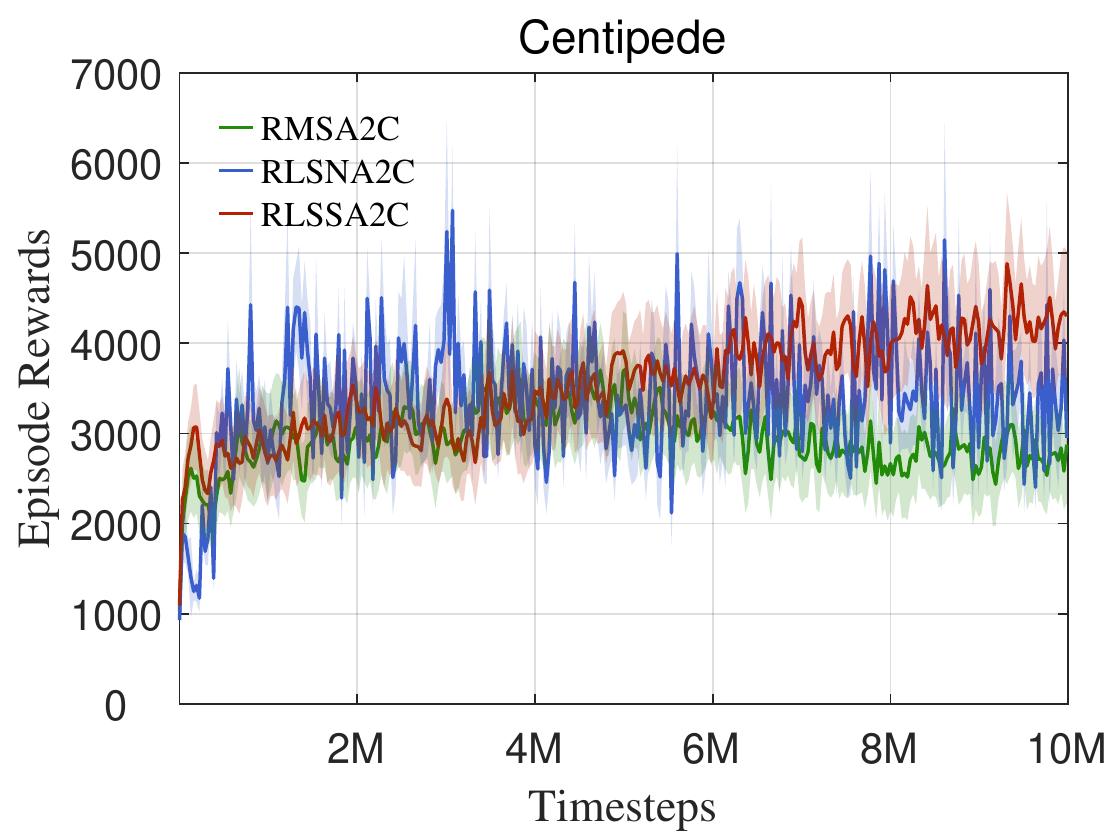}}
\subfigure{\includegraphics[width=4.2cm]{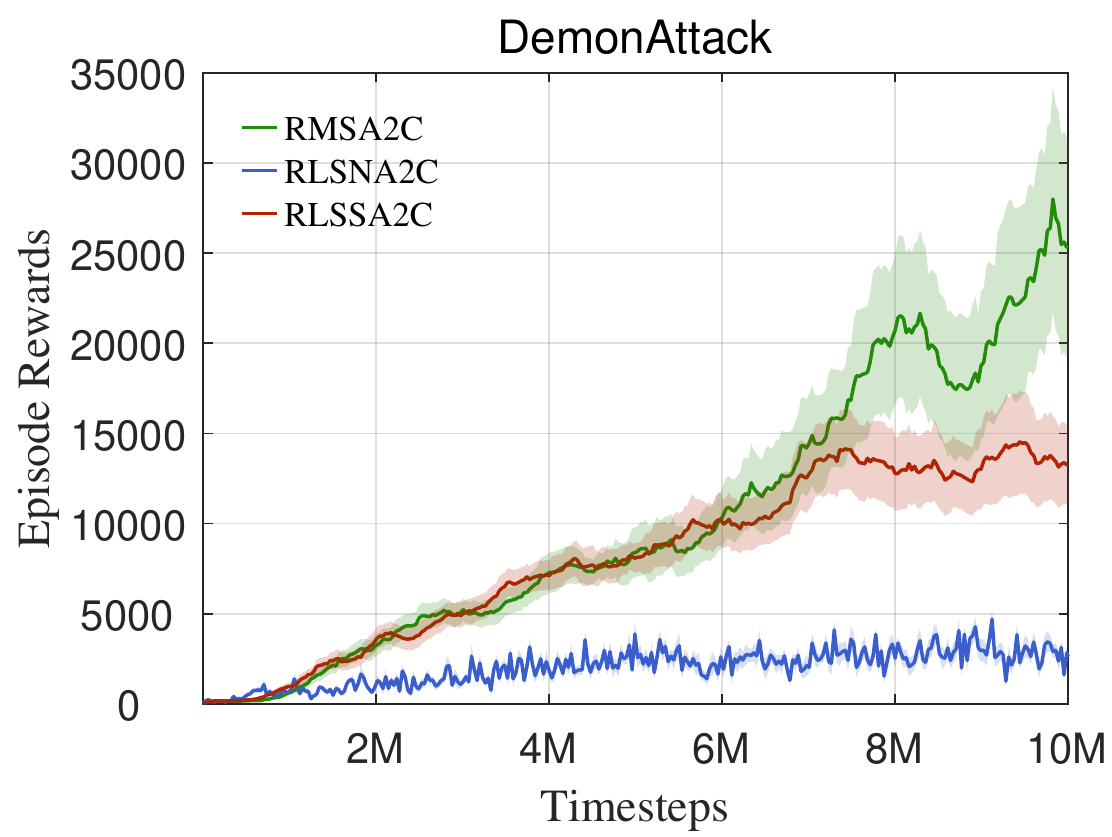}}
\subfigure{\includegraphics[width=4.2cm]{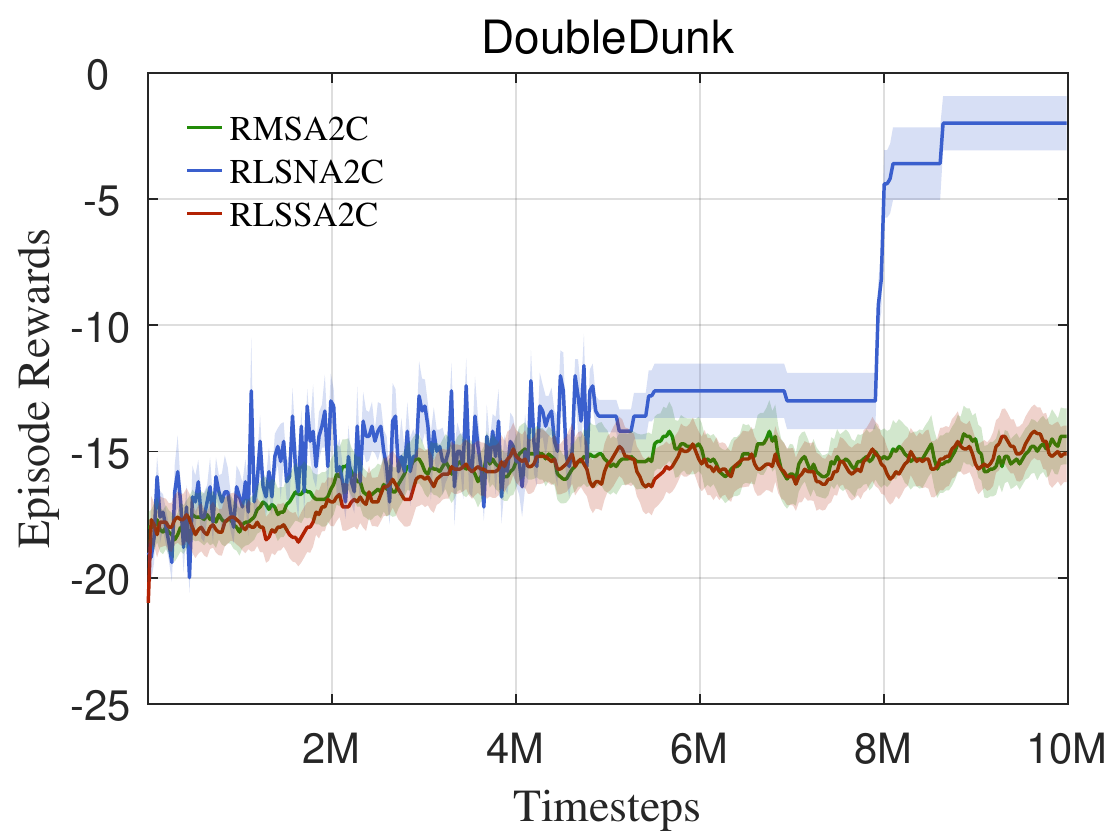}}
\subfigure{\includegraphics[width=4.2cm]{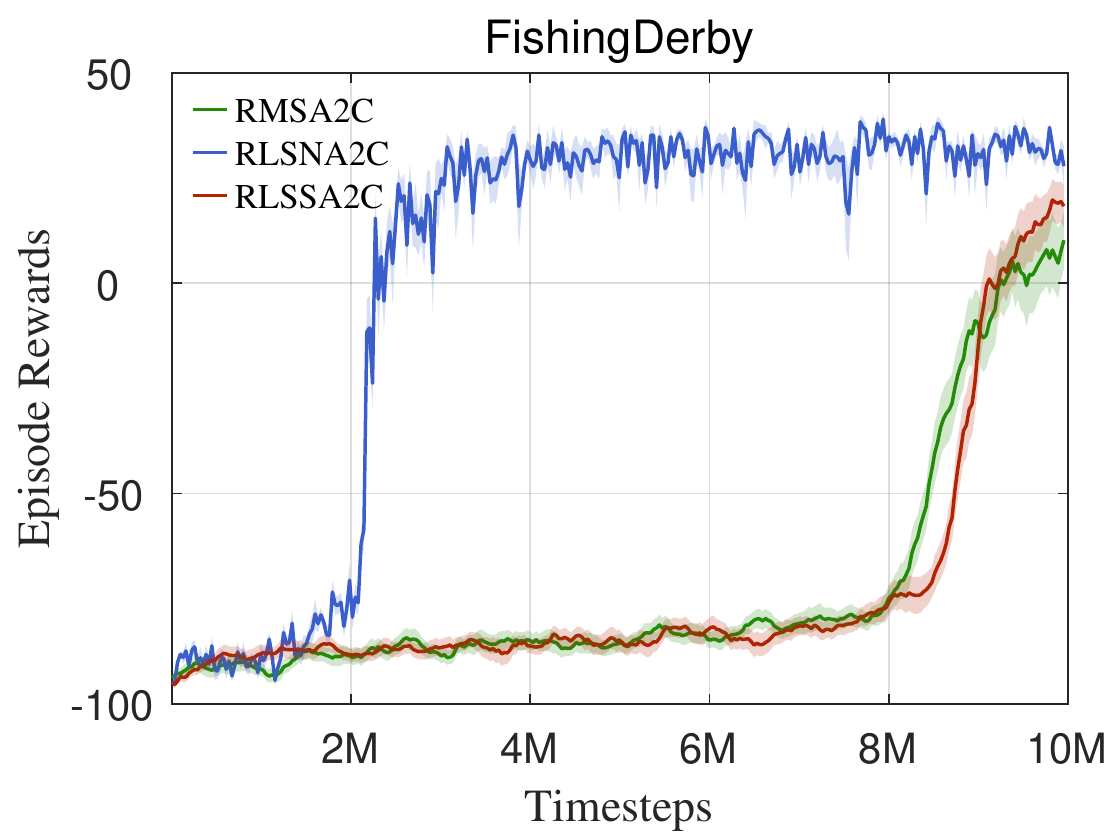}}
\centering
\subfigure{\includegraphics[width=4.2cm]{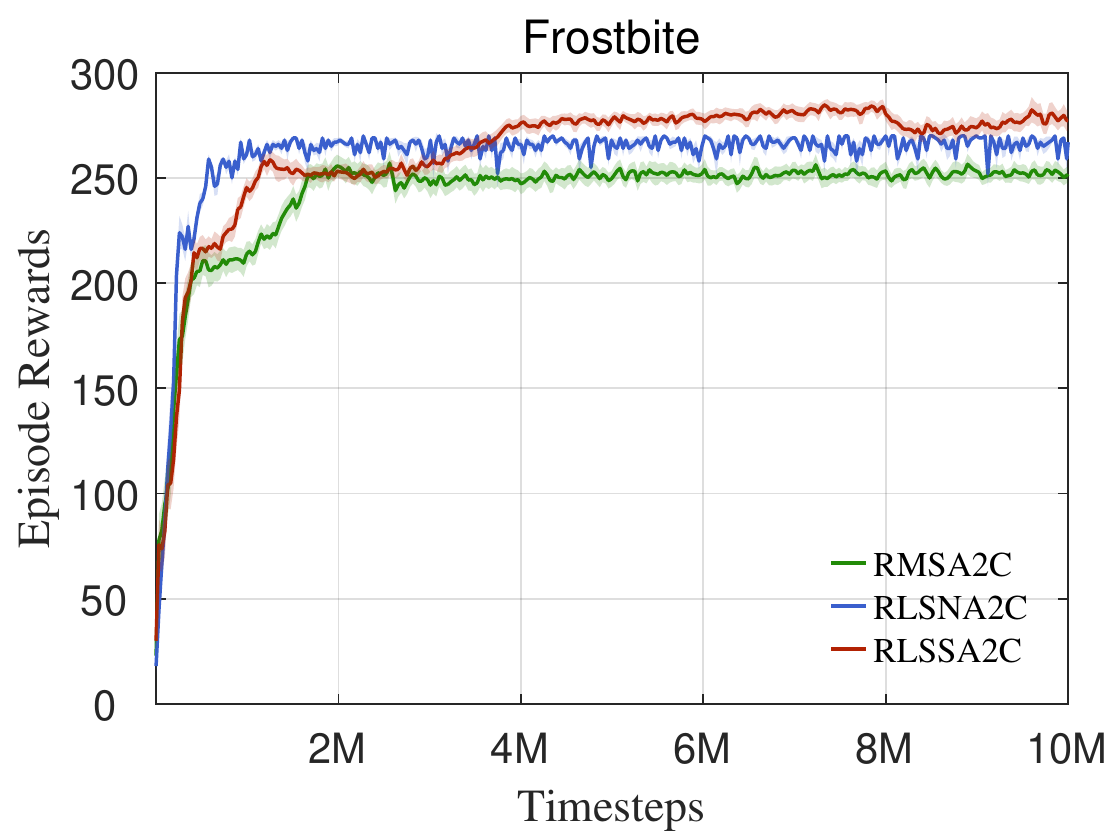}}
\subfigure{\includegraphics[width=4.2cm]{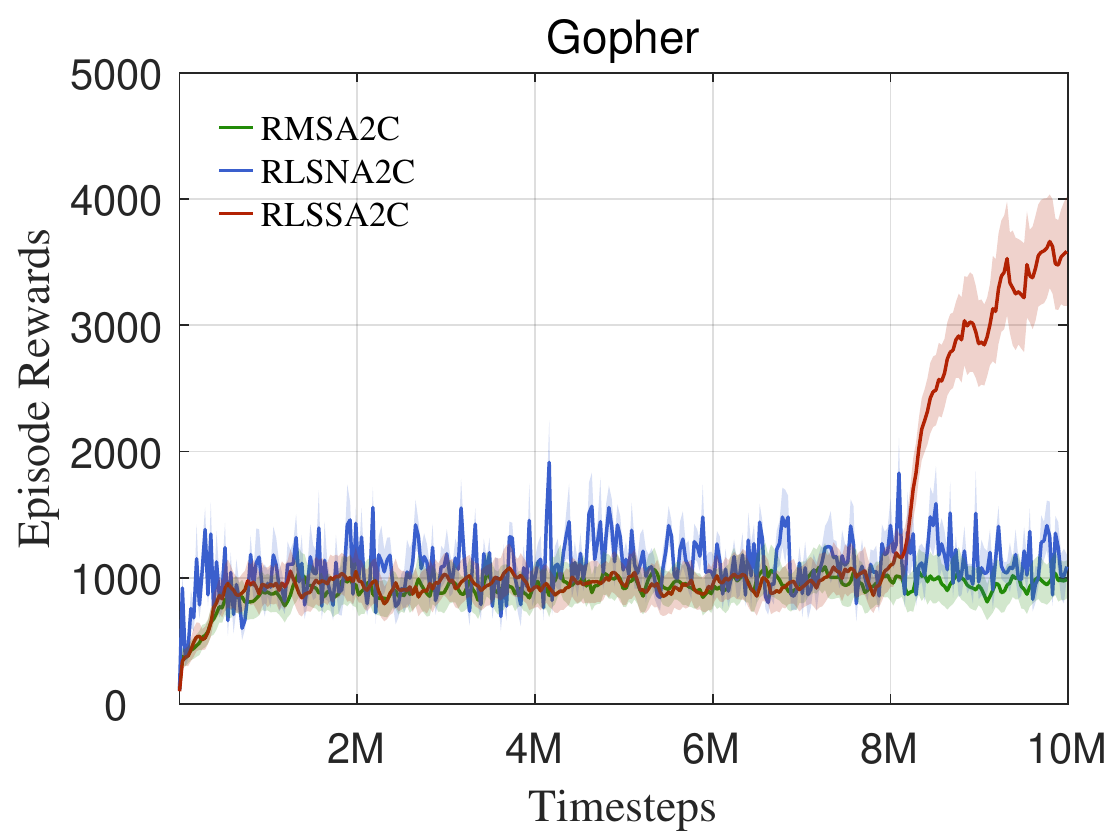}}
\subfigure{\includegraphics[width=4.2cm]{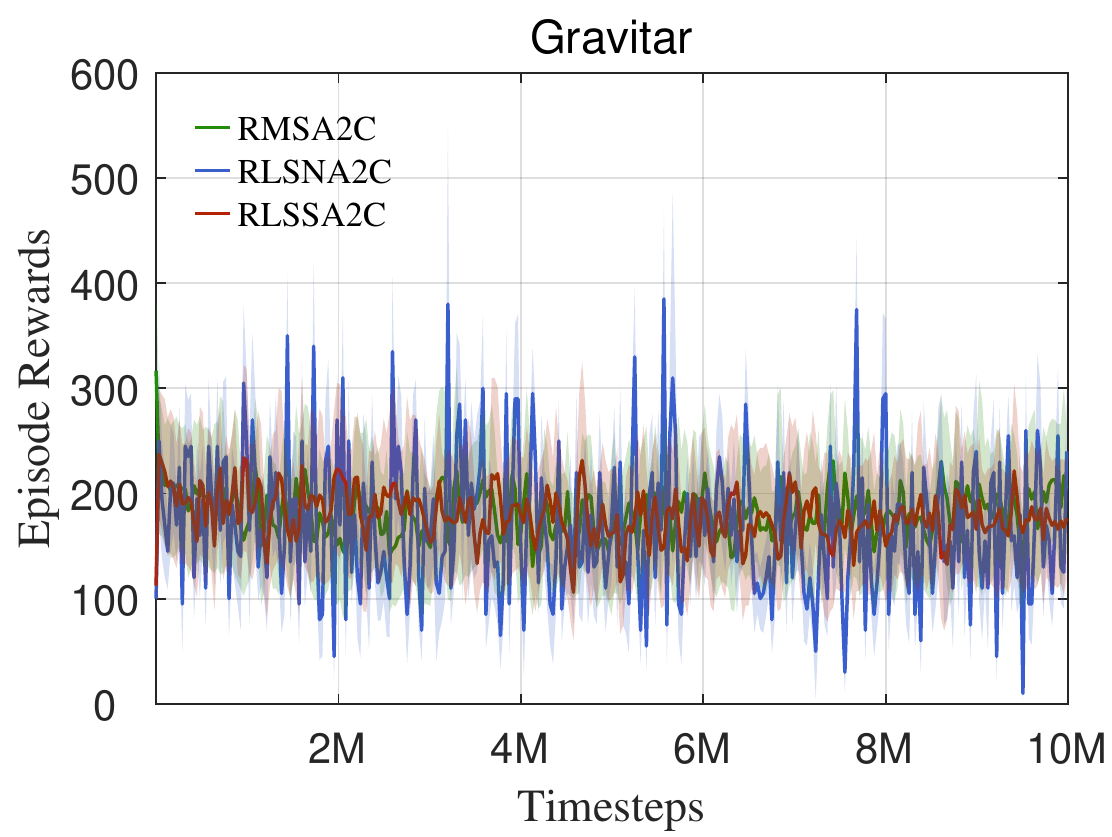}}
\subfigure{\includegraphics[width=4.2cm]{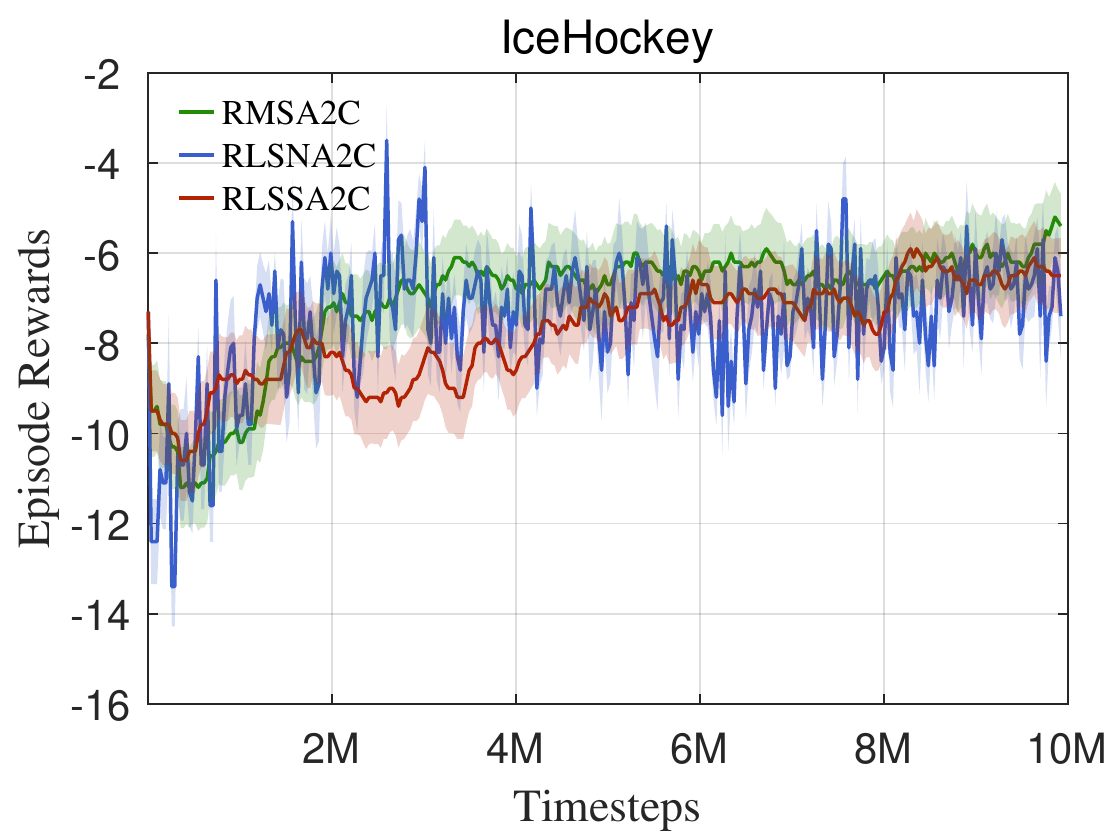}}
\centering
\subfigure{\includegraphics[width=4.2cm]{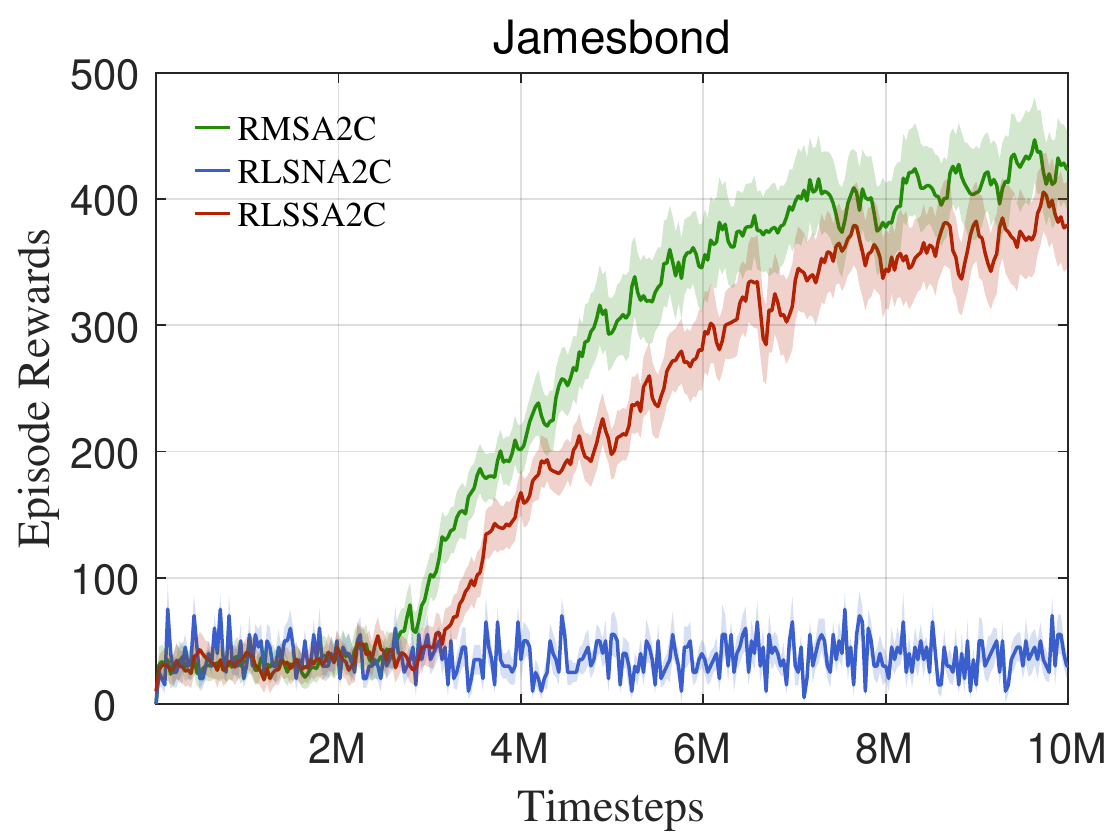}}
\subfigure{\includegraphics[width=4.2cm]{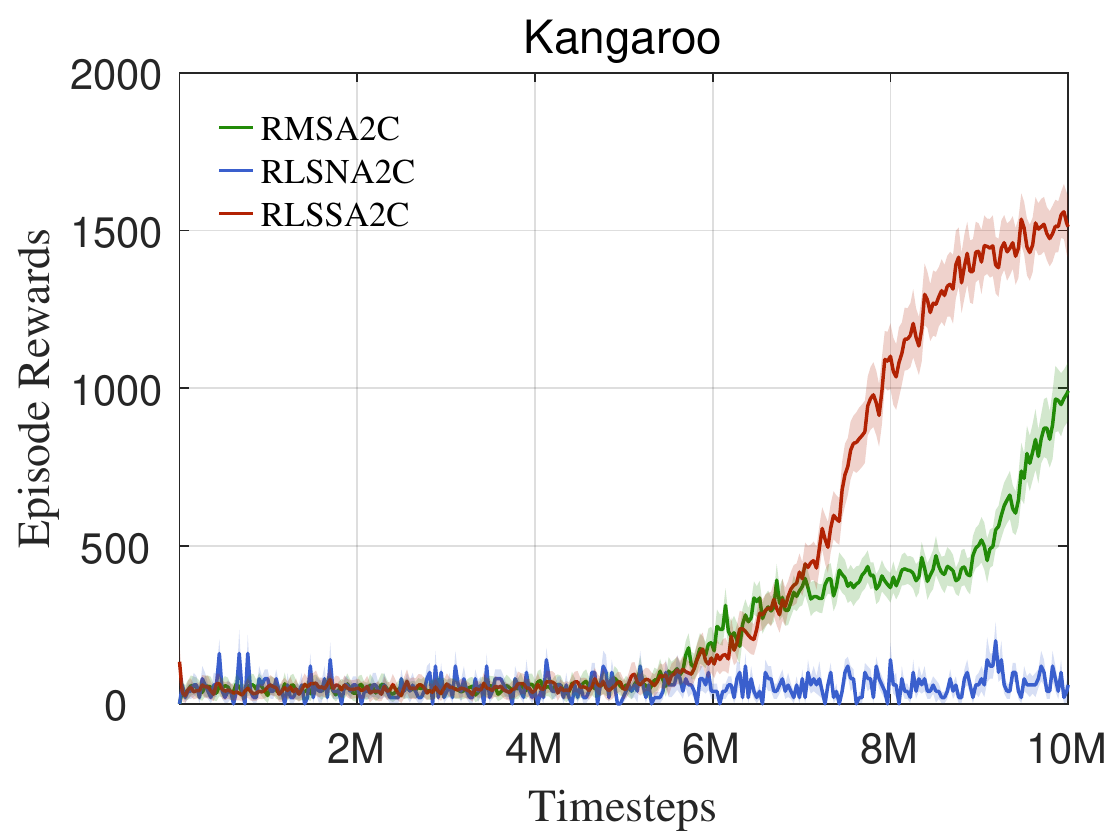}}
\subfigure{\includegraphics[width=4.2cm]{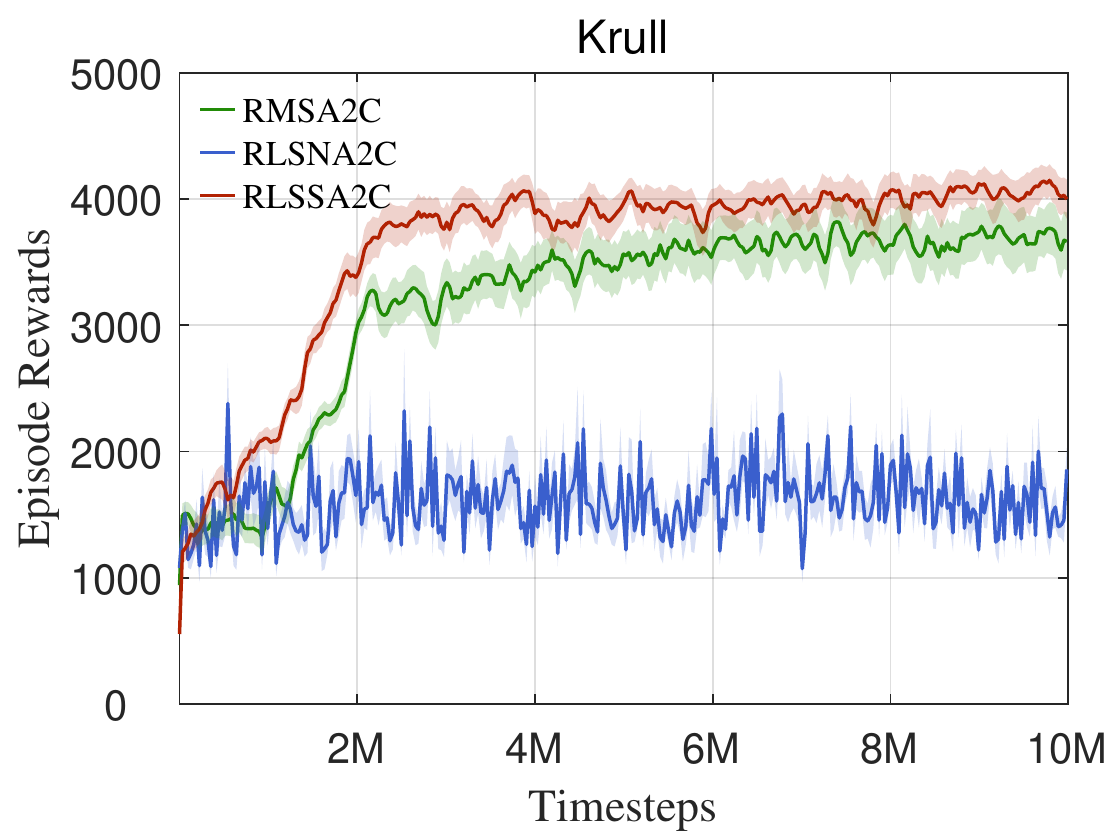}}
\subfigure{\includegraphics[width=4.2cm]{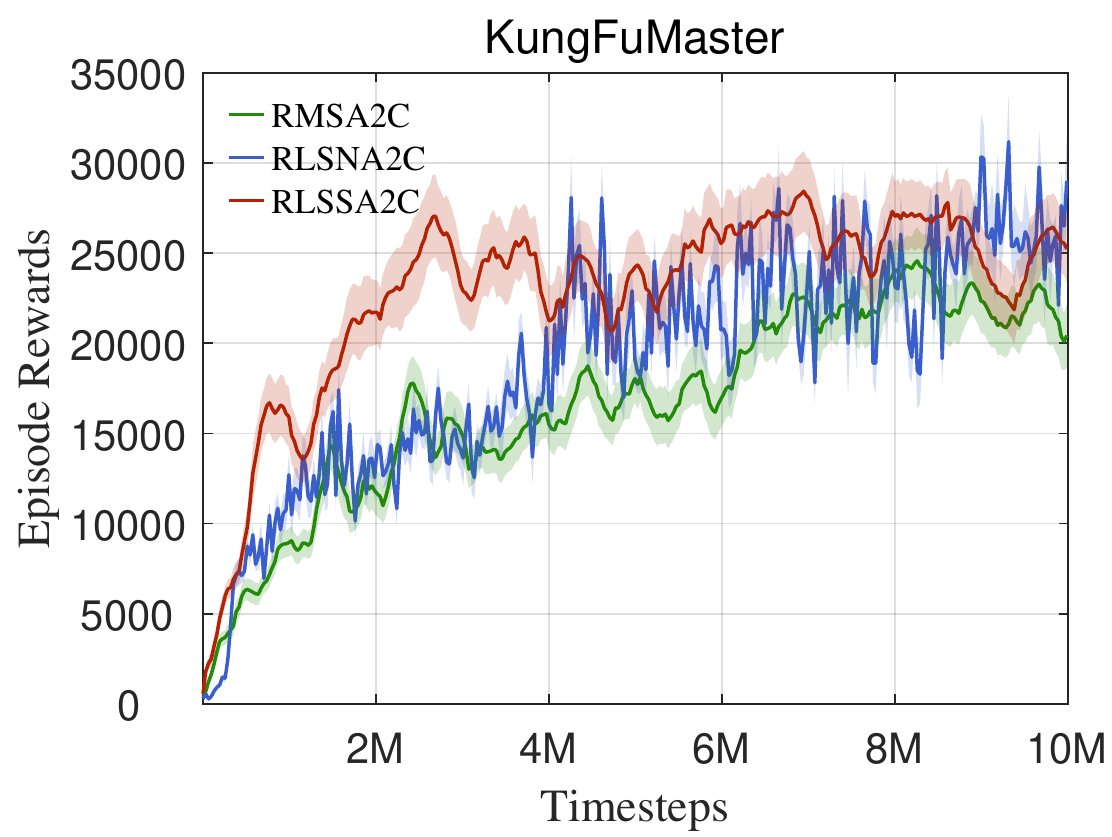}}
\centering
\subfigure{\includegraphics[width=4.2cm]{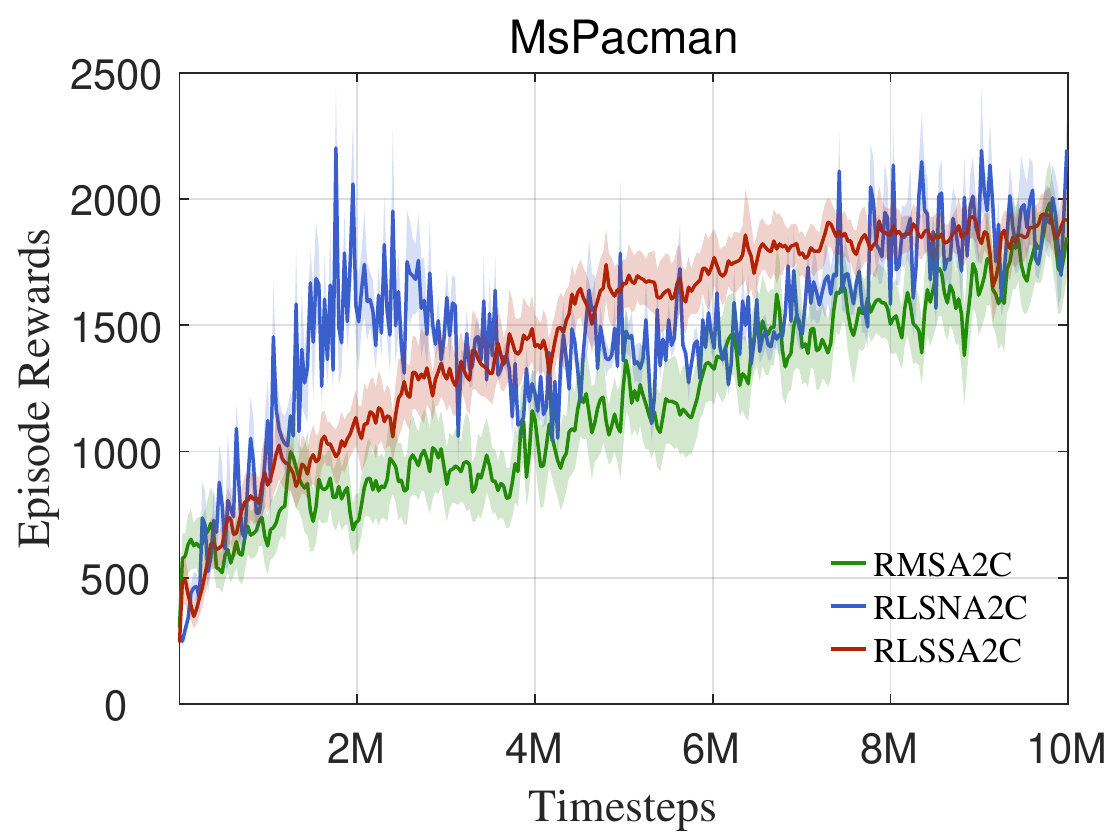}}
\subfigure{\includegraphics[width=4.2cm]{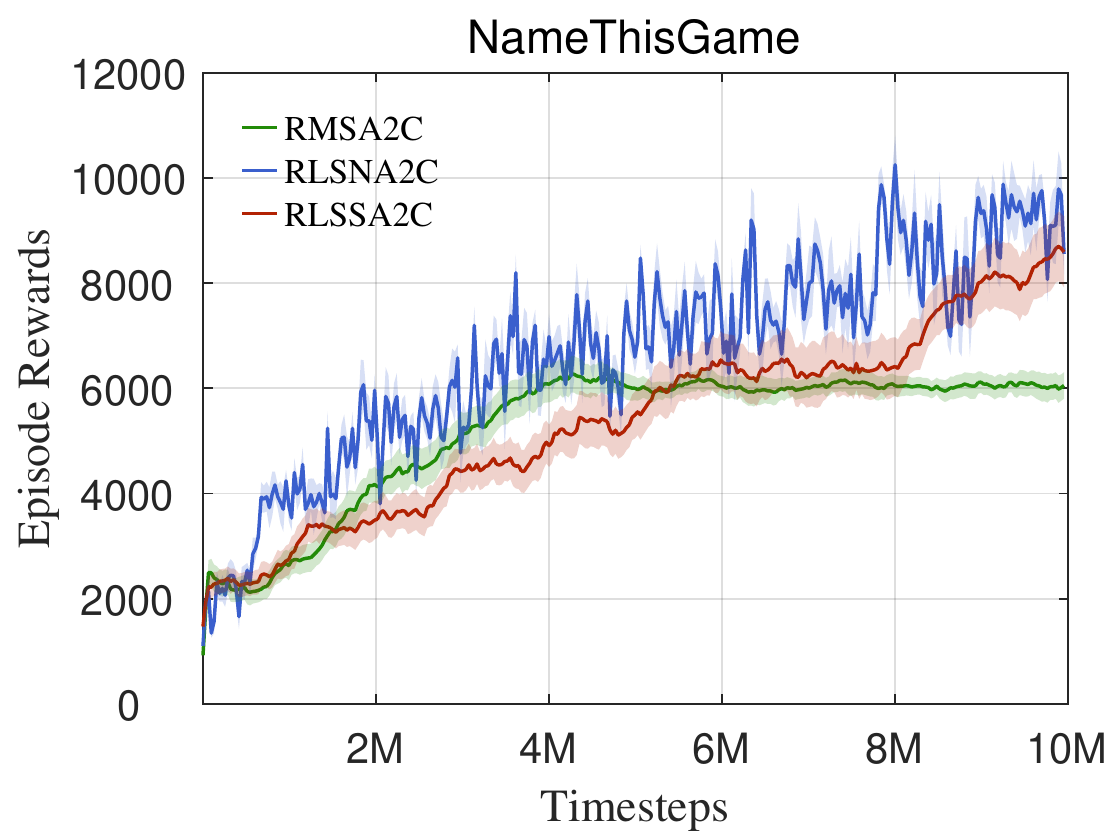}}
\subfigure{\includegraphics[width=4.2cm]{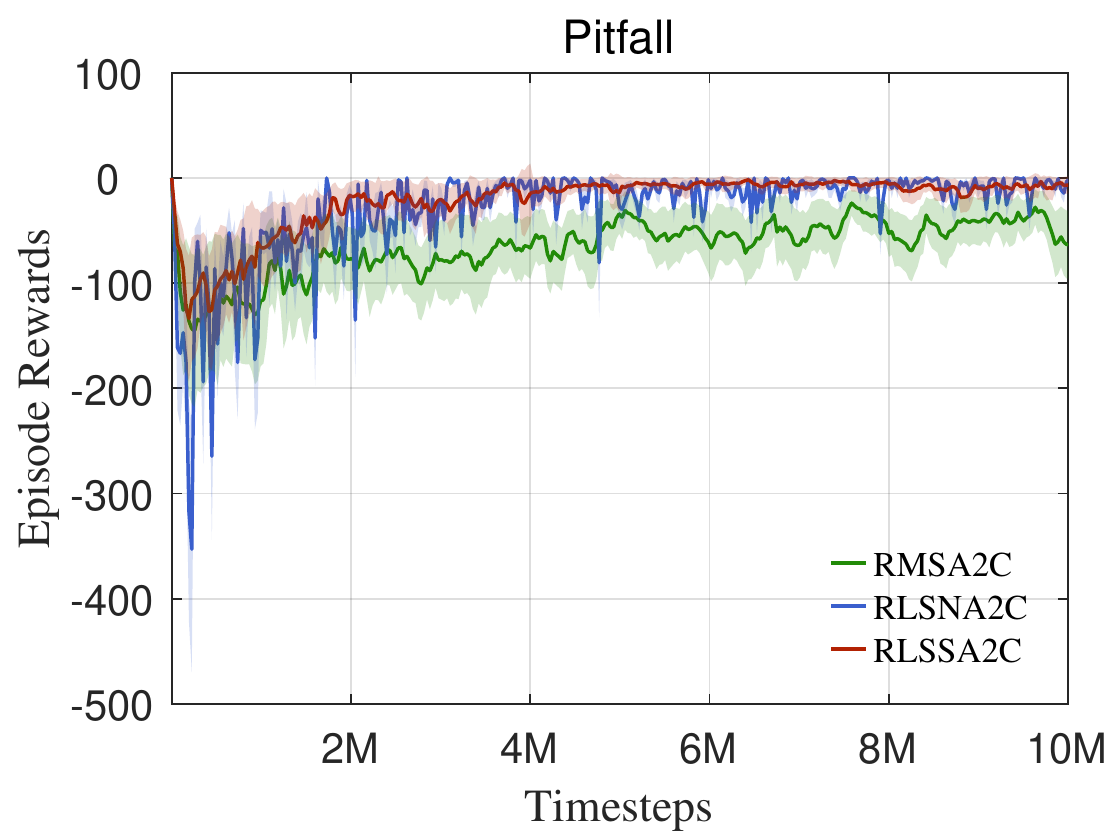}}
\subfigure{\includegraphics[width=4.2cm]{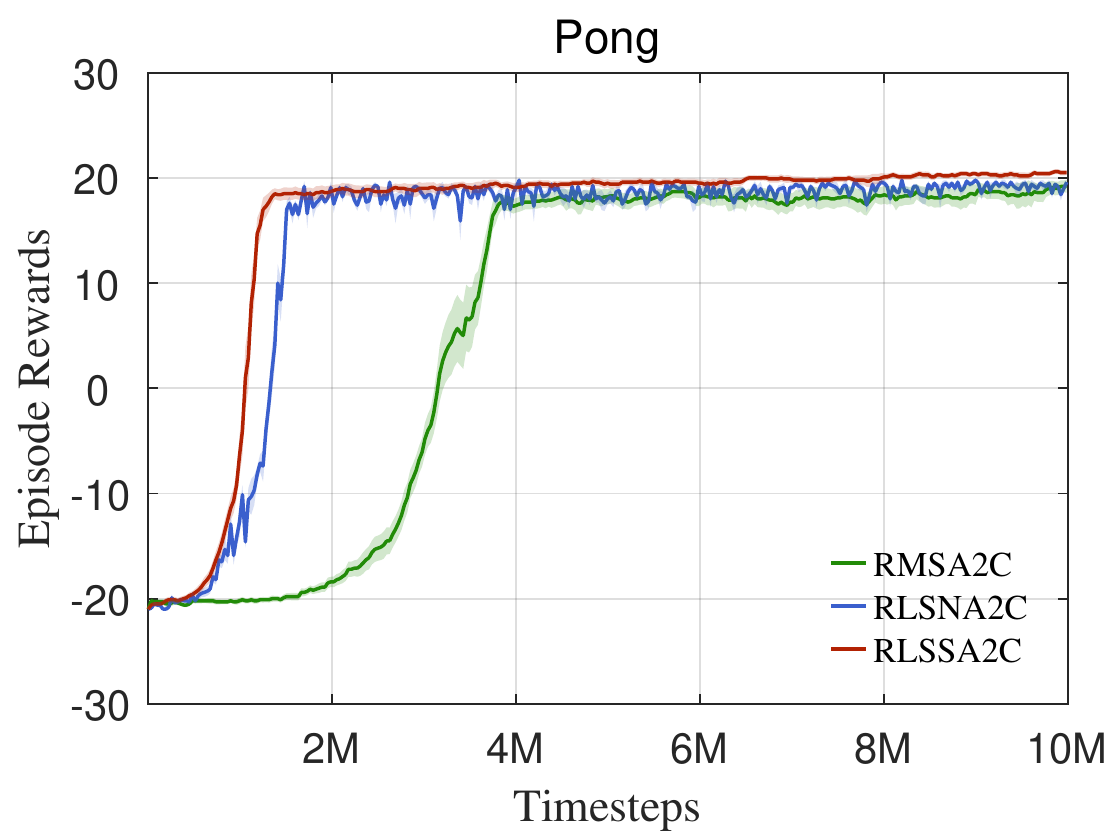}}

\captionsetup{labelformat=empty}
\caption{}
\end{figure*}
\begin{figure*}
\centering
\ContinuedFloat
\subfigure{\includegraphics[width=4.2cm]{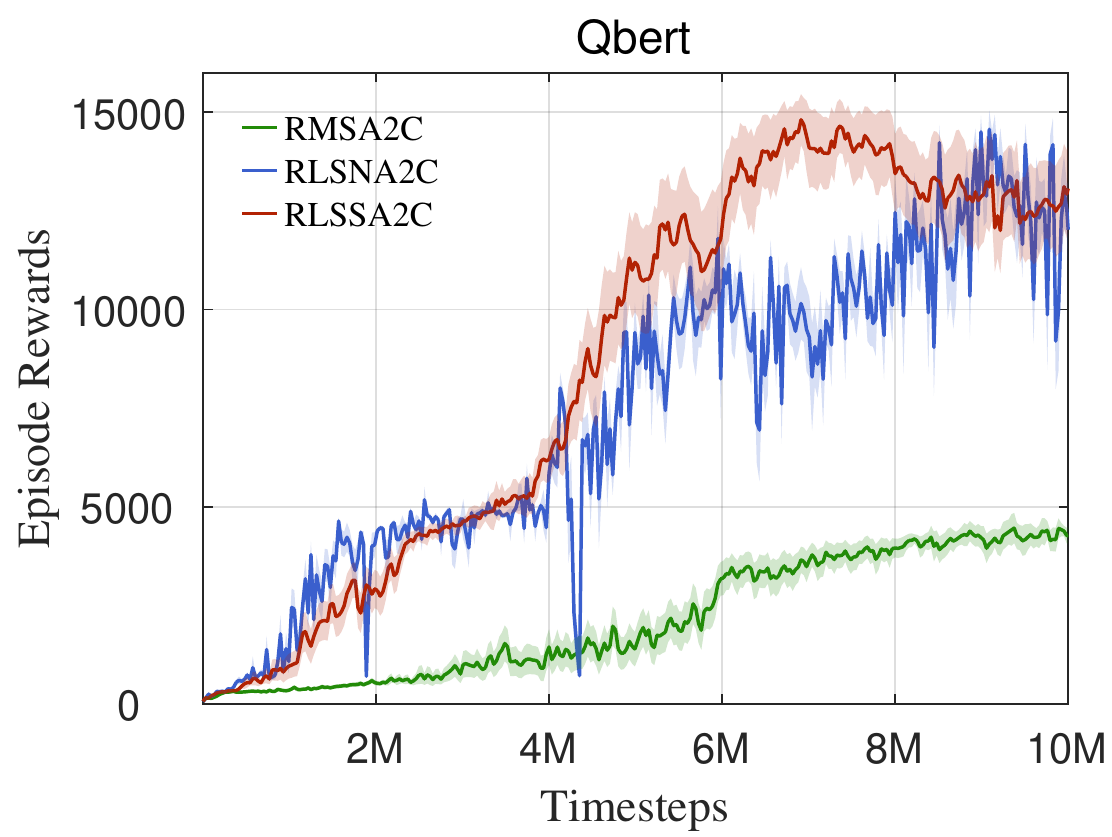}}
\subfigure{\includegraphics[width=4.2cm]{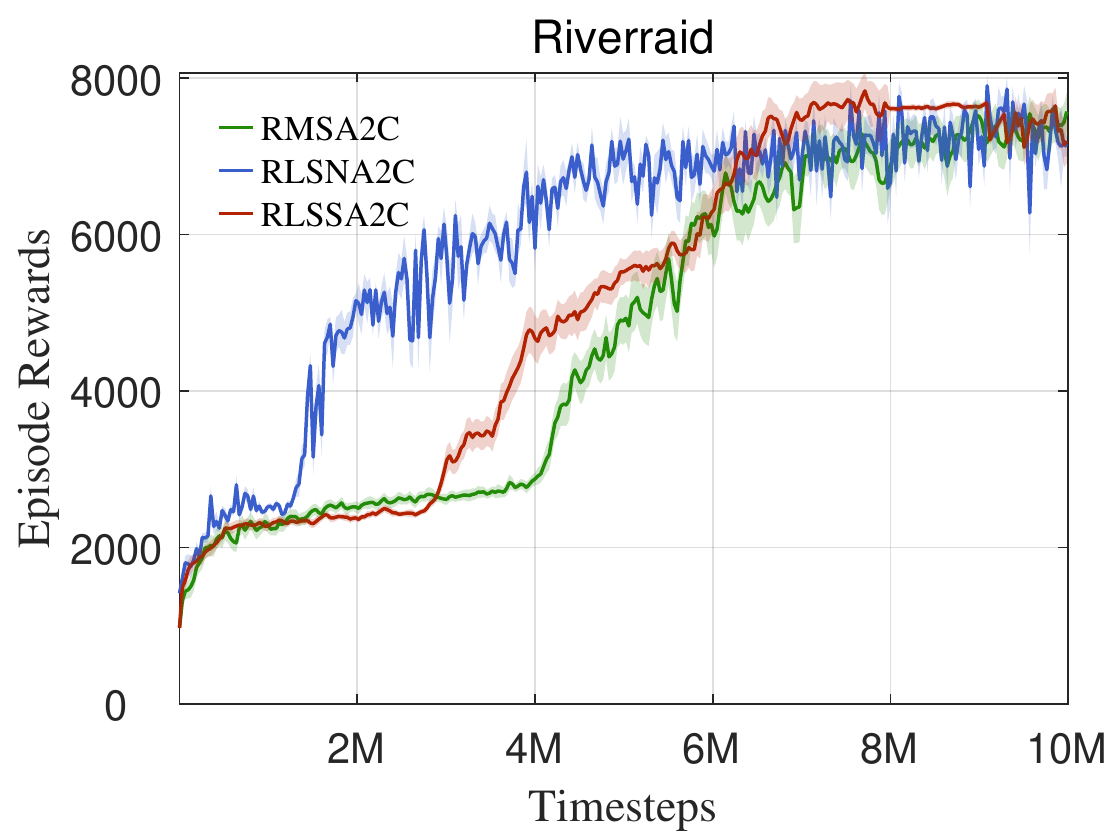}}
\subfigure{\includegraphics[width=4.2cm]{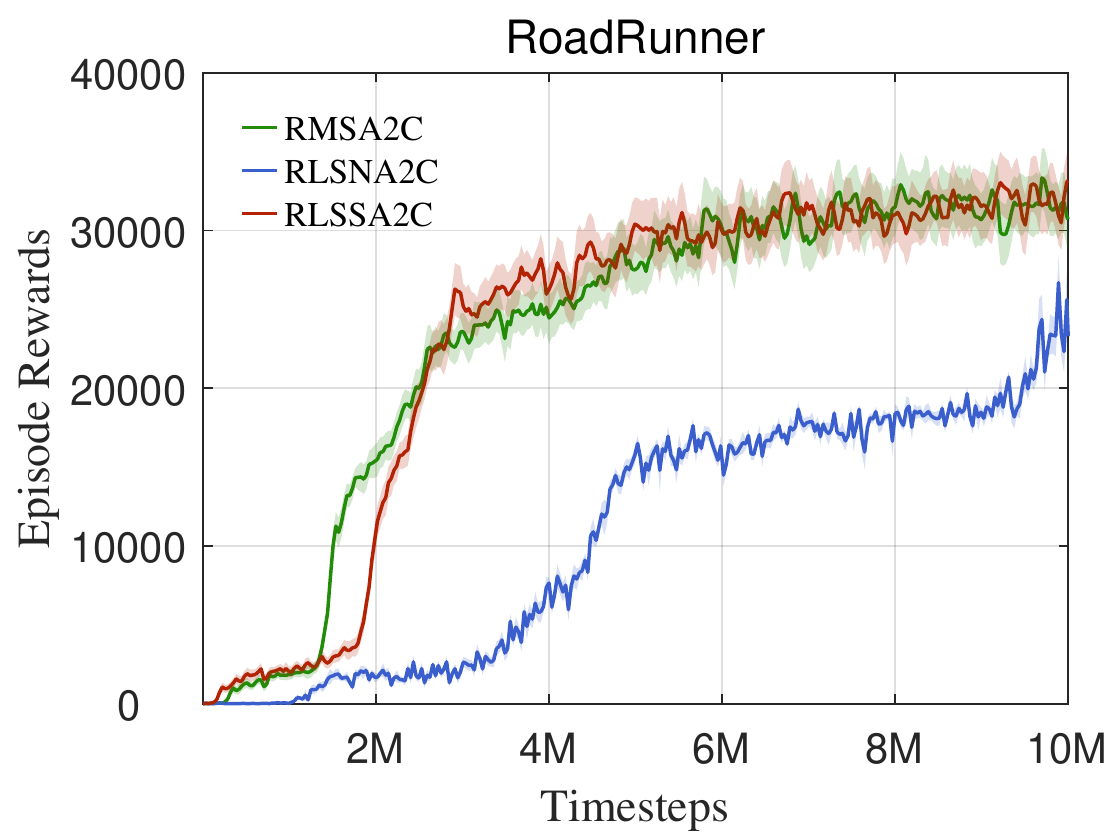}}
\subfigure{\includegraphics[width=4.2cm]{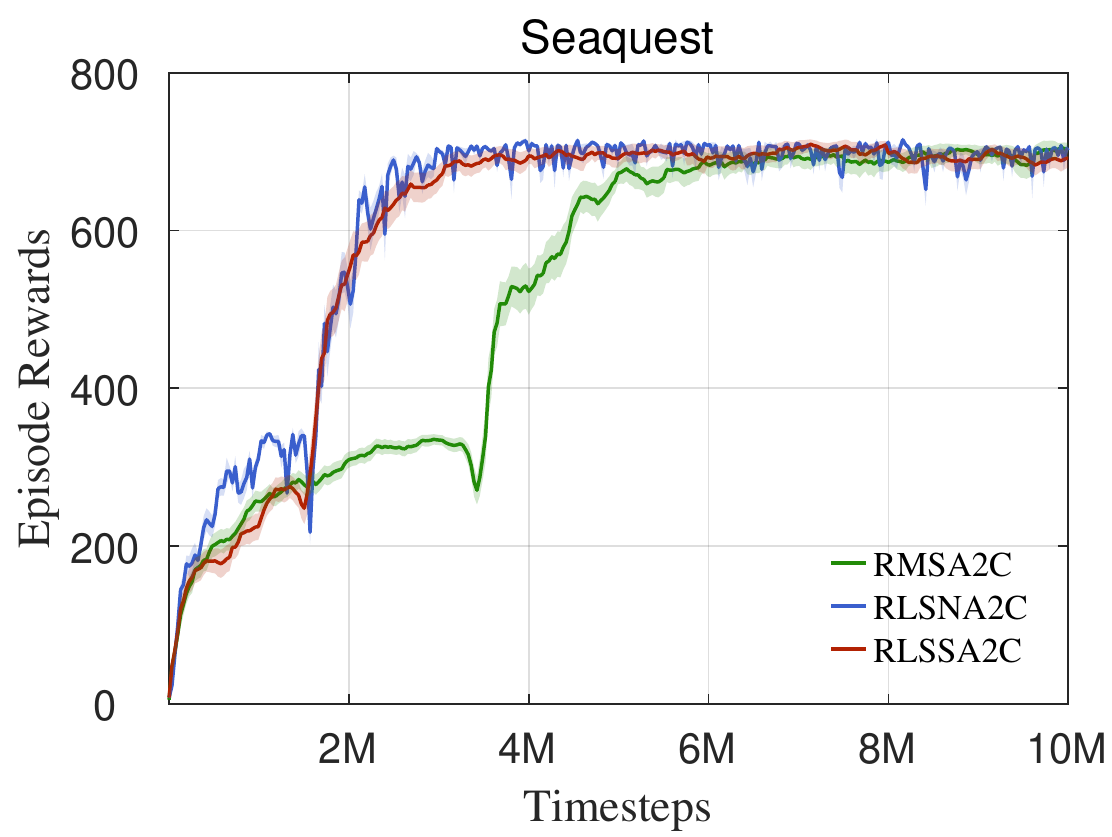}}
\centering
\subfigure{\includegraphics[width=4.2cm]{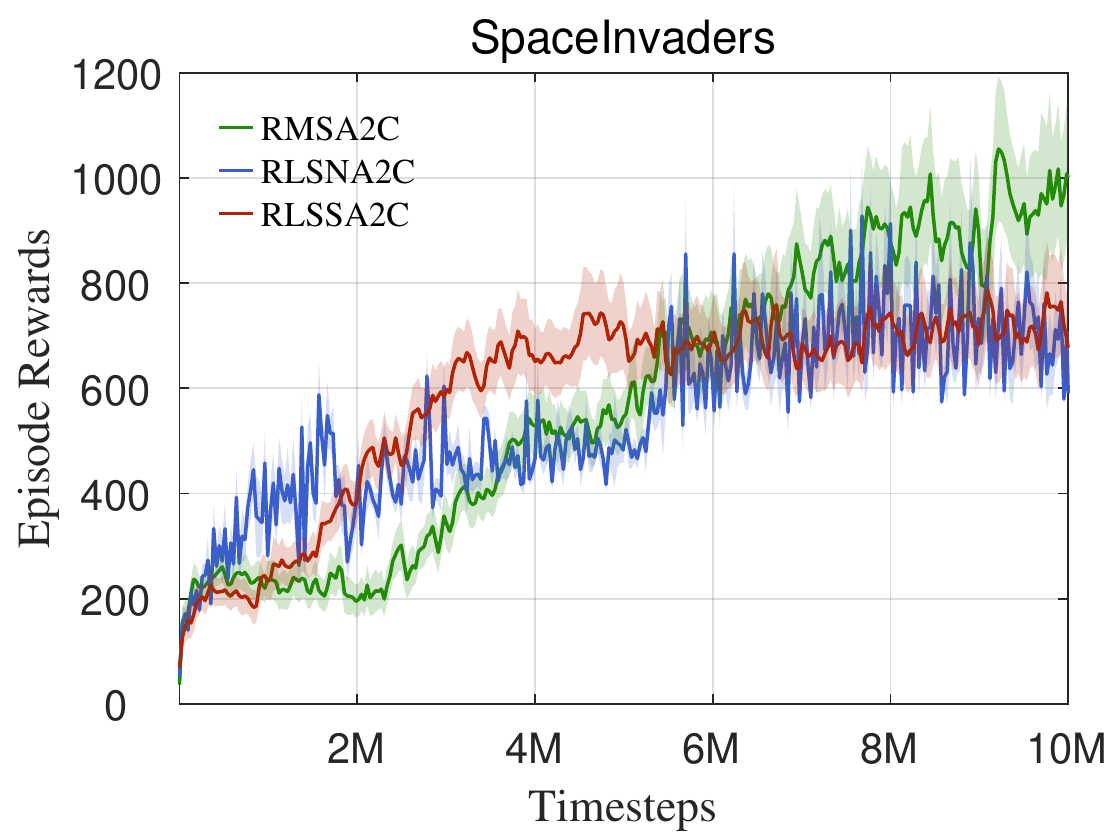}}
\subfigure{\includegraphics[width=4.2cm]{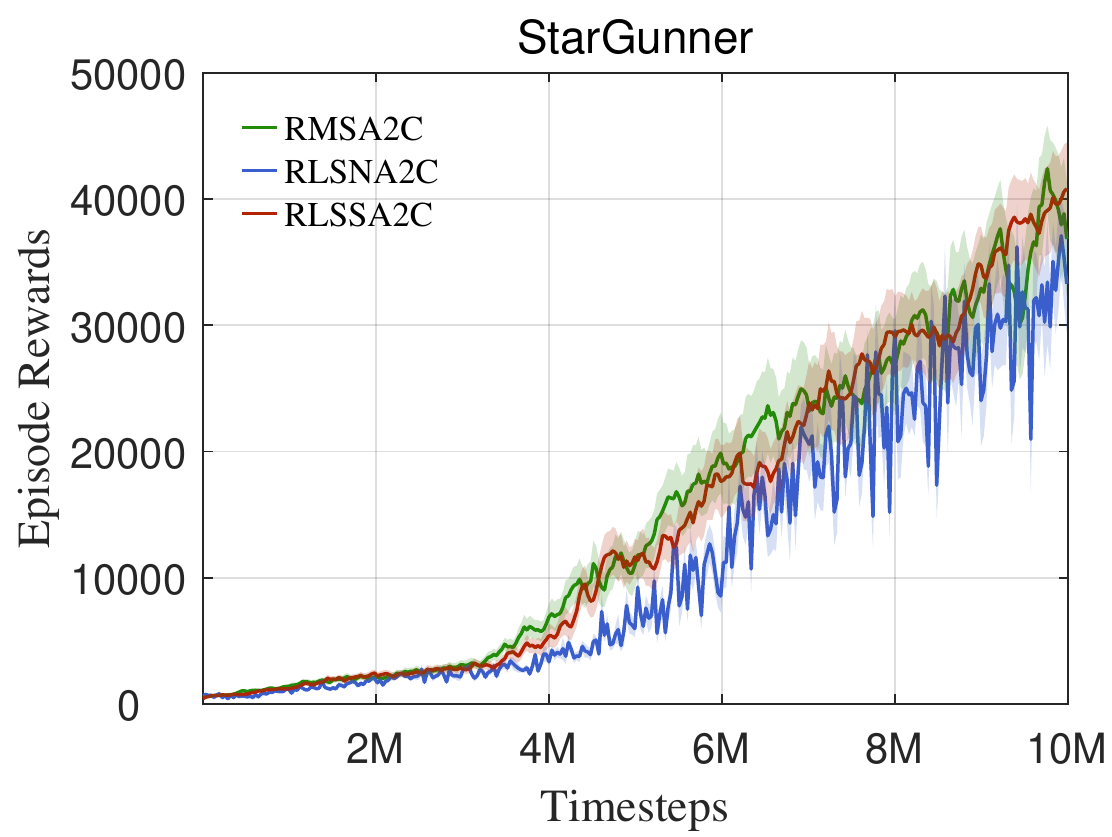}}
\subfigure{\includegraphics[width=4.2cm]{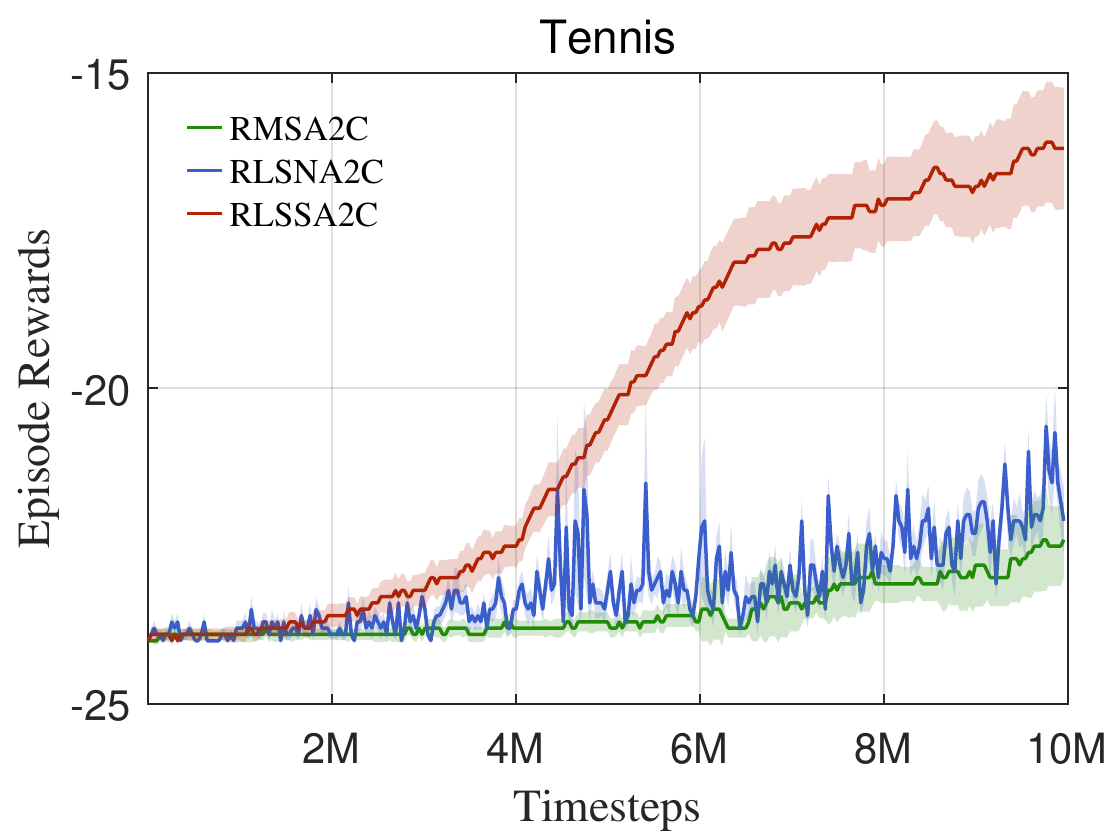}}
\subfigure{\includegraphics[width=4.2cm]{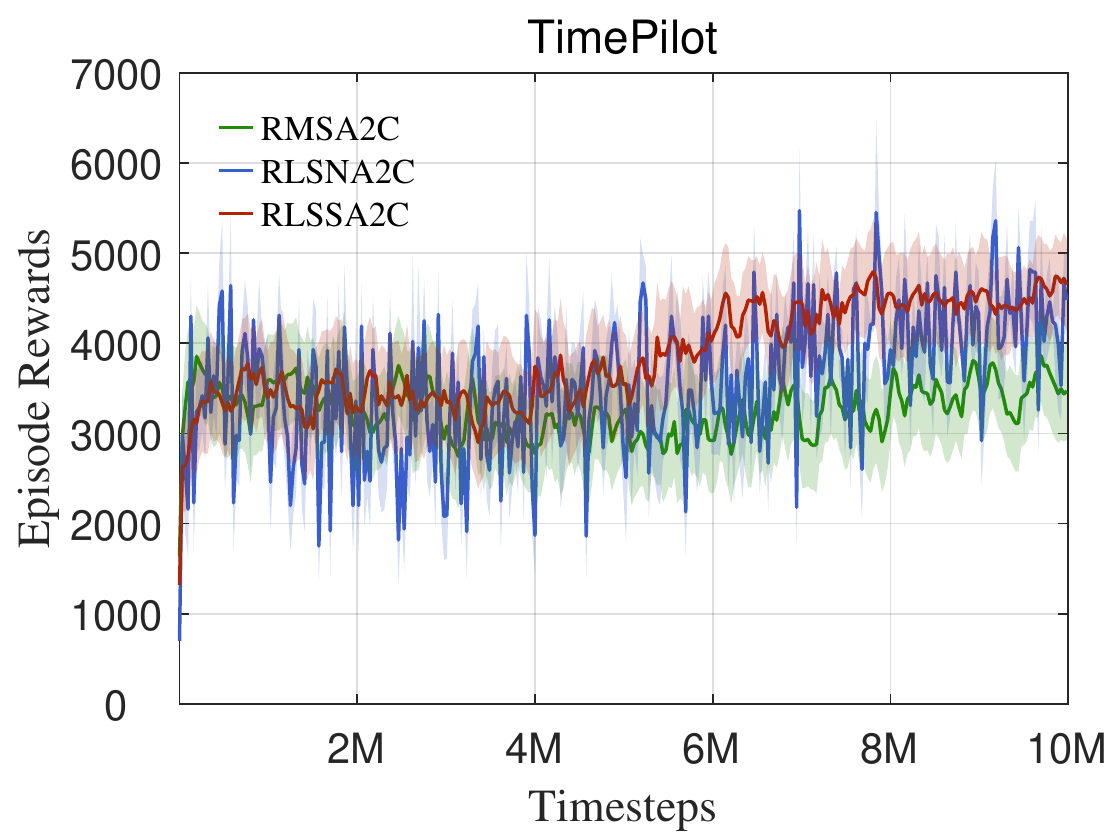}}
\centering
\subfigure{\includegraphics[width=4.2cm]{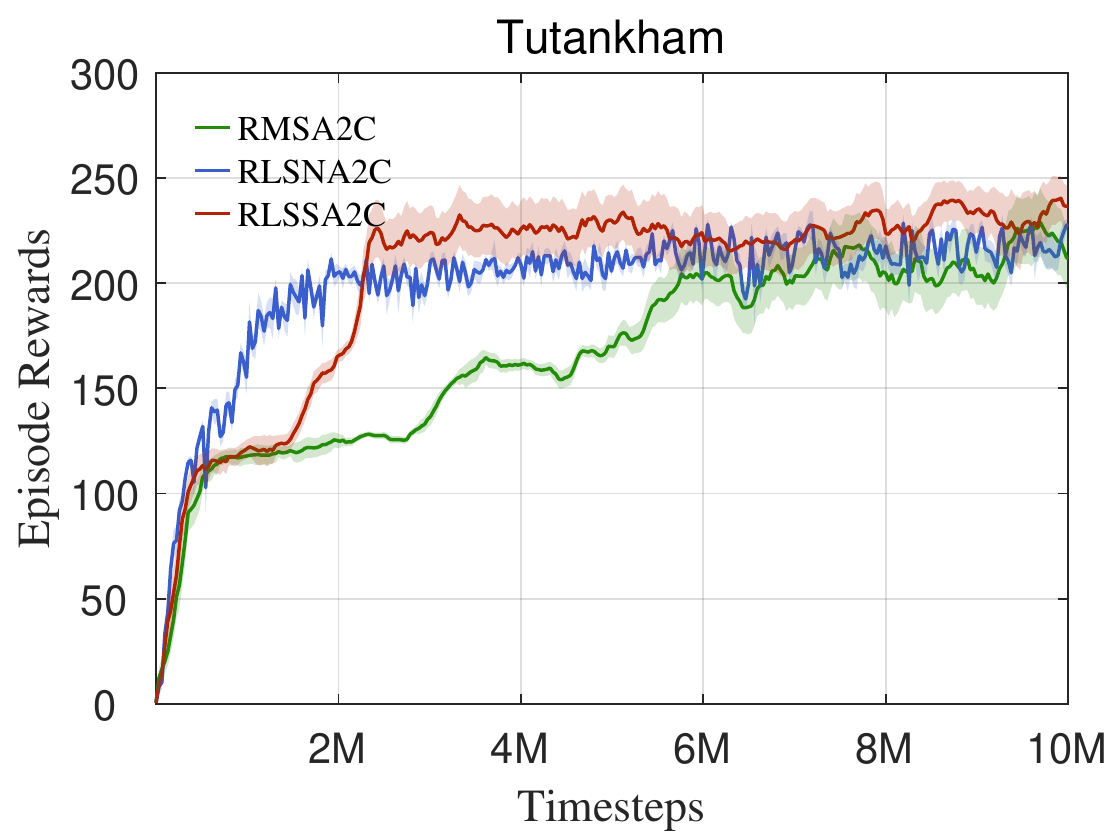}}
\subfigure{\includegraphics[width=4.2cm]{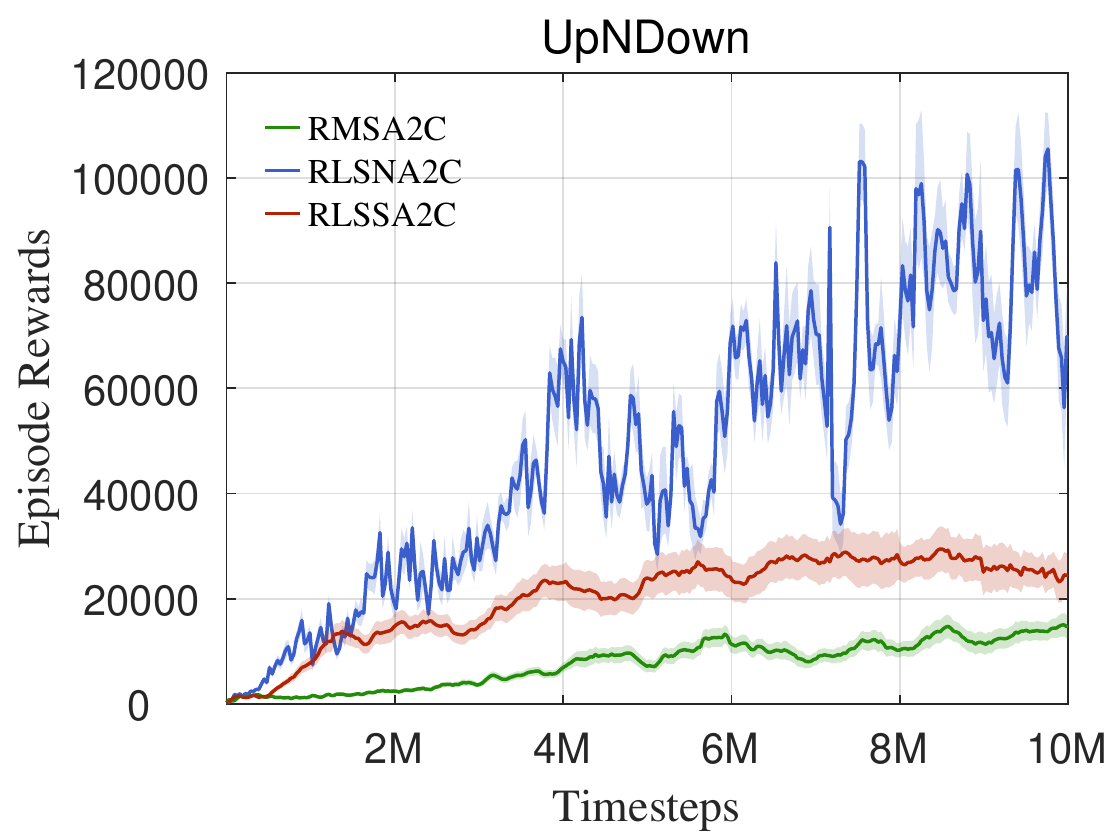}}
\subfigure{\includegraphics[width=4.2cm]{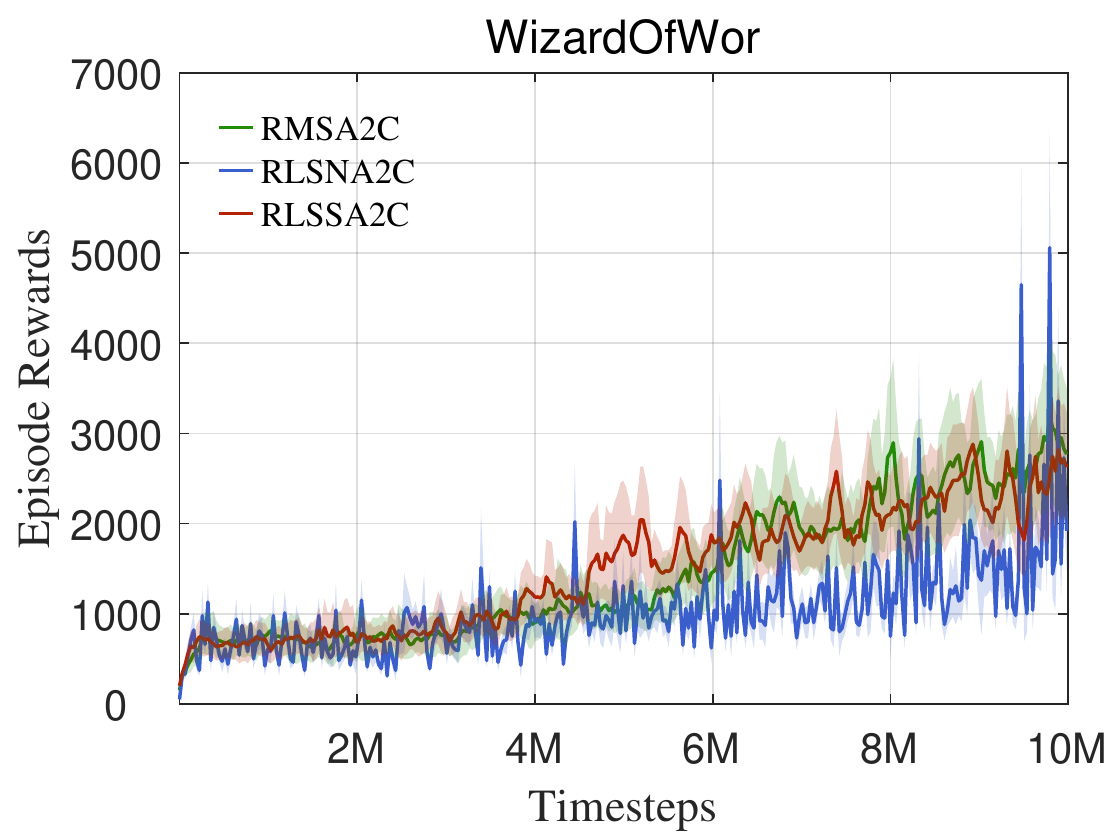}}
\subfigure{\includegraphics[width=4.2cm]{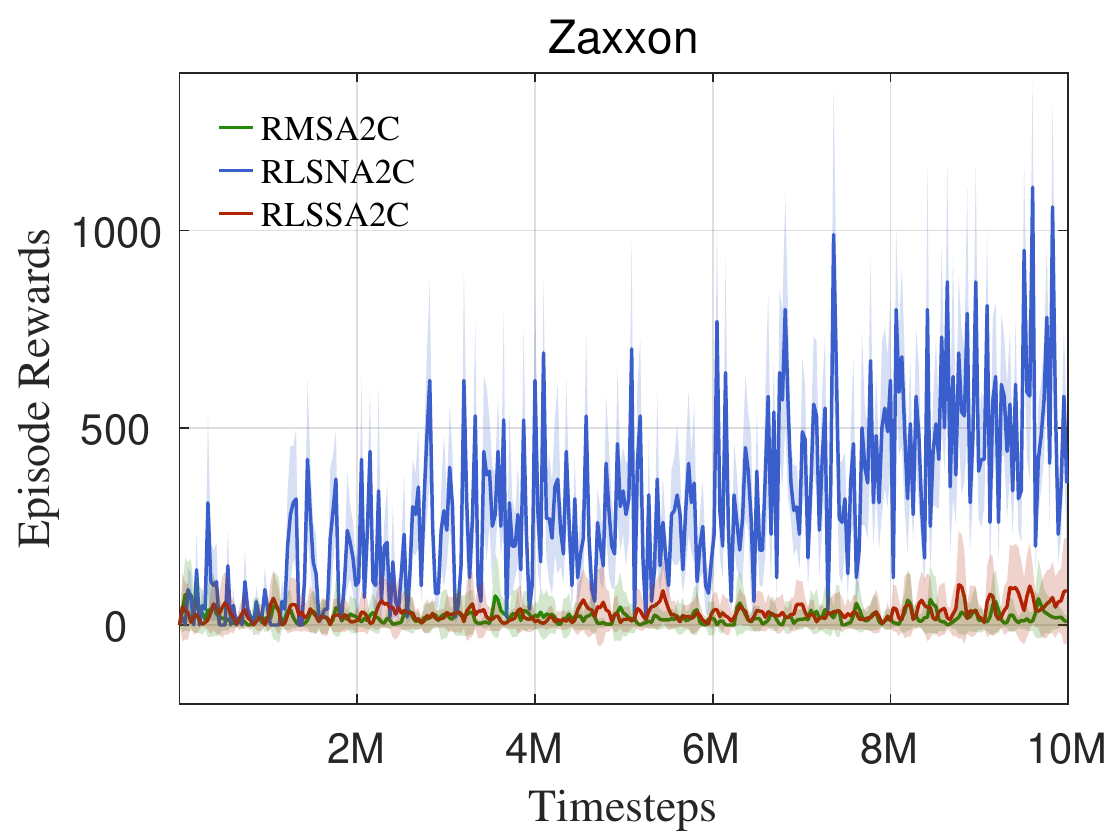}}

\caption{Convergence comparison of our algorithms against RMSA2C on 40 Atari games trained for 10M timesteps.}
\end{figure*}

\renewcommand{\arraystretch}{1.5}
\begin{table*}
  \centering
  \fontsize{7.4}{9}\selectfont
\begin{floatrow}
\capbtabbox{
    \begin{tabular*}{0.45\textwidth}{lrrr}
    \toprule[1pt]
    Game&~~~~~~~~~~RMSA2C&~~~~~~~RLSSA2C&~~~~~~~RLSNA2C\cr
    \midrule[1pt]
    Alien          &975.8           &{1375.7}   &\textbf{1675.0}            \cr
    Amidar         &234.9           &\textbf{405.6}    &301.1             \cr
	Assault        &1281.7          &\textbf{3346.9}   &3136.4            \cr
    Asterix        &2937.0          &5138.5   &\textbf{8140.0}   \cr
    Asteroids      &1427.4          &1344.8   &\textbf{1626.0}            \cr
    Atlantis       &1085676.0       &1885303.0&\textbf{2747970.0}         \cr
	BankHeist      &\textbf{1077.1}          &789.5    &662.0             \cr
    BattleZone     &4620.0          &\textbf{11200.0}  &8100.0            \cr
    BeamRider      &4602.9          &5225.5   &\textbf{5313.0}   \cr
    Bowling        &29.1          &\textbf{30.9}     &29.3              \cr
	Boxing         &93.3            &97.2     &\textbf{100.0}    \cr
    Breakout       &389.9           &\textbf{460.5}    &445.9    \cr
    Centipede      &2875.6          &\textbf{4300.1}   &2950.9            \cr
    DemonAttack    &\textbf{25309.3}&13252.4  &2619.5            \cr
    DoubleDunk     &-14.4           &-15.1    &\textbf{-2.0}     \cr
    FishingDerby   &10.2            &18.3     &\textbf{27.8}     \cr
	Frostbite      &250.7           &\textbf{277.2}    &267.0             \cr
    Gopher         &986.0           &\textbf{3585.0}   &1088.0   \cr
    Gravitar       &204.5           &176.5    &\textbf{240.0}             \cr
    IceHockey      & -11.2          &\textbf{-6.5 }    &-7.4              \cr
    \bottomrule[1pt]
    \end{tabular*}
  \hfil\hfil\hfil\hfil\hfil\hfil~~~~~~~~~~\hfil\hfil\hfil\hfil\hfil\hfil\hfil\hfil\hfil\hfil
    \begin{tabular*}{0.45\textwidth}{lrrr}
    \toprule[1pt]
    Game&~~~~~~~RMSA2C&~~~~~~~RLSSA2C&~~~~~~~RLSNA2C\cr
    \midrule[1pt]
	Jamesbond      &\textbf{423.5}           &379.0    &30.0              \cr
    Kangaroo       &992.0           &\textbf{1512.0}   &60.0              \cr
    Krull          &7327.0          &\textbf{7996.3}   &3715.0   \cr
    KungFuMaster   &20427.0         &25224.0    &\textbf{29000.0}  \cr
	MsPacman       &1846.9          &1916.4   &\textbf{2195.0}   \cr
    NameThisGame   &6054.1          &\textbf{8592.8}   &8555.0   \cr
    Pitfall        &-63.4           &-6.7     &\textbf{-2.8}     \cr
    Pong           &19.2            &\textbf{20.5}     &19.6              \cr
    Qbert          &4267.5          &\textbf{13064.5}  &12020.0           \cr
    Riverraid      &\textbf{7572.3}          &7193.4   &7125.0            \cr
    RoadRunner     &30705.0         &\textbf{33160.0}  &23300.0  \cr
	Seaquest       &1754.8          &1728.8   &\textbf{1756.0}            \cr
    SpaceInvaders  &\textbf{1001.2} &677.4    &591.0             \cr
    StarGunner     &36820.0         &\textbf{40808.0}    &33280.0\cr
    Tennis         &-22.4           &\textbf{-16.2}    &-22.1             \cr
	TimePilot      &3471.0          &\textbf{4648.0}   &4480.0   \cr
    Tutankham      &211.9           &\textbf{236.5}    &227.7             \cr
    UpNDown        &14666.5         &24466.6  &\textbf{69848.0}           \cr
    WizardOfWor    &\textbf{2770.0}          &2626.0   &1920.0            \cr
	Zaxxon         &8.0             &87.0     &\textbf{360.0}             \cr
    \bottomrule[1pt]
    \end{tabular*}
}
{
 \caption{\textsc{Last 100 Average Episode Rewards of our algorithms and RMSA2C on 40 Atari Games Trained for 10M Timesteps}}
 \label{tab.tb2}
}
\end{floatrow}
\end{table*}

\renewcommand{\arraystretch}{1.5}
\begin{table}[tp]
  \centering
  \fontsize{7.4}{9}\selectfont
  \begin{threeparttable}
  \caption{\textsc{Running Speed Comparison on Six Atari Games}\\{( Timesteps\! /\! Second )}}
  \label{tab:performance_comparison}
    \begin{tabular}{lccccc}
    \toprule[1pt]
    Game&RMSA2C&PPO&ACKTR&RLSSA2C&RLSNA2C\cr
    \midrule[1pt]
    Alien        &2897 &320 &1690 &2056 &2021 \cr
    Breakout     &3033 &321 &1668 &2057 &2150 \cr
    Pong         &3265 &324 &1754 &2411 &2402 \cr
    SpaceInvaders&3124 &328 &1731 &2293 &2306 \cr
    StarGunner   &3436 &341 &1794 &2515 &2512 \cr
	Zaxxon       &3201 &336 &1750 &2301 &2311 \cr
    \midrule[1pt]
    Mean         &3159 &328 &1731 &2272 &2284 \cr
    \bottomrule[1pt]
    \end{tabular}
    \end{threeparttable}
\end{table}

\subsection{Continuous Control Evaluation}
MuJoCo is a high-dimensional continuous control benchmark platform.
In this set of experiments, we select 11 tasks from the MuJoCo environment for performance evaluation.
In contrast to in Atari, the state in MuJoCo is a multiple dimensional vector, and the action space is continuous.

\begin{figure*}
\centering
\subfigure{\includegraphics[width=4.2cm]{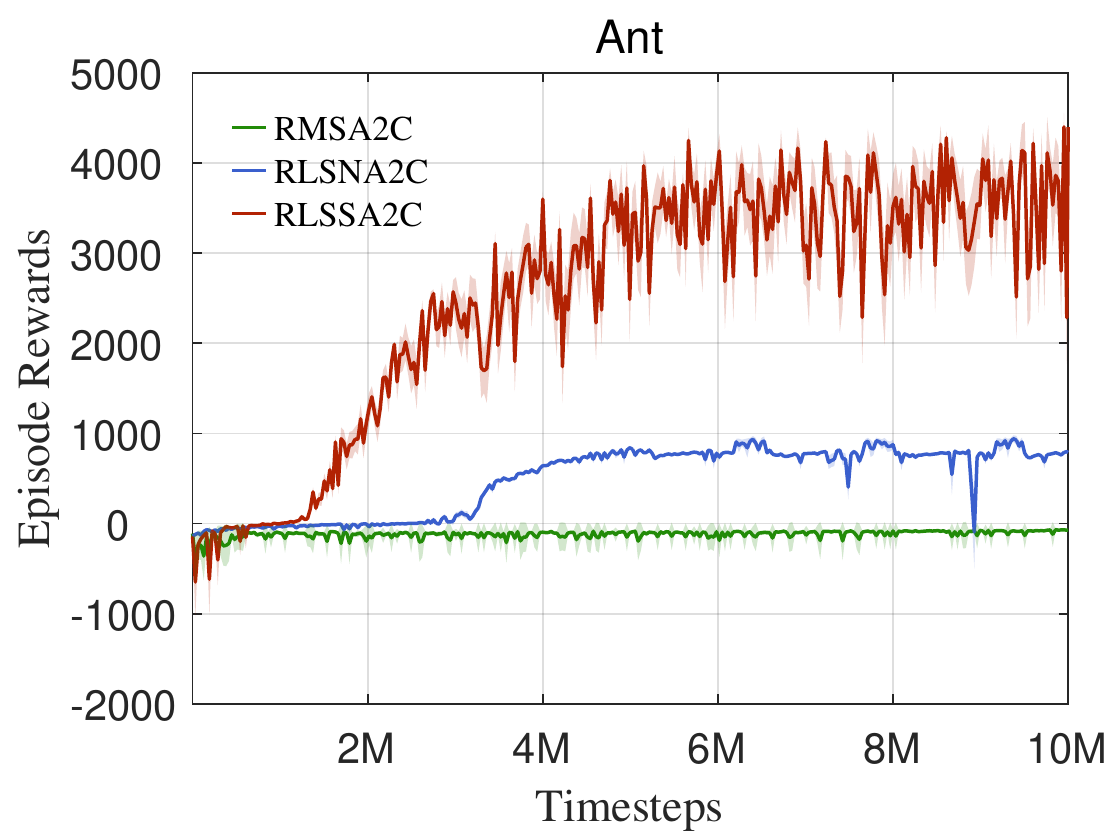}}
\subfigure{\includegraphics[width=4.2cm]{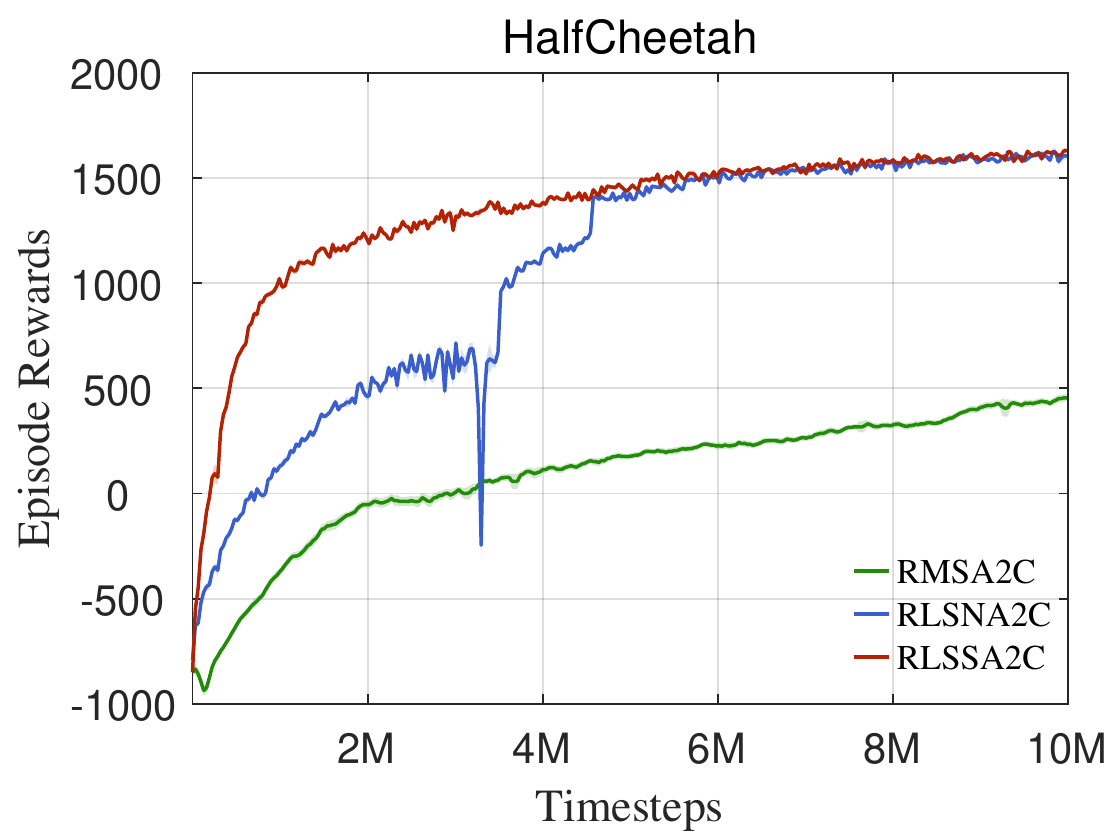}}
\subfigure{\includegraphics[width=4.2cm]{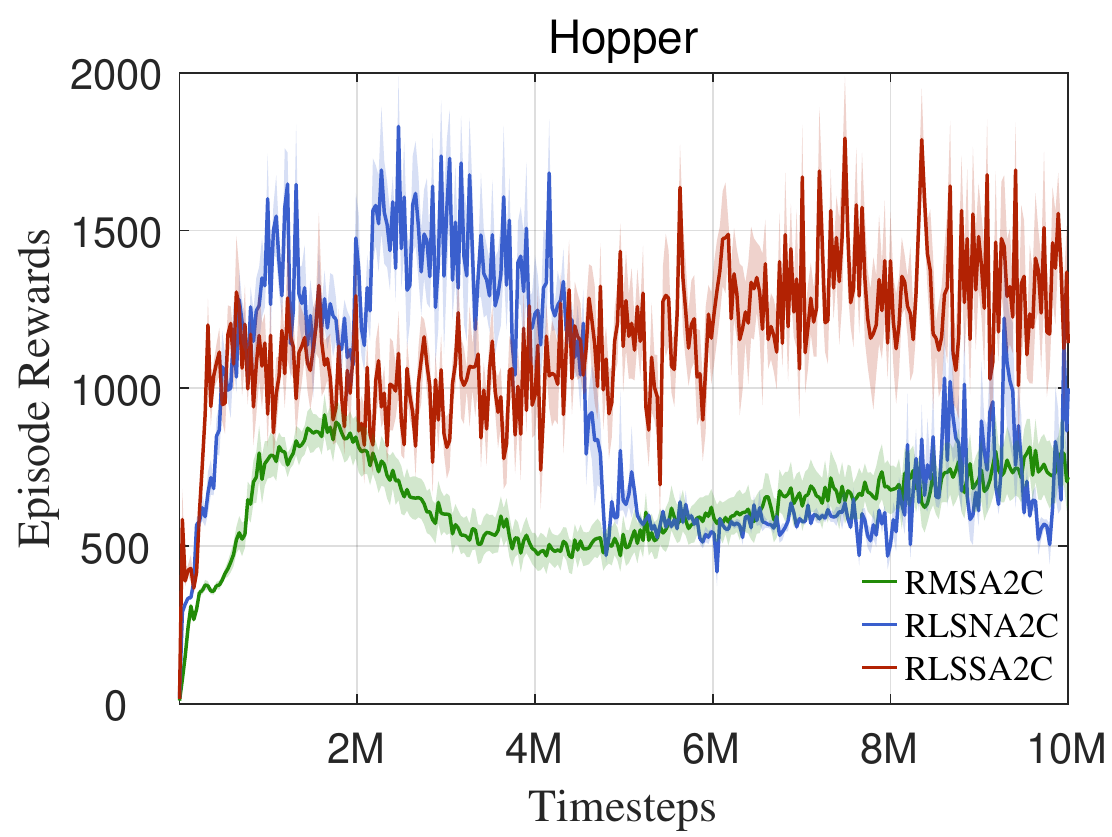}}
\subfigure{\includegraphics[width=4.2cm]{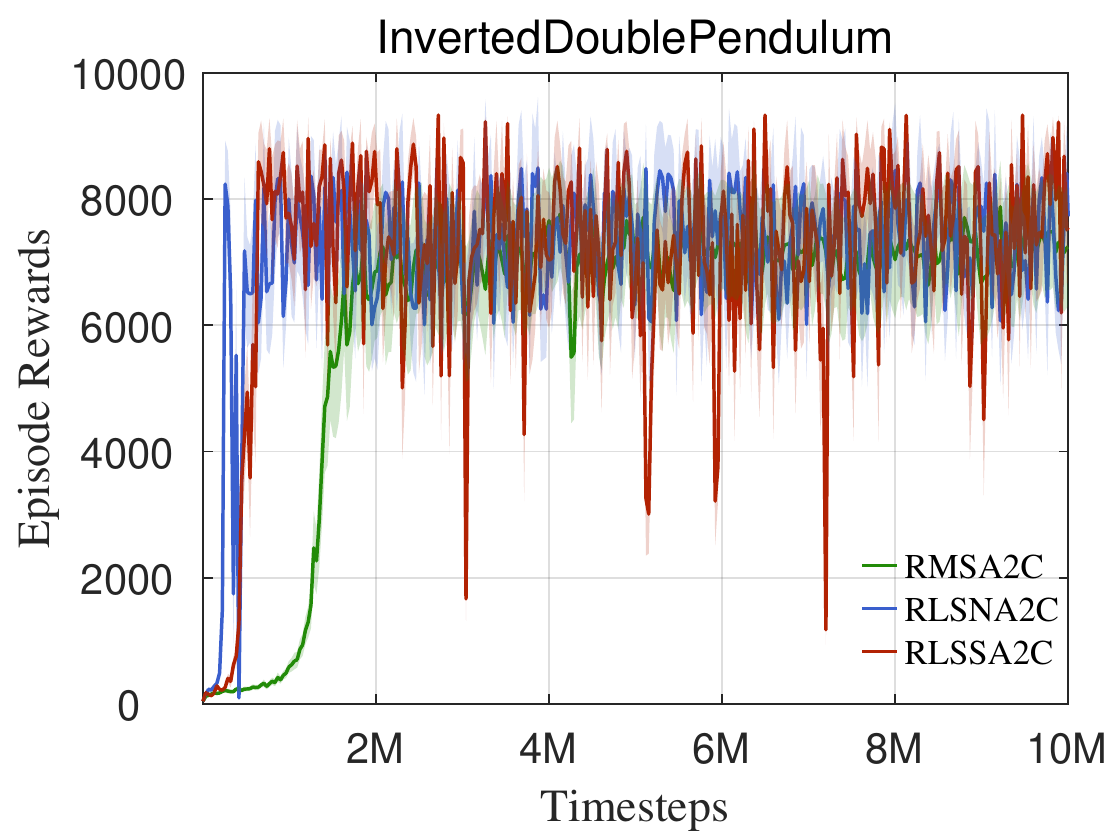}}
\centering
\subfigure{\includegraphics[width=4.2cm]{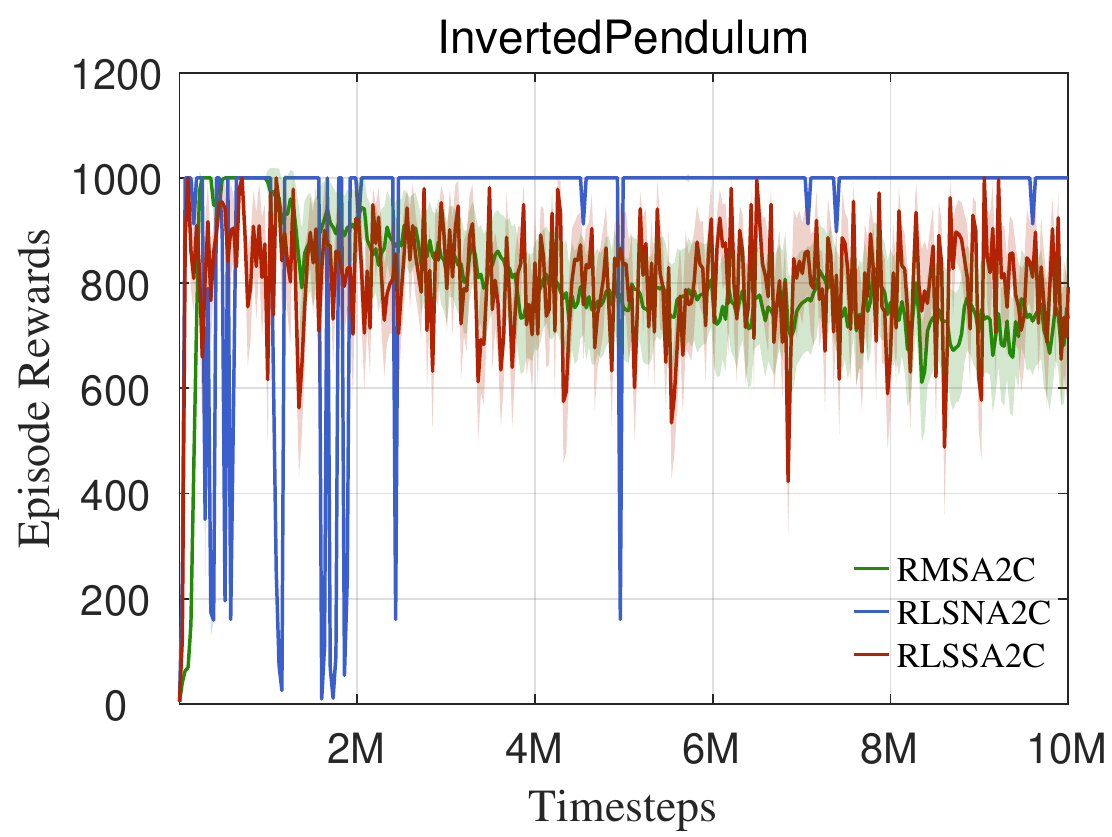}}
\subfigure{\includegraphics[width=4.2cm]{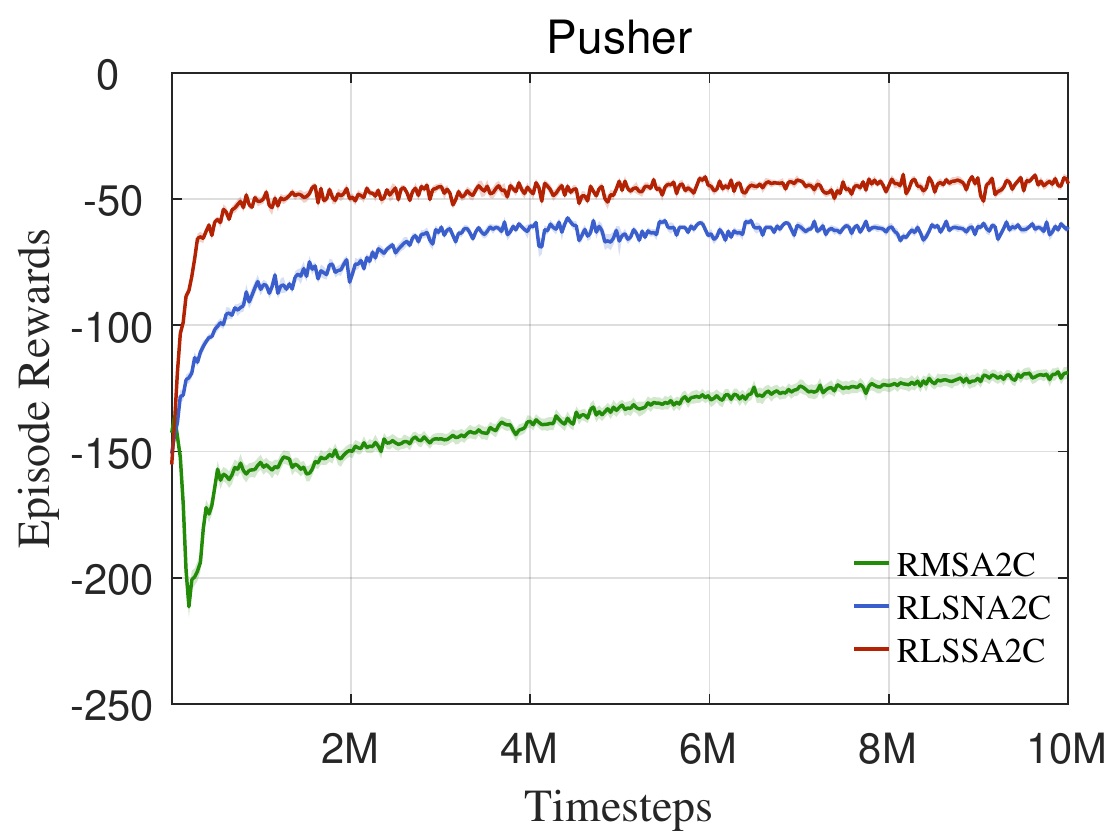}}
\subfigure{\includegraphics[width=4.2cm]{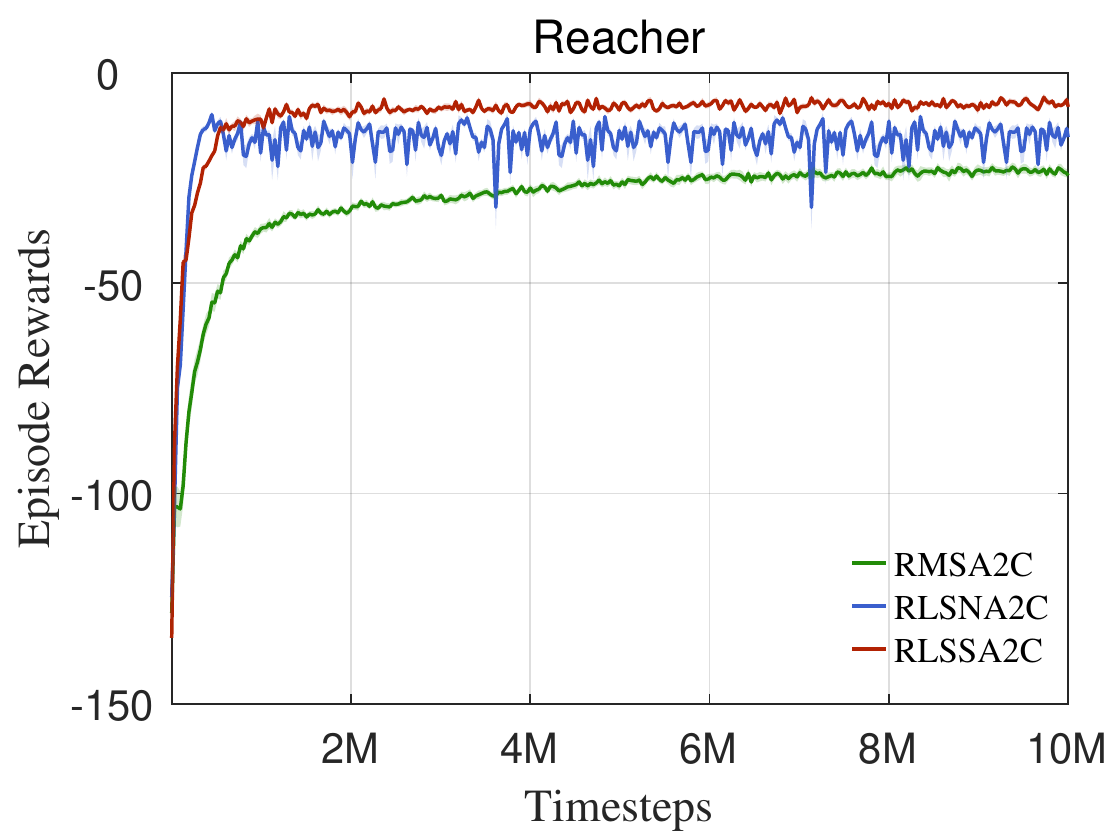}}
\subfigure{\includegraphics[width=4.2cm]{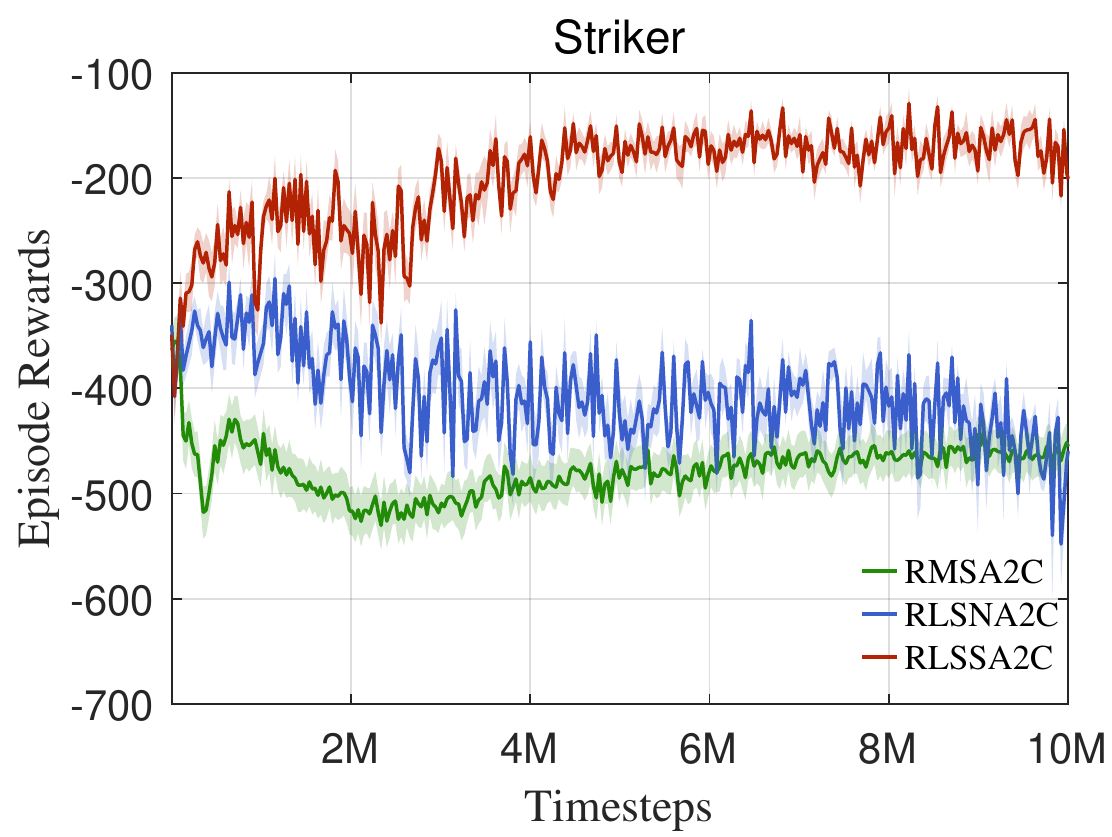}}
\centering
\subfigure{\includegraphics[width=4.2cm]{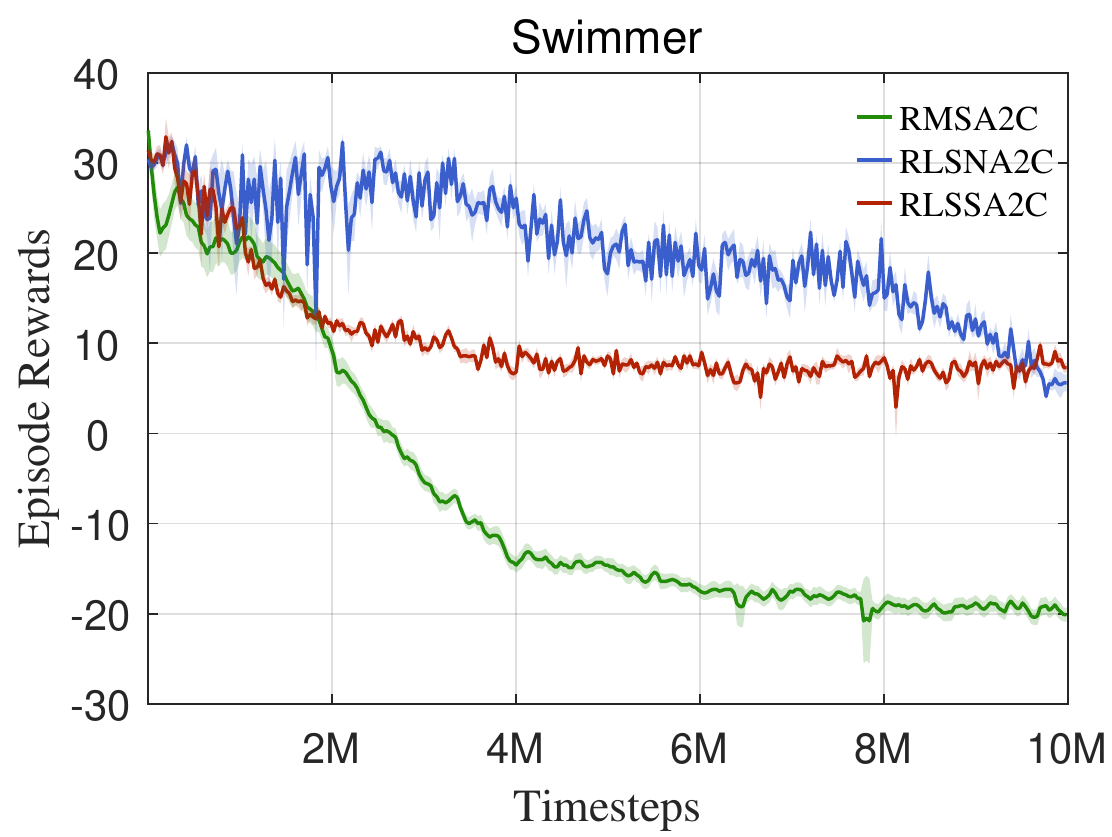}}
\subfigure{\includegraphics[width=4.2cm]{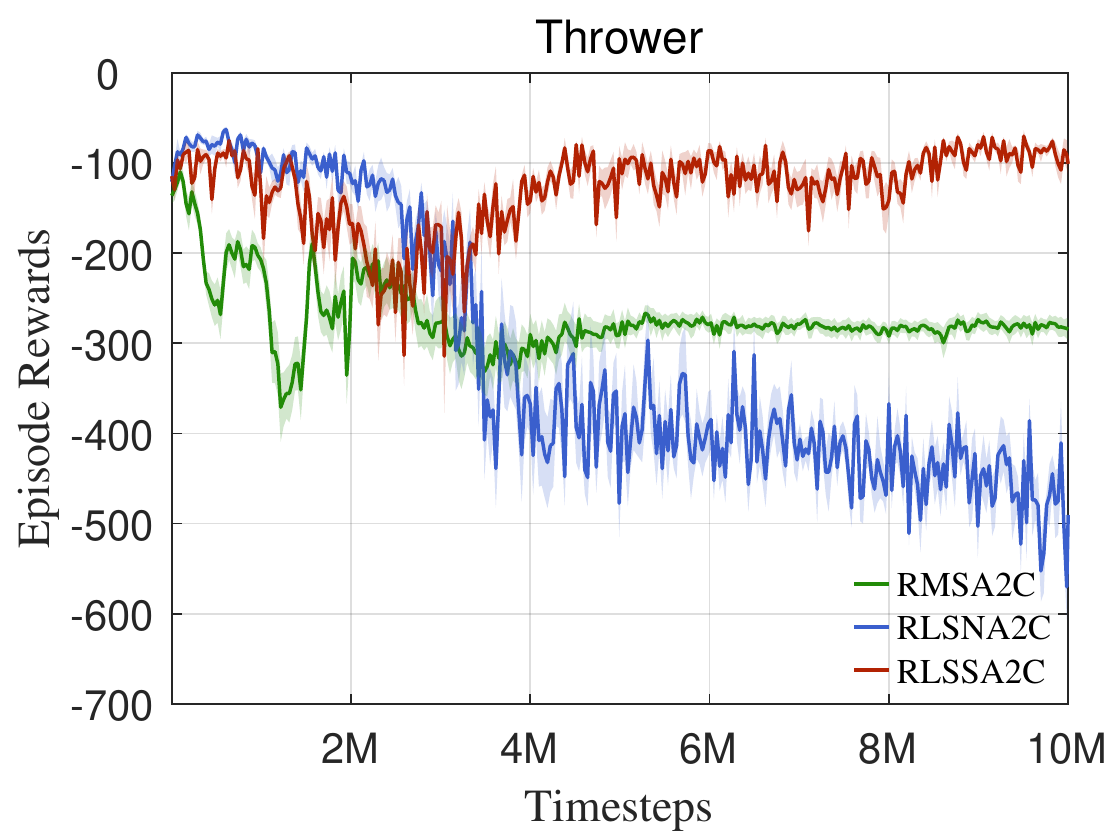}}
\subfigure{\includegraphics[width=4.2cm]{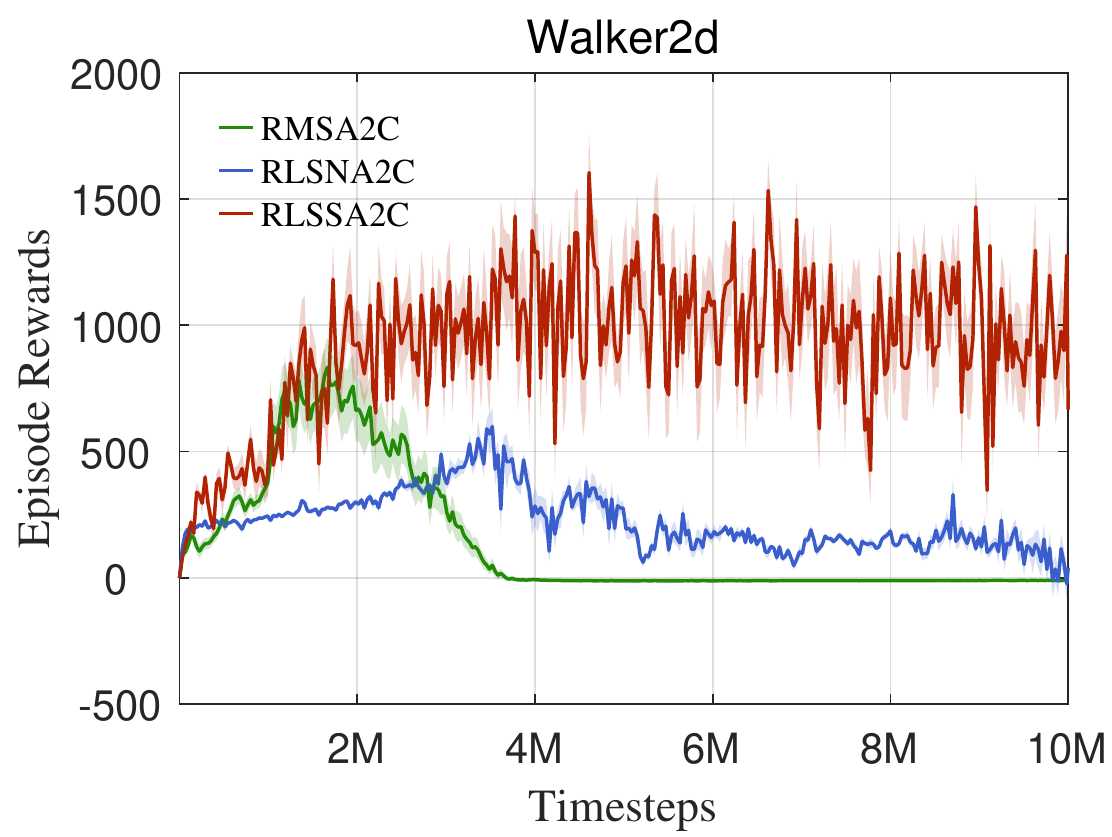}}

\caption{Convergence comparison of our algorithms against RMSA2C on 11 MuJoCo tasks trained for 10M timesteps.}
\end{figure*}

\renewcommand{\arraystretch}{1.5}
\begin{table*}[tp]
  \centering
  \fontsize{7.4}{9}\selectfont
\begin{floatrow}
\capbtabbox{
    \begin{tabular*}{0.45\textwidth}{lrrr}
    \toprule[1pt]
    Game&RMSA2C&~~~~RLSSA2C&~~~~RLSNA2C\cr
    \midrule[1pt]
    Ant                   &-62.0 &\textbf{4607.0} &941.5  \cr
    HalfCheetah           &454.6 &\textbf{1630.3} &1625.4  \cr
    Hopper                &921.9 &2068.9 &\textbf{2133.6}  \cr
    InvertedDoublePendulum&7957.0 &\textbf{9333.4} &8500.0  \cr
    InvertedPendulum      &1000.0 &1000.0 &1000.0  \cr
	Pusher                &-117.7 &\textbf{-39.0} &-56.6  \cr
    \bottomrule[1pt]
    \end{tabular*}
    \hfil\hfil\hfil\hfil\hfil\hfil~~~~~~~~~~\hfil\hfil\hfil\hfil\hfil\hfil\hfil\hfil\hfil\hfil
    \begin{tabular*}{0.45\textwidth}{lrrr}
    \toprule[1pt]
    Game&~~~~~~~~~~~~~~~~~~~~~RMSA2C&~~~~~RLSSA2C&~~~~~RLSNA2C\cr
    \midrule[1pt]
    Reacher               &-21.9 &\textbf{-4.9} &-9.2 \cr
    Striker               &-336.1 &\textbf{-124.7} &-264.3  \cr
    Swimmer               &\textbf{33.6} &32.9 &32.3  \cr
    Thrower               &-106.2 &-68.5 &\textbf{-59.3}  \cr
    Walker2d              &1682.0 &\textbf{3306.9} &1363.6  \cr
    \cr
    \bottomrule[1pt]
    \end{tabular*}
}{
    \caption{\textsc{Top 10 Average Episode Rewards of our Algorithms and RMSA2C on 11 MuJoCo Tasks Trained for 10M Timesteps}}}
\end{floatrow}
\end{table*}

\renewcommand{\arraystretch}{1.5}
\begin{table}[tp]
  \centering
  \fontsize{7.4}{9}\selectfont
  \begin{threeparttable}
  \caption{\textsc{Running Speed Comparison on Six MuJoCo Tasks}\\{( Timesteps\! /\! Second )}}
  \label{tab:performance_comparison}
    \begin{tabular*}{\textwidth}{lccccc}
    \toprule[1pt]
    Game&RMSA2C&PPO&ACKTR&RLSSA2C&RLSNA2C\cr
    \midrule[1pt]
    Ant             &5911 &372 &4584 &5646 &5380 \cr
    Hopper          &7735 &375 &5574 &7160 &6909 \cr
    InvertedPendulum&10076 &384 &6633 &8997 &8399 \cr
    Pusher          &7873 &368 &5558 &7185 &7073 \cr
    Swimmer         &9033 &384 &6068 &7996 &7927 \cr
	Walker2d        &7716 &373 &5438 &7052 &6855 \cr
    \midrule[1pt]
    Mean            &8057 &376 &5643 &7339 &7091 \cr
    \bottomrule[1pt]
    \end{tabular*}
    \end{threeparttable}
\end{table}

All tested algorithms also use two disjoint FNNs defined in \cite{15}. One is the critic network, the other is the actor network.
Both networks have two same fc hidden layers with 64 Tanh neurons. The critic output layer is a linear fc layer for predicting the value function.
The actor network uses a linear fc layer with bias to represent the mean and the standard deviation of the Gaussian policy \cite{24}.
All settings for five tested algorithms are the same as those in Section \uppercase\expandafter{\romannumeral5}. \textit{A}, except for $\alpha=0.001$ in RLSNA2C.

The convergence comparison of our algorithms against RMSA2C on 11 MuJoCo tasks trained for 10 million timesteps is shown in Fig. 2.
It is clear that RLSSA2C and RLSNA2C outperform RMSA2C on almost all tasks.
Among these three algorithms, RLSSA2C has the best convergence performance and stability.
Compared with RMSA2C, RLSSA2C wins on 10 tasks. On Ant, HalfCheetah, Hopper, Pusher, Reacher, Striker, Swimmer, Thrower and Walker2d tasks, RLSSA2C is significantly superior to RMSA2C in terms of convergence speed and convergence quality. Compared with RMSA2C, RLSNA2C also wins on 9 tasks. On Pusher,
Reacher, and Swimmer tasks, RLSNA2C performs very well.

In Table \uppercase\expandafter{\romannumeral3}, we present the top 10 average episode rewards of our algorithms and RMSA2C on 11 MuJoCo tasks trained for 10 million timesteps. From Table \uppercase\expandafter{\romannumeral3}, RLSSA2C and RLSNA2C obtain the highest rewards on 7 and 2 tasks respectively, but RMSA2C only wins on 1 task.

The running speed comparison of our algorithms against RMSA2C, PPO and ACKTR on six MuJoCo tasks is shown in Table \uppercase\expandafter{\romannumeral4}. Among these five algorithms, RMSA2C has the highest computational efficiency, RLSSA2C and RLSNA2C are listed 2nd and 3rd respectively, and PPO is listed last. In detail, RLSSA2C is only 8.9\% slower than RMSA2C, but is  1851.9\% and 30.1\% faster than PPO and ACKTR.  RLSNA2C is only 12.0\% slower than RMSA2C, but is  1785.9\% and 25.7\% faster than PPO and ACKTR.

Obviously, RLSSA2C and RLSNA2C also perform very well on MuJoCo tasks. In summary, our both algorithms have better
sample efficiency than RMSA2C, and have higher computational efficiency than PPO and ACKTR.

\section{Conclusion}
In this paper, we proposed two RLS-based A2C algorithms, called RLSSA2C and RLSNA2C. To the best of our knowledge,
they are the first RLS-based DAC algorithms. Our both algorithms use the RLS method to train the critic network and
hidden layers of the actor network. Their main difference is their policy learning. RLSSA2C uses SPG and an ordinary first-order gradient descent algorithm to learn the policy parameter, but RLSNA2C uses NPG, the Kronecker-factored approximation and the RLS method to learn the
compatible parameter and the policy parameter. We also analyzed the complexity and convergence of our both algorithms, and presented three tricks for
further accelerating their convergence.  We tested our both algorithms on 40 Atari discrete control games and 11 MuJoCo continuous control tasks.
Experimental results show that our both algorithms have better sample efficiency than RMSA2C on most games or tasks, and have higher computational efficiency than PPO and ACKTR. In future work, we will try to establish the convergence of our both algorithms and improve the stability of RLSNA2C.

\end{document}